\documentclass[final]{cvpr}

\usepackage{times}
\usepackage{epsfig}
\usepackage{graphicx}
\usepackage{amsmath}
\usepackage{amssymb}
\usepackage{verbatim}
\usepackage{caption}
\usepackage{subcaption}
\usepackage{rotating}
\usepackage{multirow}
\usepackage{color}
\usepackage{url}
\usepackage{booktabs}
\usepackage{bbm}
\usepackage{sidecap}
\usepackage{enumitem}
\usepackage{enumerate}

\usepackage[pagebackref=true,breaklinks=true,colorlinks,bookmarks=false]{hyperref}

\newif\ifshowcomments
\showcommentstrue

\ifshowcomments

\newcommand{\TODO}[1]{{\color{red}{[TODO: #1]}}}

\newcommand{\revised}[1]{{\color[rgb]{0.2,0.7,0.2}{#1}}}

\else
\newcommand{\TODO}[1]{}
\newcommand{\revised}[1]{}
\newcommand{\phil}[1]{}
\fi

\begin{document}

\title{Deep Texture-Aware Features for Camouflaged Object Detection}


\author{Jingjing Ren$^{1, \ast}$, Xiaowei Hu$^{2, }$\thanks{Jingjing Ren and Xiaowei Hu are the joint first authors of this work.}, Lei Zhu$^{2}$, Xuemiao Xu$^{1, }$\thanks{Corresponding author (xuemx@scut.edu.cn)}, Yangyang Xu$^{1, }$, \\ Weiming Wang$^{3}$, Zijun Deng$^{1, }$, and Pheng-Ann Heng$^{2, }$ \\
        $^1$ South China University of Technology,  $^2$ The Chinese University of Hong Kong,\\
        $^3$ The Open University of Hong Kong\\}

\maketitle

\begin{abstract}

Camouflaged object detection is a challenging task that aims to identify objects having similar texture to the surroundings. 
This paper presents to amplify the subtle texture difference between camouflaged objects and the background for camouflaged object detection by formulating multiple texture-aware refinement modules to learn the texture-aware features in a deep convolutional neural network.
The texture-aware refinement module computes the covariance matrices of feature responses to extract the texture information, designs an affinity loss to learn a set of parameter maps that help to separate the texture between camouflaged objects and the background, and adopts a boundary-consistency loss to explore the object detail structures.
We evaluate our network on the benchmark dataset for camouflaged object detection both qualitatively and quantitatively.
Experimental results show that our approach outperforms various state-of-the-art methods by a large margin.

\end{abstract}



\begin{figure*}[h]
	\centering
	\vspace*{5mm}
	\begin{subfigure}{0.19\textwidth} 
	\includegraphics[width=\textwidth, height=0.9\textwidth]{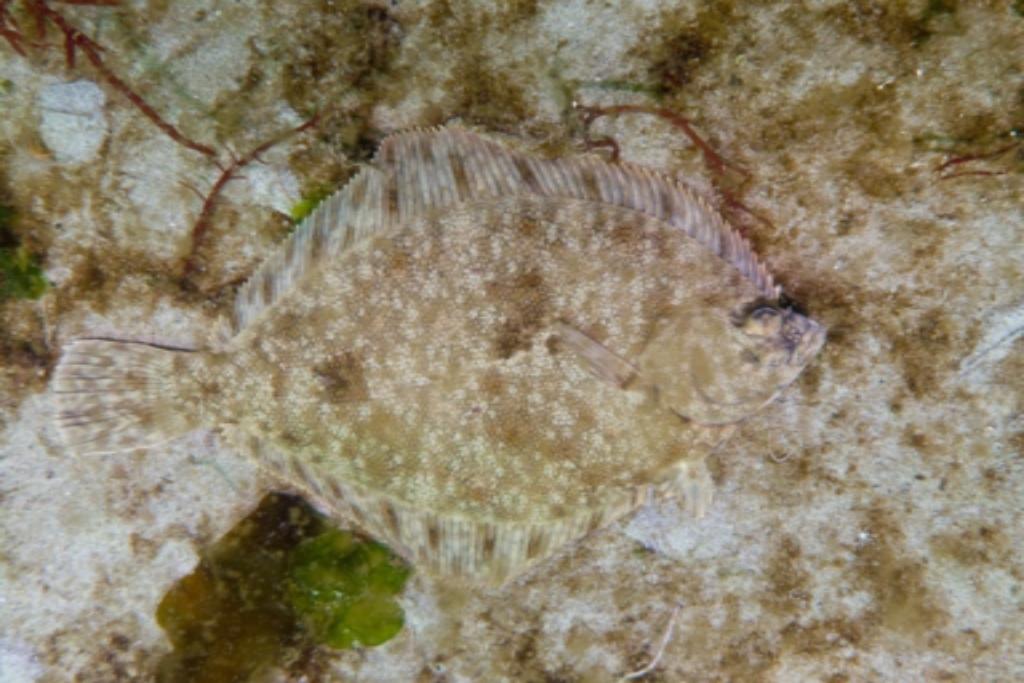}\vspace*{-1mm}\caption{input image}
	\end{subfigure}
	\begin{subfigure}{0.19\textwidth}
		\includegraphics[width=\textwidth, height=0.9\textwidth]{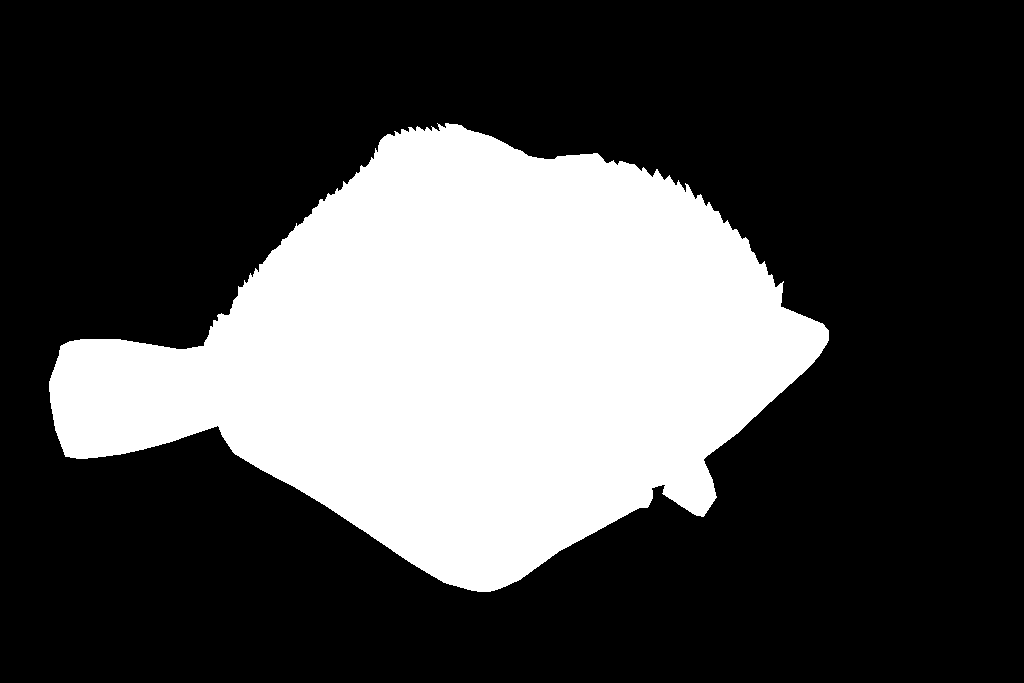}\vspace*{-1mm}\caption{ground truth}
	\end{subfigure}
	\begin{subfigure}{0.19\textwidth}
		\includegraphics[width=\textwidth, height=0.9\textwidth]{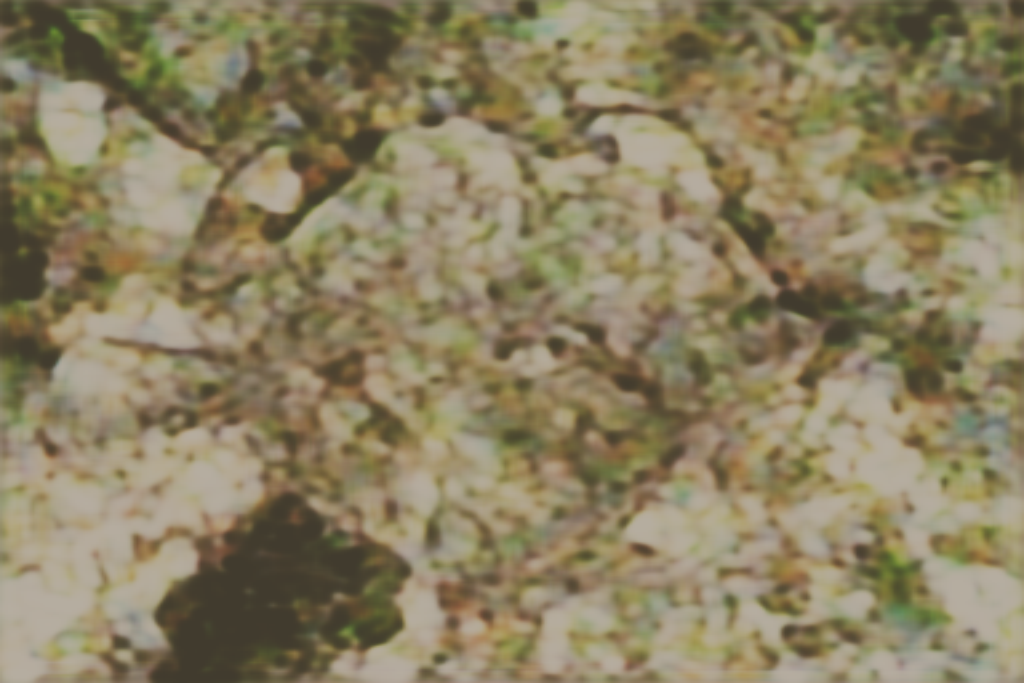}\vspace*{-1mm}\caption{convolutional feature}
	\end{subfigure}
	\begin{subfigure}{0.19\textwidth}
		\includegraphics[width=\textwidth, height=0.9\textwidth]{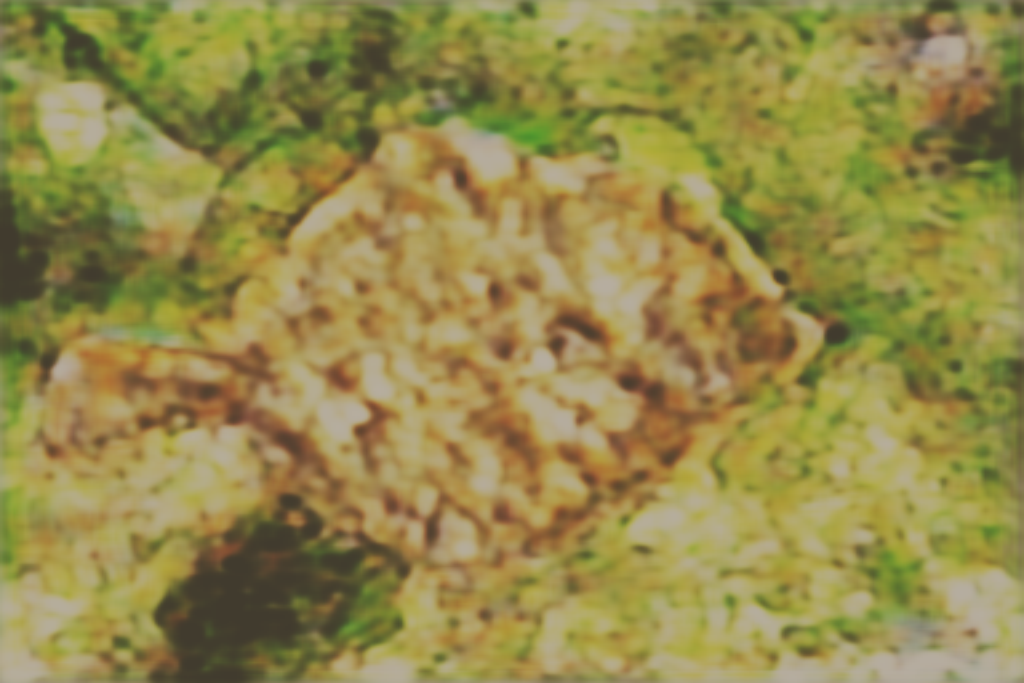}\vspace*{-1mm}\caption{textrue-aware feature}
	\end{subfigure}
	\begin{subfigure}{0.19\textwidth}
		\includegraphics[width=\textwidth, height=0.9\textwidth]{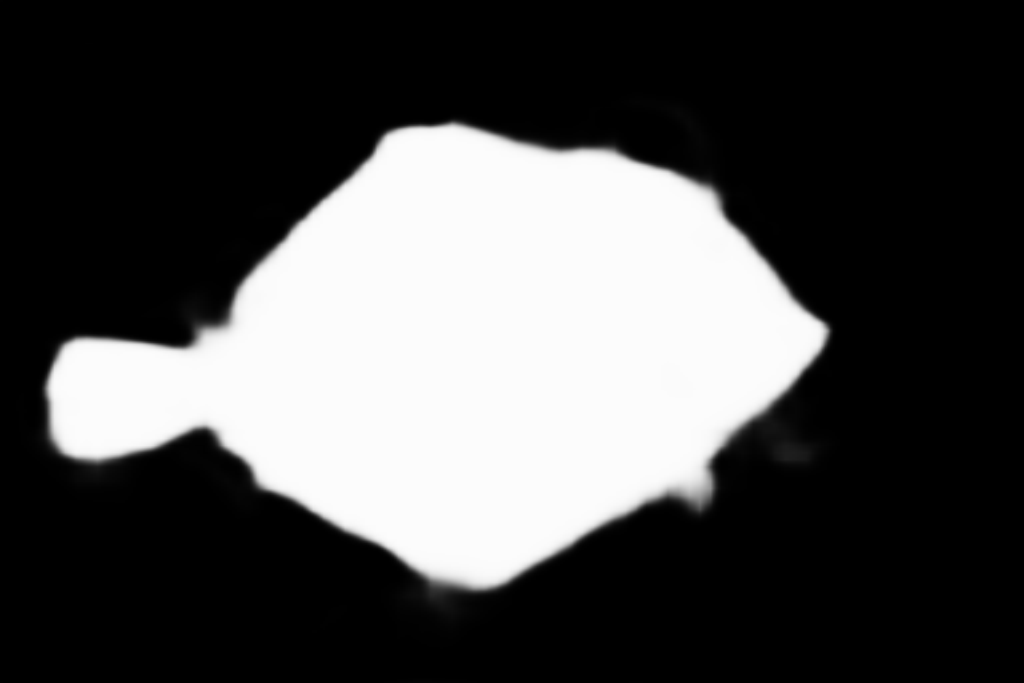}\vspace*{-1mm}\caption{our result}
	\end{subfigure}
	\ \\
	\caption{Visualization results of the reconstructed images from the original convolutional features and texture-aware features produced from our TARM.}
	\label{fig:fish}
\end{figure*}
\section{Introduction}
\label{sec:introduction}

In nature, animals try to conceal themselves by adapting the texture of their bodies to the texture of the surroundings, which helps them avoid being recognized by predators.
This strategy is easily used to deceive the visual perceptual system \cite{stevens2009animal} and current vision algorithms may fail to distinguish camouflaged objects from the background.

Hence, camouflaged object detection~\cite{fan2020camouflaged} has been a great challenge and solving this problem could benefit a lot of applications in computer vision, such as polyp segmentation \cite{fan2020pranet}, lung infection segmentation \cite{fan2020inf, wu2020jcs}, photo-realistic blending \cite{ge2018image}, and recreational art \cite{chu2010camouflage}.

To solve this problem, Fan~\textit{et al}.~\cite{fan2020camouflaged} collects the first large-scale dataset for camouflaged object detection.
The dataset includes $10,000$ images with over $78$ object categories in various natural scenes, which make it possible to apply deep learning algorithms to learn to recognize camouflaged objects from large data. 
Camouflaged objects usually have similar texture to the surroundings, and deep learning algorithms designed for generic object detection \cite{ren2015faster, redmon2016you, he2017mask, chen2019hybrid,zhao2017pyramid, ronneberger2015u, chen2017deeplab} and salient object detection \cite{zhao2019pyramid, zhao2019egnet} generally do not perform well to detect camouflaged objects in such difficult situations.
Recently, algorithms for camouflaged object detection based on deep neural networks have been proposed.
SINet~\cite{fan2020camouflaged} first adopts a search module to find the candidate regions of camouflaged objects and then uses an identification module to precisely detect camouflaged objects.
ANet~\cite{le2019anabranch} first leverages a classification network to identify whether the image contains camouflaged objects or not, and then adopts a fully convolutional network for camouflaged object segmentation.
However, these algorithms may still misunderstand camouflaged objects as the background due to their similar texture.

%
Texture refers to a particular way how visual primitives are organized spatially in natural images \cite{liu2019bow}.
Essentially, there are subtle differences in the texture between camouflaged objects and the background. 
%
%
%
As shown in Figure~\ref{fig:fish} (a), the texture of the fish involves a combination of dense and small white particles and brown regions while the texture of background has the combination of white and brown regions. 
Based on this observation, we present to amplify the texture difference between camouflaged objects and the background by learning texture-aware features from the deep neural network, thus improving the performance of camouflaged object detection; see Figure~\ref{fig:fish} (c)-(e).
%

To achieve this, we design the texture-aware refinement module (TARM) in a deep neural network, where we first compute the covariance matrices of feature responses to extract the texture information from the convolutional features, and then learn a set of affinity functions to amplify the texture difference  between camouflaged objects and the background. 
Moreover, we design a boundary-consistency loss in TARM to improve the segmentation quality by revisiting the image patches across boundaries on high-resolution feature maps. 
After that, we adopt multiple TARMs in a deep network (TANet) to learn deep texture-aware features in different layers and predict a detection map per layer for camouflaged object detection.
Finally, we qualitatively and quantitatively compare our approach with $13$ state-of-the-art methods designed for camouflaged object detection, salient object detection, and semantic segmentation on the benchmark dataset, showing the superiority of our network. 

We summarize the contributions of this work as follow:
\begin{itemize}[]
	
	\vspace*{-0.25mm}
	\item
	First, we design a novel texture-aware refinement module (TARM) to amplify the texture difference between camouflaged objects and the background, which significantly enhances camouflaged object recognition.
	
	\item
	Second, we design a boundary-consistency loss to enhance detail information across boundaries without extra computation overhead in testing. 
	
	\item
	Third, we evaluate our network on the benchmark dataset and compare it with $13$ state-of-the-art methods on camouflaged object detection, salient object detection, and semantic segmentation.
	Qualitative and quantitative results show that our approach outperforms previous methods by a large margin.

\end{itemize}


\section{Related Work}
\label{sec:related}

\vspace*{3mm}
\noindent
{\bf Camouflaged object detection.}
Early work for camouflaged object detection (COD) adopted various hand-crafted features, \eg, color \cite{huerta2007improving, siricharoen2010robust}, convex intensity \cite{tankus2001convexity}, edge \cite{siricharoen2010robust}, and texture~\cite{bhajantri2006camouflage, kavitha2011efficient}. 
%
%
Recently, deep convolution neural network achieves great success with the help of large-scale camouflaged object datasets~\cite{skurowski2018animal, le2019anabranch, fan2020camouflaged}. 
SINet~\cite{fan2020camouflaged} found the candidate regions of camouflaged objects by a search module and precisely detected  camouflaged objects through an identification module.
ANet~\cite{le2019anabranch} first identified the image that contains camouflaged objects through a classification network and then adopted a fully convolutional network for camouflaged object segmentation.
These methods, however, do not consider the subtle texture difference between camouflaged objects and the background, and may fail to detect camouflaged objects in complex situations.
%

\vspace*{3mm}
\noindent
{\bf Salient object detection and semantic segmentation.}
Salient object detection (SOD) predicts a binary mask to indicate the saliency regions
while
semantic segmentation (SS) aims to generate masks with category labels to identify the image regions with different classes.
The deep-learning-based methods designed for SOD~\cite{liu2018picanet,qin2019basnet,wu2019cascaded,zhao2019egnet,zhao2019pyramid} and SS~\cite{he2017mask,zhou2018unet++,zhao2017pyramid, chen2019hybrid, huang2019mask} can be employed for camouflaged object detection by retraining them on camouflaged object datasets, and we compare our network with these methods for camouflaged object detection on the benchmark dataset; see Section~\ref{sec:experiments}.

\begin{figure}[htbp]
	\centering
	\includegraphics[width=1\linewidth]{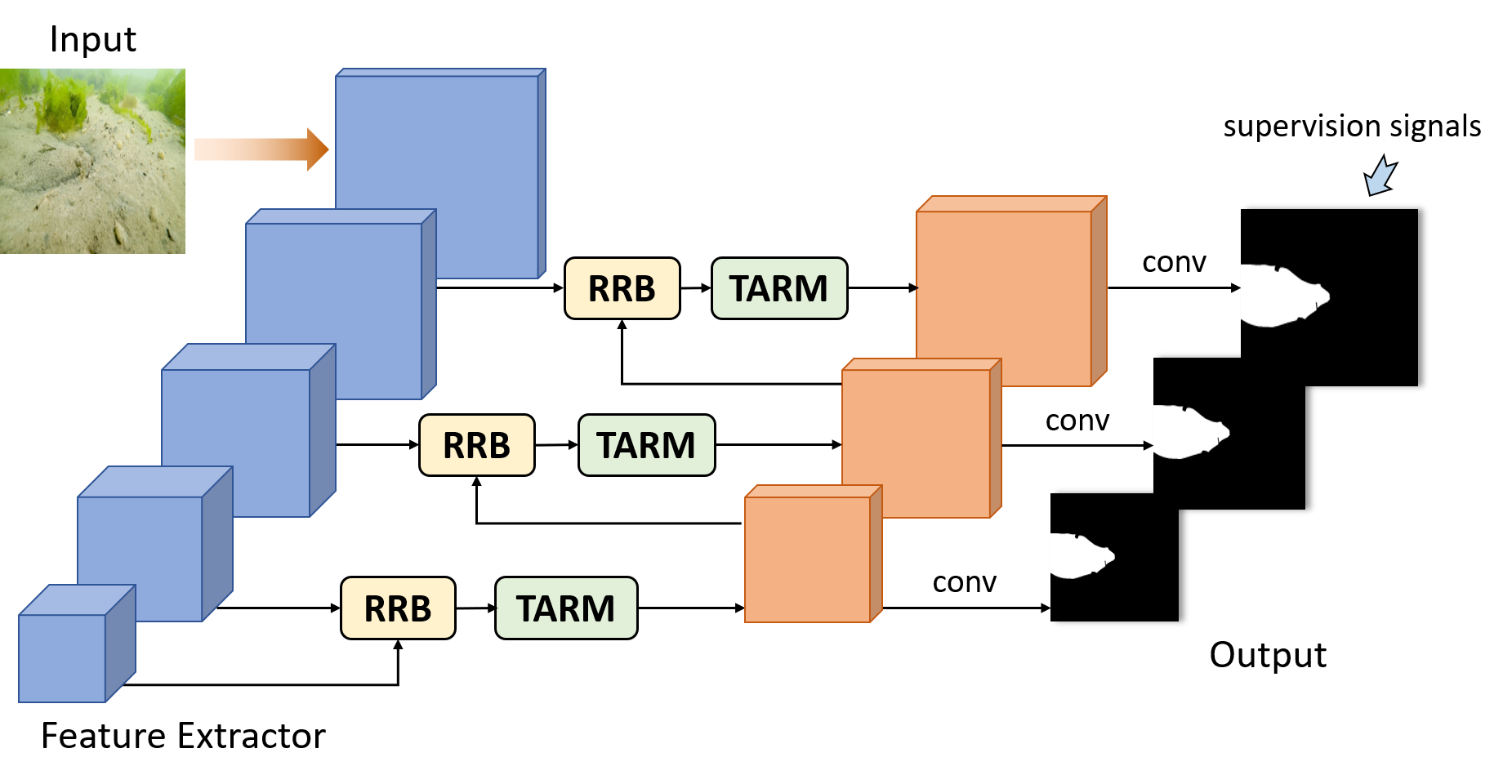}
	\caption{The schematic illustration of the overall network architecture (TANet) to learn the deep texture-aware features for camouflaged object detection. Given the input image, we use a feature extractor to produce the feature maps with multiple resolutions. At each layer, the feature map is first refined by the residual refine block (RRB) and then further enhanced by our texture-aware refinement module (TARM) in a texture-aware manner. We adopt the predicted mask with highest resolution as the final output of our network.}
	\label{fig:overview}
\end{figure}

\begin{figure*}[htbp]
	\centering
	\includegraphics[width=1\textwidth]{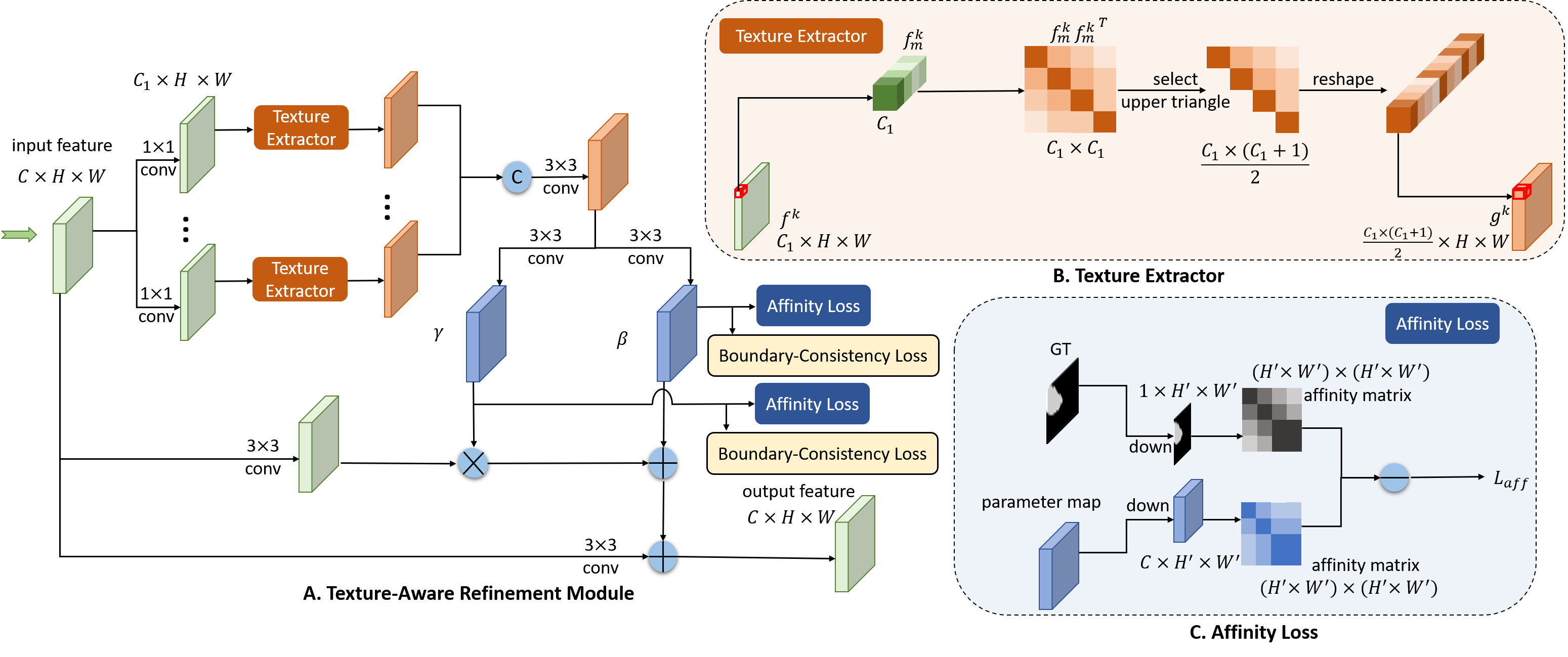}
	\caption{The schematic illustration of the texture-aware refinement module (TARM). We compute the deep texture-aware features by formulating covariance matrices to extract the texture features in different aspects and learning the parameter maps through affinity and boundary-consistency losses to amplify the texture difference between the camouflaged objects and their surroundings.}
	\label{fig:TRM}
\end{figure*}

\section{Methodology}

Figure~\ref{fig:overview} shows the overall network architecture (TANet) with texture-aware refinement modules (TARM) for camouflaged object detection. 
Given the input image, we adopt the feature extractor to extract the feature maps with multiple resolutions, and then use the residual refine blocks (RRB)  to refine the feature maps at different layers to enhance the fine details and remove the background noises. 

We ignore to refine the feature maps at the first layer due to the large memory footprint. 
Next, we present the texture-aware refinement module (TARM) to learn the texture-aware features, which help to improve the visibility of camouflaged objects.
Lastly, we predict the binary masks to indicate the camouflaged objects at each layer by adding the supervision signals at multiple layers. 

In the following subsections, we will elaborate on the texture-aware refinement module (Section~\ref{subsec:TARM}) in detail, present the training and testing strategies in Section~\ref{subsec:train_test}, and visualize the learned texture-aware features in Section~\ref{subsec:visualization}.

\subsection{Texture-Aware Refinement Module}
\label{subsec:TARM}

As shown in Figure~\ref{fig:fish}, the camouflaged objects have similar texture with the surroundings.
%
%
However, there still exists subtle texture difference between the camouflaged objects and the background.
Hence, we present a texture-aware refinement module to extract texture information and amplify the texture difference of camouflaged objects and background, thus improving the performance for camouflaged object detection. 

\subsubsection{Architecture}
Figure~\ref{fig:TRM} shows the architecture of the proposed texture-aware refinement module.
First, it takes a feature map with the resolution $C\times H\times W$ as the input, and then use multiple convolution operations with $1\times 1$ kernel to obtain multiple kinds of feature maps \cite{song2019autoint} and each with the size of $C_1\times H\times W$. 
Note that $C_1$ is smaller than $C$ for the computational efficiency and these feature maps are used to learn multiple aspects of textures in the following operations.  
Next, we compute the co-variance matrix among the feature channels at each position to capture the correlations between different responses on the convolutional features.
The co-variance matrix among features measures the co-occurrence of features,  describes the combination of features, and is used to represent texture information ~\cite{karacan2013structure,gatys2016image}.
%
%
As shown in Figure~\ref{fig:TRM}(b), for each pixel $f_m^k$ on the feature map $f^k \in C_1 \times H \times W$, we compute its covariance matrix as the the inner product between $f_m^k$ and ${f_m^k}^T$.
Since the covariance matrix ($C_1 \times C_1$) has the property of diagonal symmetry, we just adopt the upper triangle of this matrix to represent the texture feature, and reshape the result into a feature vector. 
We perform the same operations for each pixel on the feature map and concatenate the results to obtain $g^k \in \frac{C_1 \times (C_1+1)}{2} \times H \times W$, which contain the texture information.
Then, we fuse all the covariance matrices computed from different feature maps by a $3\times 3$ convolution.
After that, we adopt two sets of $3 \times 3$ convolution from the texture features to learn two parameter maps $\gamma \in C'\times H\times W$ and $\beta \in C'\times H\times W$ ($C'$ denotes the channel number) which are used to amplify the texture difference of camouflaged objects and their surroundings by adjusting the texture of input features $f_{in}$~\cite{huang2017arbitrary, park2019semantic}. Finally, we obtain the output feature $f_{out}$ by:
\begin{equation}
f_{out} \ =  \ conv(\gamma\frac{f_{in}^{'}  \ - \ \mu(f_{in}^{'} )}{\sigma(f_{in}^{'})} + \beta) \ + \ f_{in} \ , 
\end{equation}
where $f_{in}^{'}$ is obtained by applying a $3 \times 3$ convolution on $f_{in}$. $\mu(f_{in}^{'}) $ and $\sigma(f_{in}^{'})$ are the mean and variance of $f_{in}^{'}$, which are used to normalize the feature map. 
Finally, we add the original feature map with the feature maps refined by the above operations as the output of our texture-aware refinement module.

\begin{figure}[tp]
	\centering
	\includegraphics[width=1\linewidth]{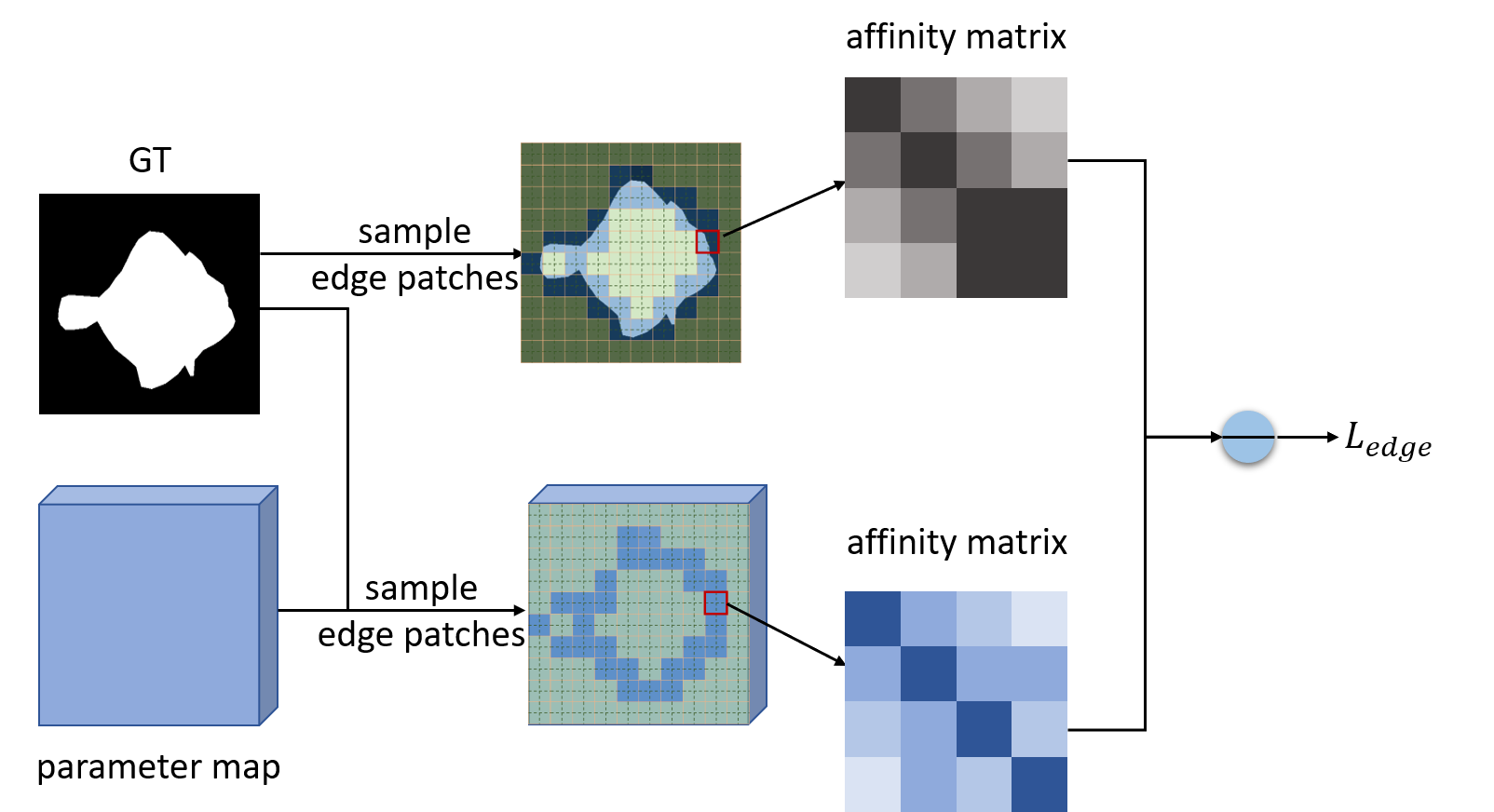}
	\caption{The schematic illustration of how boundary-consistency loss works. Affinity loss is computed within small patches across boundary, which is used to enhance detail information.}
	\label{fig:edge}
\end{figure}

\subsubsection{Loss Function}

\paragraph{Affinity loss.}
To make the parameter maps $\gamma$ and $\beta$ capture the texture difference between the camouflaged objects and the background, we adopt the affinity loss~\cite{yu2020context} on $\gamma$ and $\beta$ to explicitly amplify the difference between their texture features.
As shown in Figure~\ref{fig:TRM} (c), we first use the pooling operations to downsample the parameter map and then compute the affinity matrix $A^h_{m,n}$ at the position $m,n$:
\begin{equation}
A^h_{m,n} \ = \ \frac{{h_m}^Th_n}{\parallel h_m \parallel_2\parallel h_n \parallel_2} \ ,
\end{equation}
where $h_m$ and $h_n$ are the parameter vectors of the downsampled map.
$A^h$ is the result matrix that captures the pairwise texture similarity. 
Next, we calculate the ground truth affinity matrix $A^{gt}$ by:
\begin{equation}
A^{gt}_{m,n} \ = \ 2 \ \times \ \mathbbm{1}_{\{C_m=C_n\}} \ - \ 1 \ ,
\end{equation}
where $\mathbbm{1}$ is a indicator, which is equal to one when the labels ($C_m$ and $C_n$) of positions $m$ and $n$ are the same, otherwise it is equal to zero.

In natural images, camouflaged objects usually occupy small regions, and we formulate the affinity loss as follow to solve the  class imbalance problem:
\begin{equation}
L_{aff} \ = \ \sum_{m}\sum_{n}w_m w_n d(A^h_{m,n}, A^{gt}_{m,n}) \ ,
\end{equation}
where $w_m=1-\frac{N_{C_m}}{H'W'}$ and $w_n=1-\frac{N_{C_n}}{H'W'}$; 
$N_{C_m}$ and  $N_{C_n}$ are the numbers of pixels that have the same class label with the pixel $m$ and $n$;
$H'$ and $W'$ are the height and width of the parameter map.
%
%
From this loss function, we can see that 
the parameter map will learn to maximize the texture difference,

\begin{figure}[tp]
	\centering
	\begin{minipage}{0.05\linewidth}
		\begin{sideways}input\end{sideways}
	\end{minipage}
	\begin{subfigure}{0.3\linewidth}
		\includegraphics [width=\linewidth, height =\linewidth ] {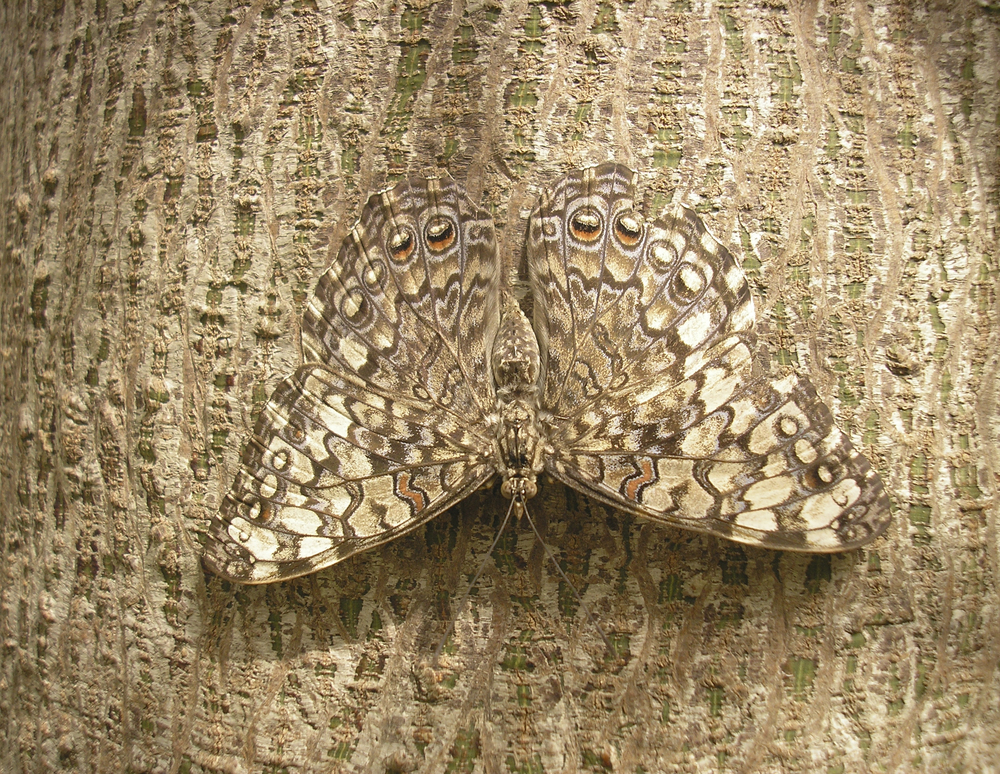}
	\end{subfigure}
	\begin{subfigure}{0.3\linewidth}
		\includegraphics[width=\linewidth, height =\linewidth ] {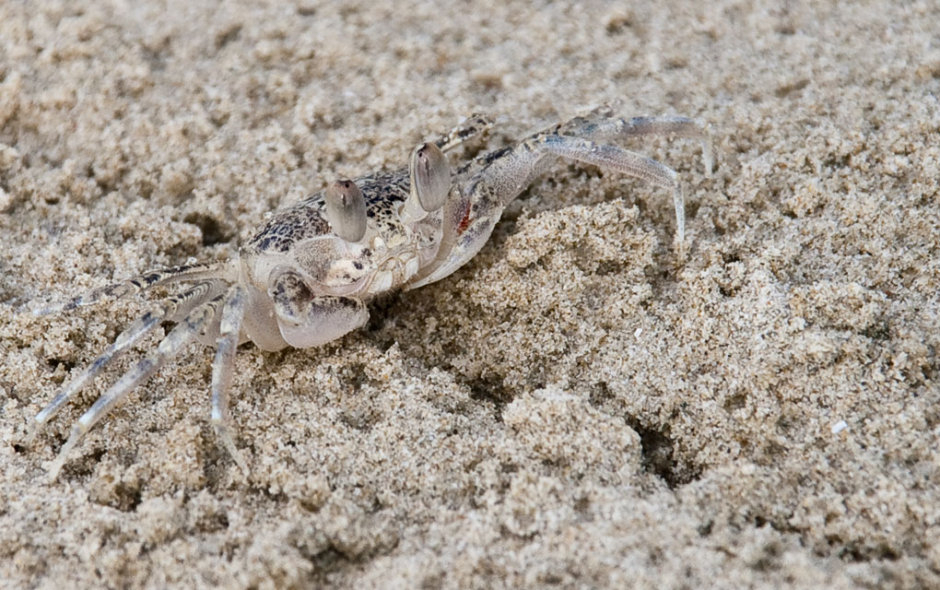}
	\end{subfigure}
	\begin{subfigure}{0.3\linewidth}
		\includegraphics[width=\linewidth, height =\linewidth ]{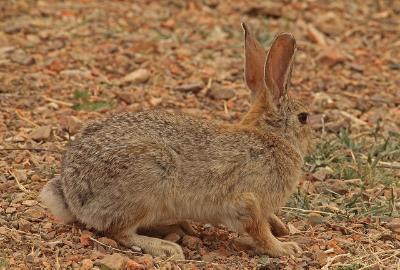}
	\end{subfigure}
	\ \\
	\vspace*{0.5mm}
	\begin{minipage}{0.05\linewidth}
		\begin{sideways}ground truth\end{sideways}
	\end{minipage}
	\begin{subfigure} {0.3\linewidth}
		\includegraphics [width=\linewidth, height =\linewidth ]{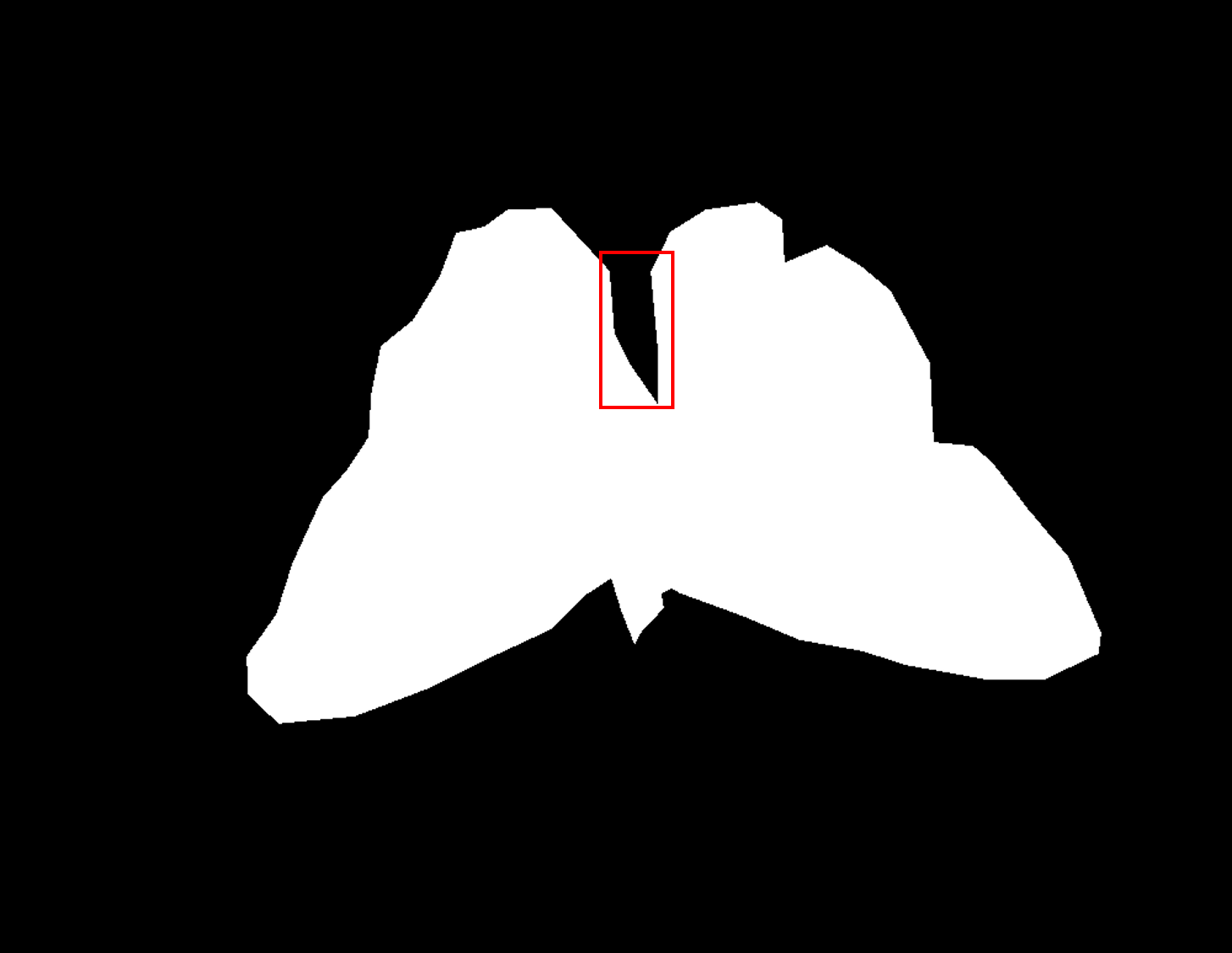}
	\end{subfigure}
	\begin{subfigure}{0.3\linewidth}
		\includegraphics [width=\linewidth, height =\linewidth ]{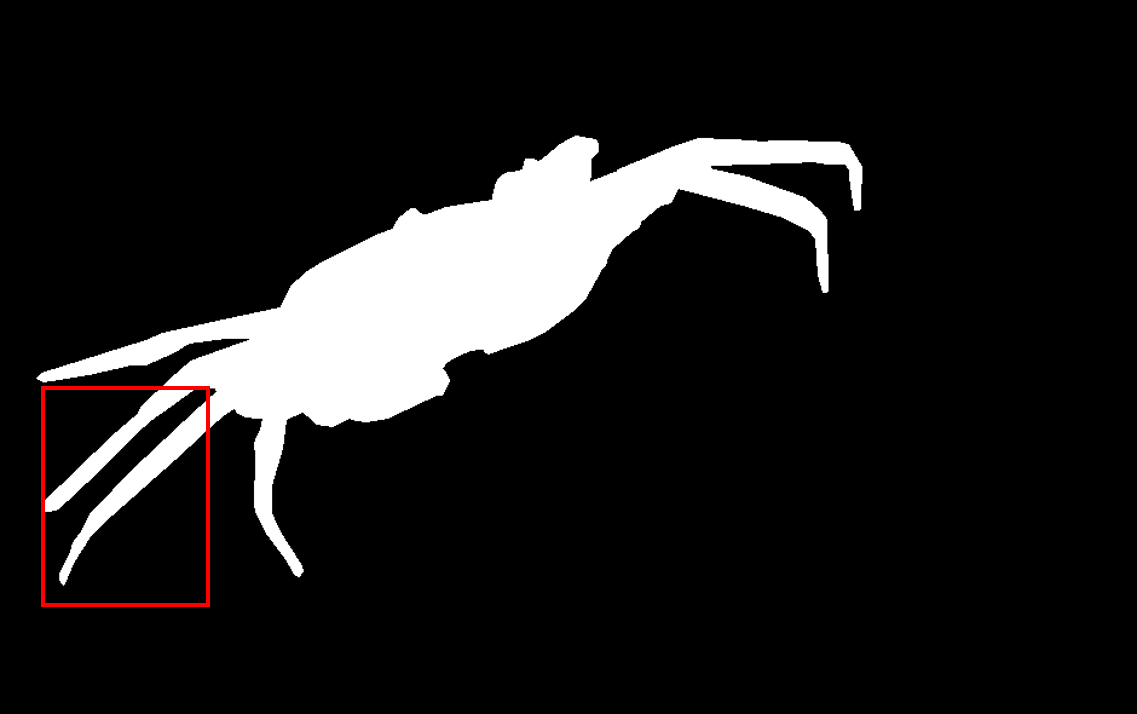}
	\end{subfigure}
	\begin{subfigure}{0.3\linewidth}
		\includegraphics [width=\linewidth, height =\linewidth ]{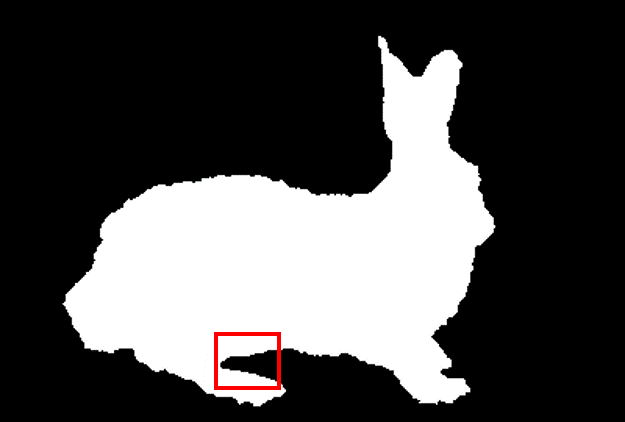}
	\end{subfigure}
	\ \\
	\vspace*{0.5mm}
	\begin{minipage}{0.05\linewidth}
		\begin{sideways}w/o edge\end{sideways}
	\end{minipage}
	\begin{subfigure}{0.3\linewidth} 
		\includegraphics [width=\linewidth, height =\linewidth ]{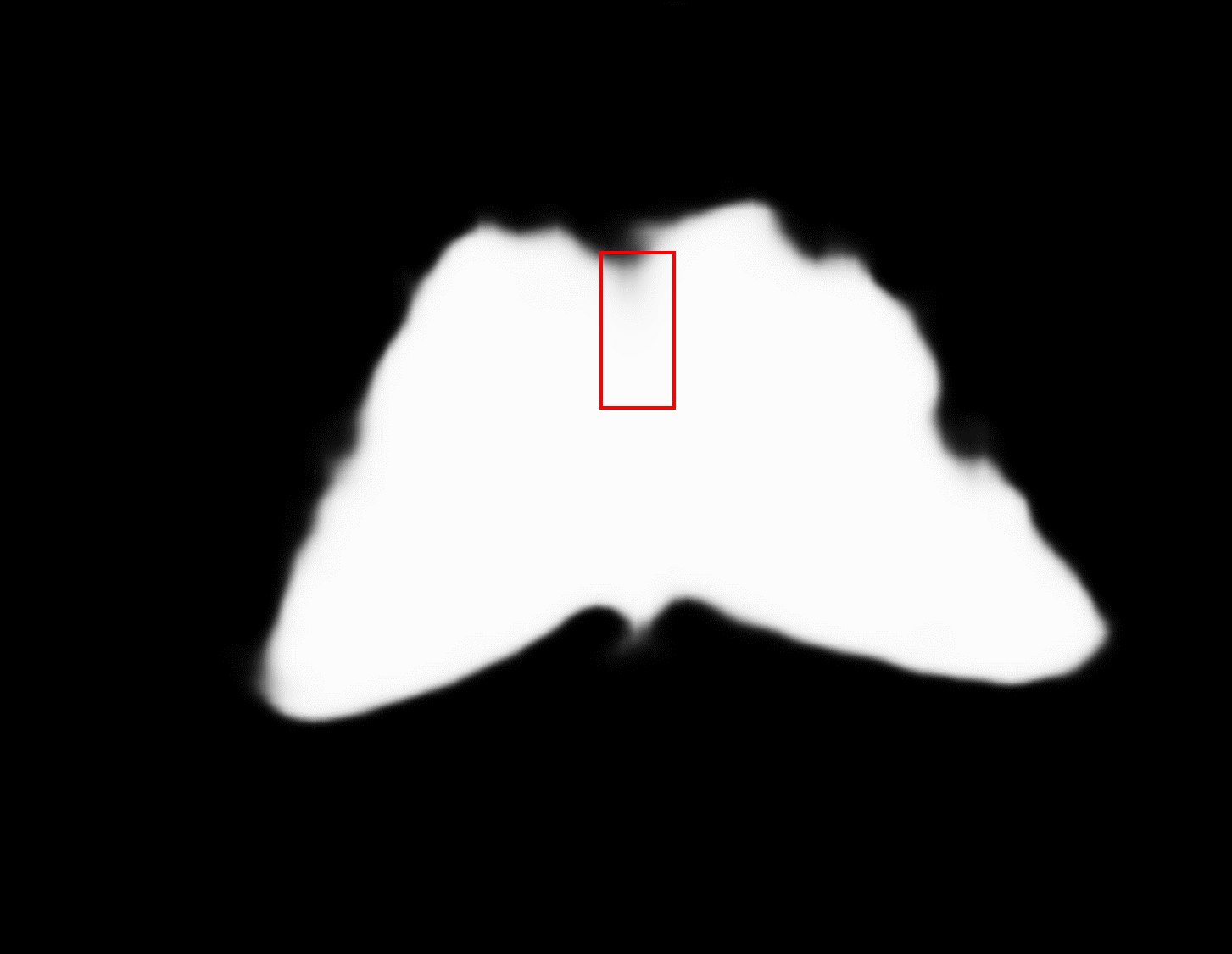}
	\end{subfigure}
	\begin{subfigure}{0.3\linewidth}
		\includegraphics [width=\linewidth, height =\linewidth ]{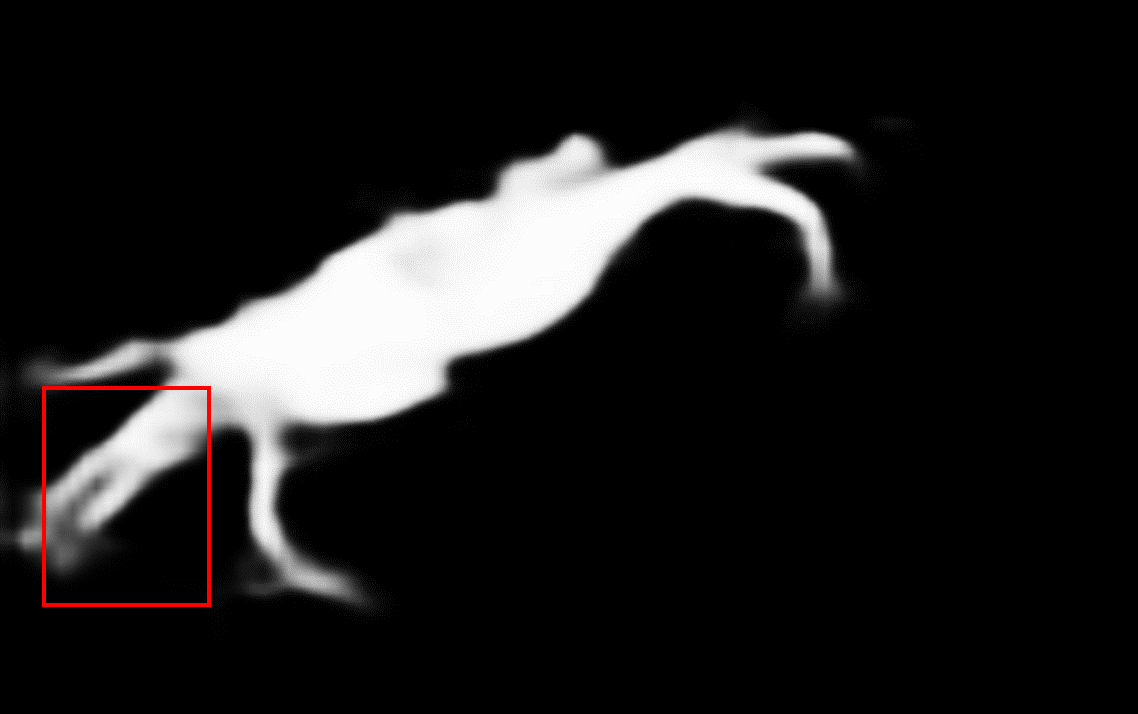}
	\end{subfigure}
	\begin{subfigure}{0.3\linewidth}
		\includegraphics [width=\linewidth, height =\linewidth ]{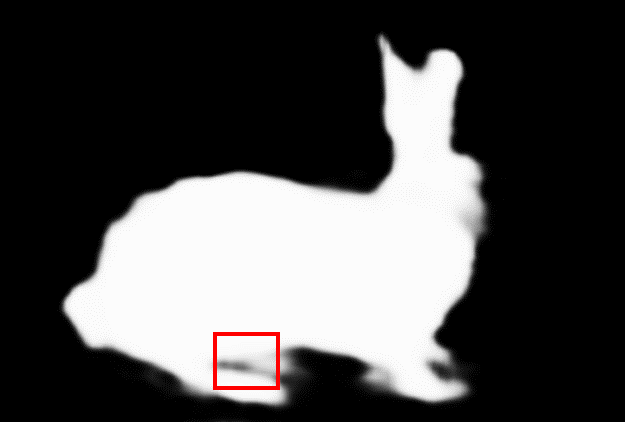}
	\end{subfigure}
	\ \\
	\vspace*{0.5mm}
	\begin{minipage}{0.05\linewidth}
		\begin{sideways}with edge\end{sideways}
	\end{minipage}
	\begin{subfigure}{0.3\linewidth} 
		\includegraphics [width=\linewidth, height =\linewidth ]{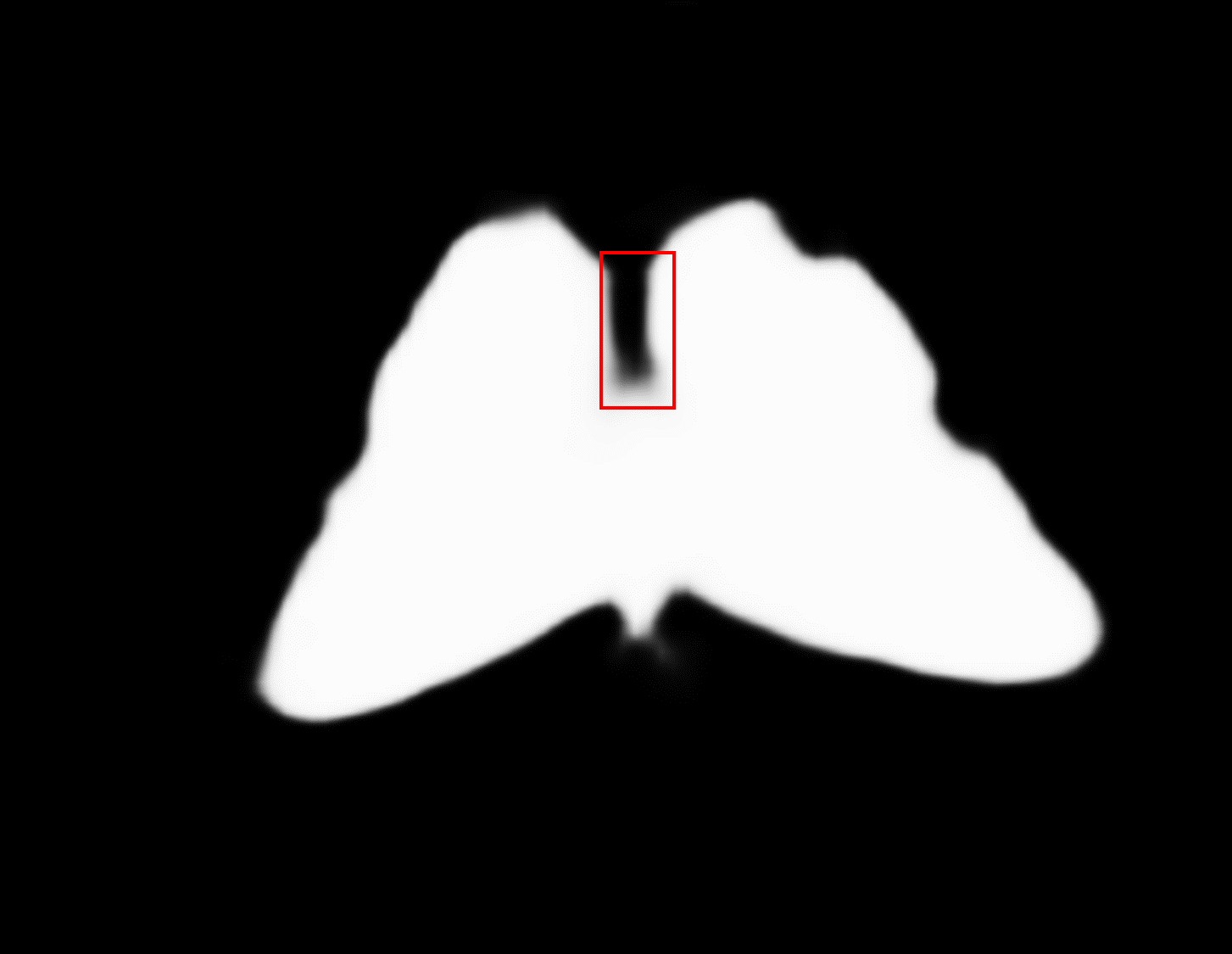}
	\end{subfigure}
	\begin{subfigure}{0.3\linewidth}
		\includegraphics [width=\linewidth, height =\linewidth ]{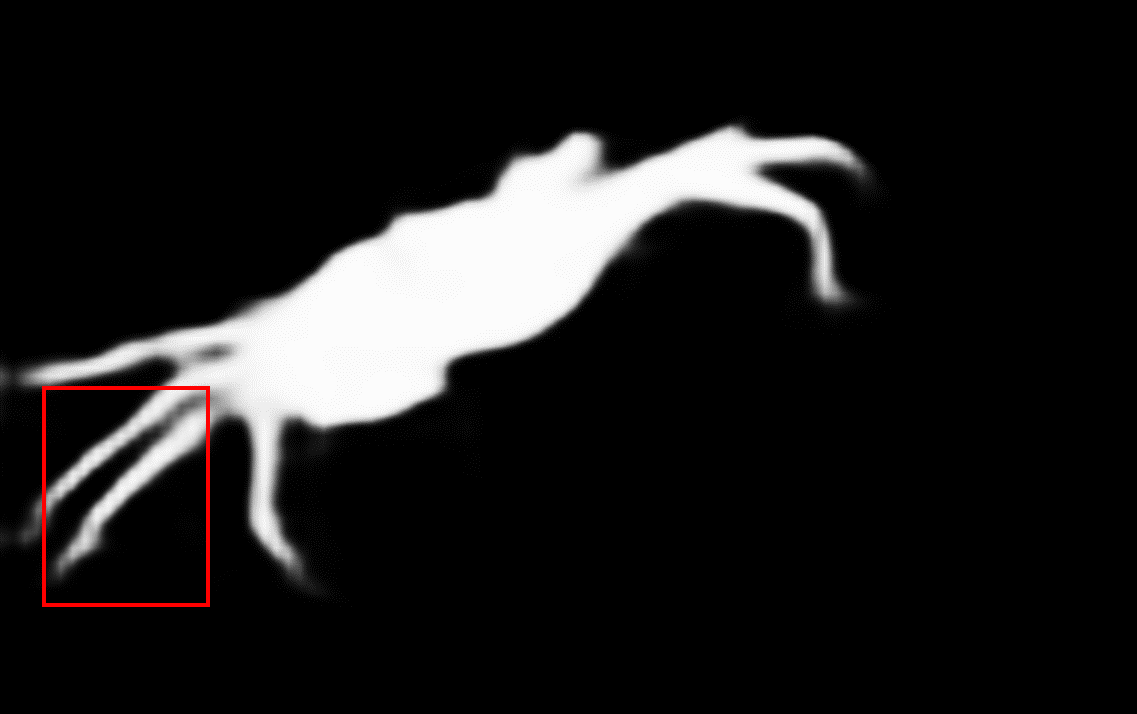}
	\end{subfigure}
	\begin{subfigure}{0.3\linewidth}
		\includegraphics [width=\linewidth, height =\linewidth ]{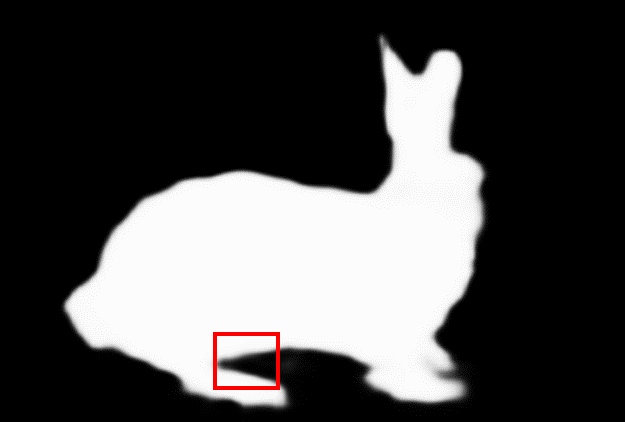}
	\end{subfigure}
	
	\caption{Visual comparison results of our method with and without boundary-consistency loss.}
	\label{fig:edgeref}
\end{figure}

\begin{figure*}[tp]
	\centering
	
	\begin{subfigure}{0.18\textwidth} 
		\includegraphics[width=\textwidth, height=\textwidth]{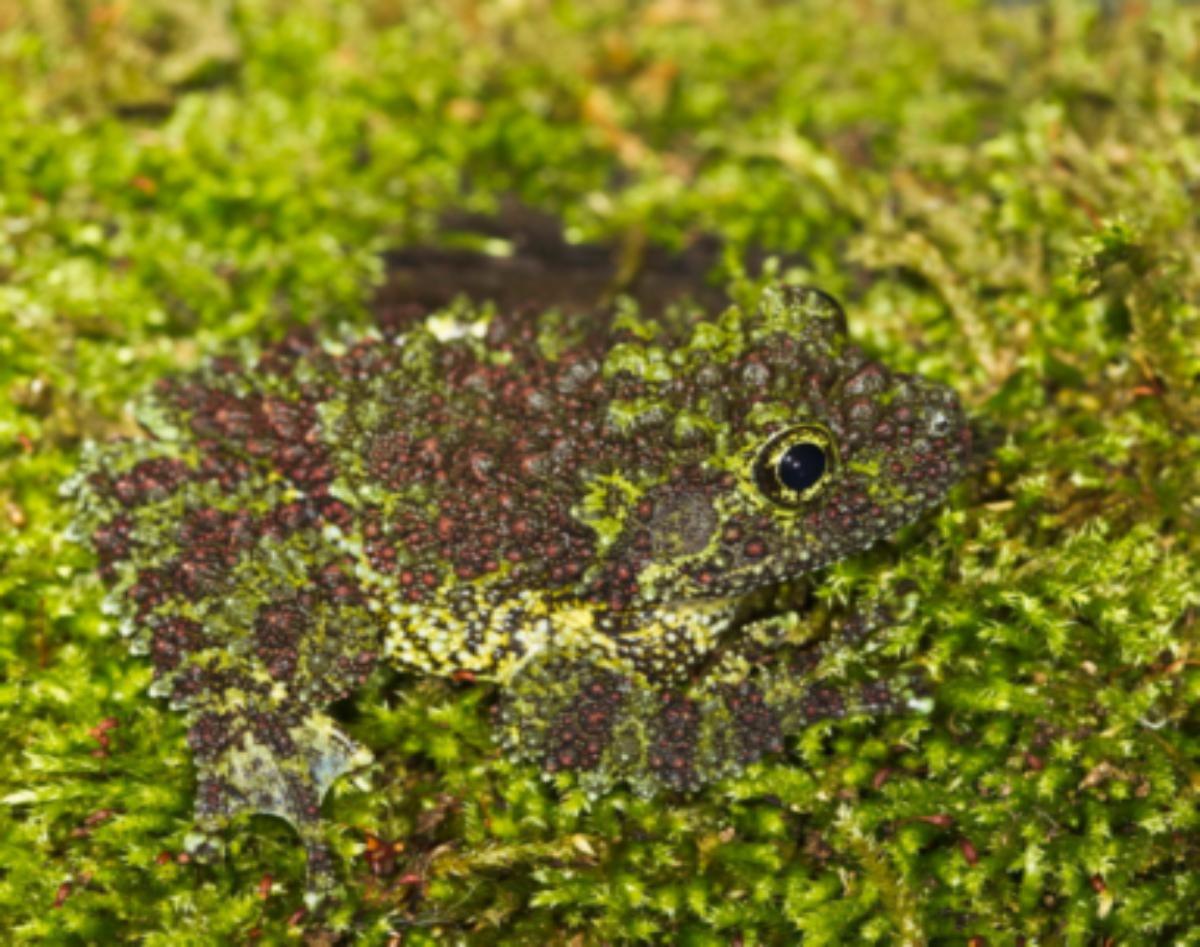}\vspace*{-1mm}
	\end{subfigure}
	\begin{subfigure}{0.18\textwidth}
		\includegraphics[width=\textwidth, height=\textwidth]{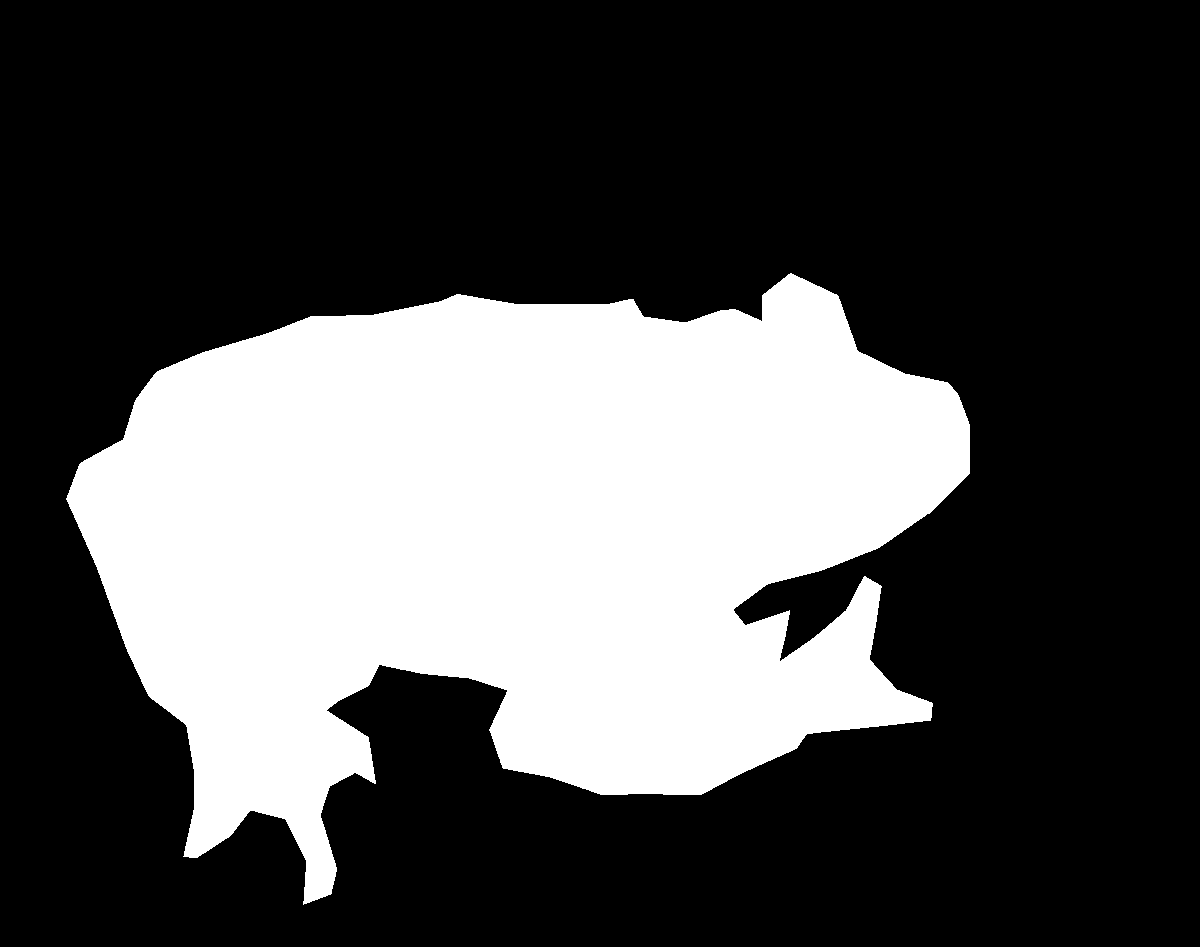}\vspace*{-1mm}
	\end{subfigure}
	\begin{subfigure}{0.18\textwidth}
		\includegraphics[width=\textwidth, height=\textwidth]{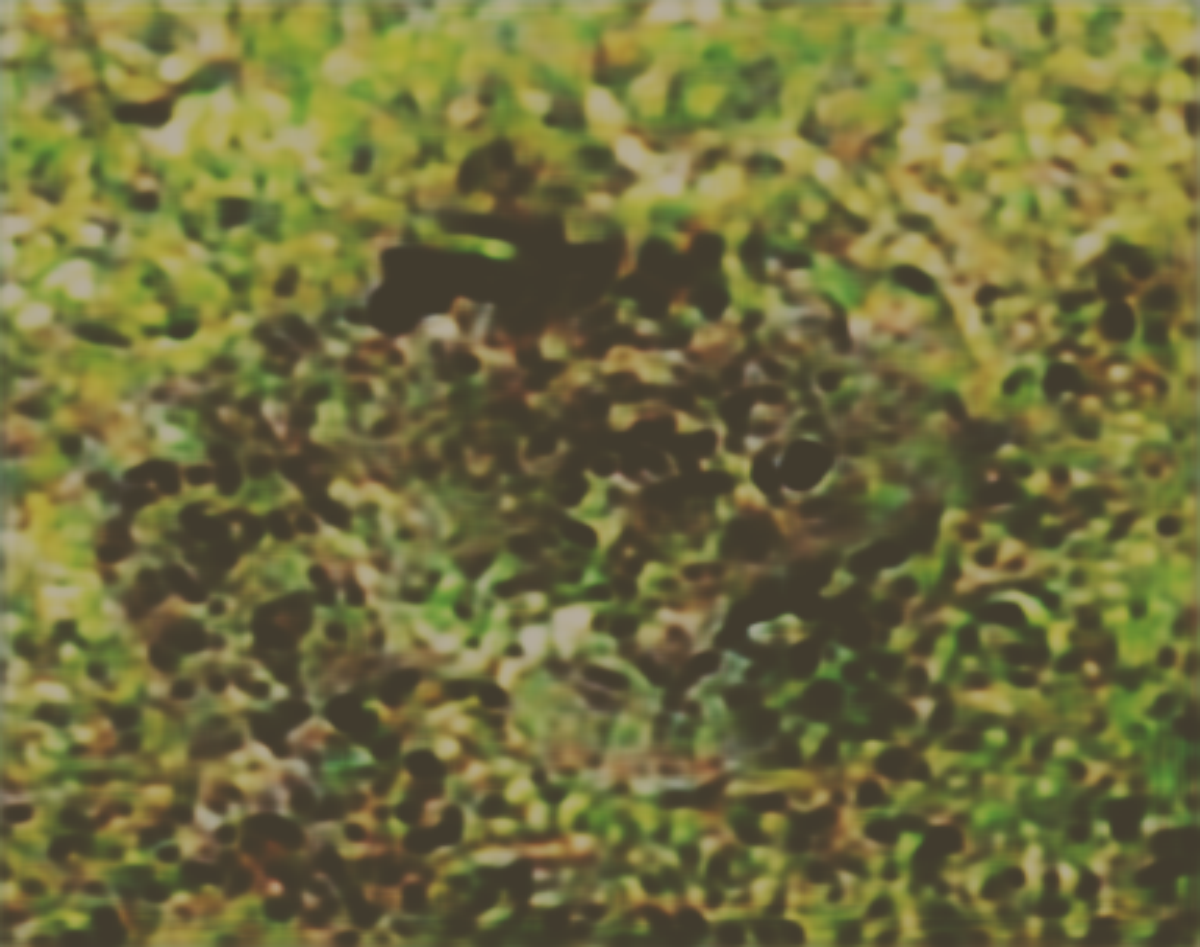}\vspace*{-1mm}
	\end{subfigure}
	\begin{subfigure}{0.18\textwidth}
		\includegraphics[width=\textwidth, height=\textwidth]{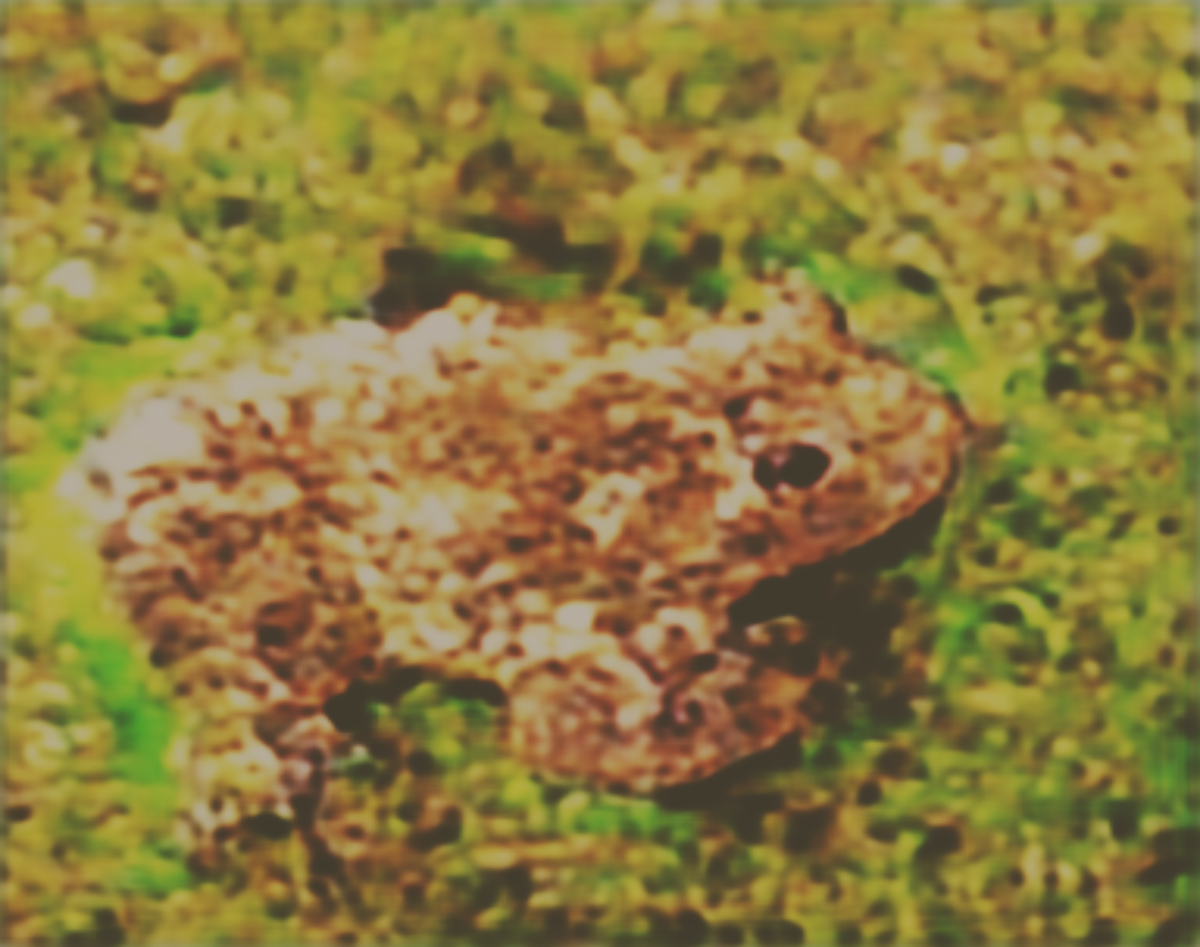}\vspace*{-1mm}
	\end{subfigure}
	\begin{subfigure}{0.18\textwidth}
		\includegraphics[width=\textwidth, height=\textwidth]{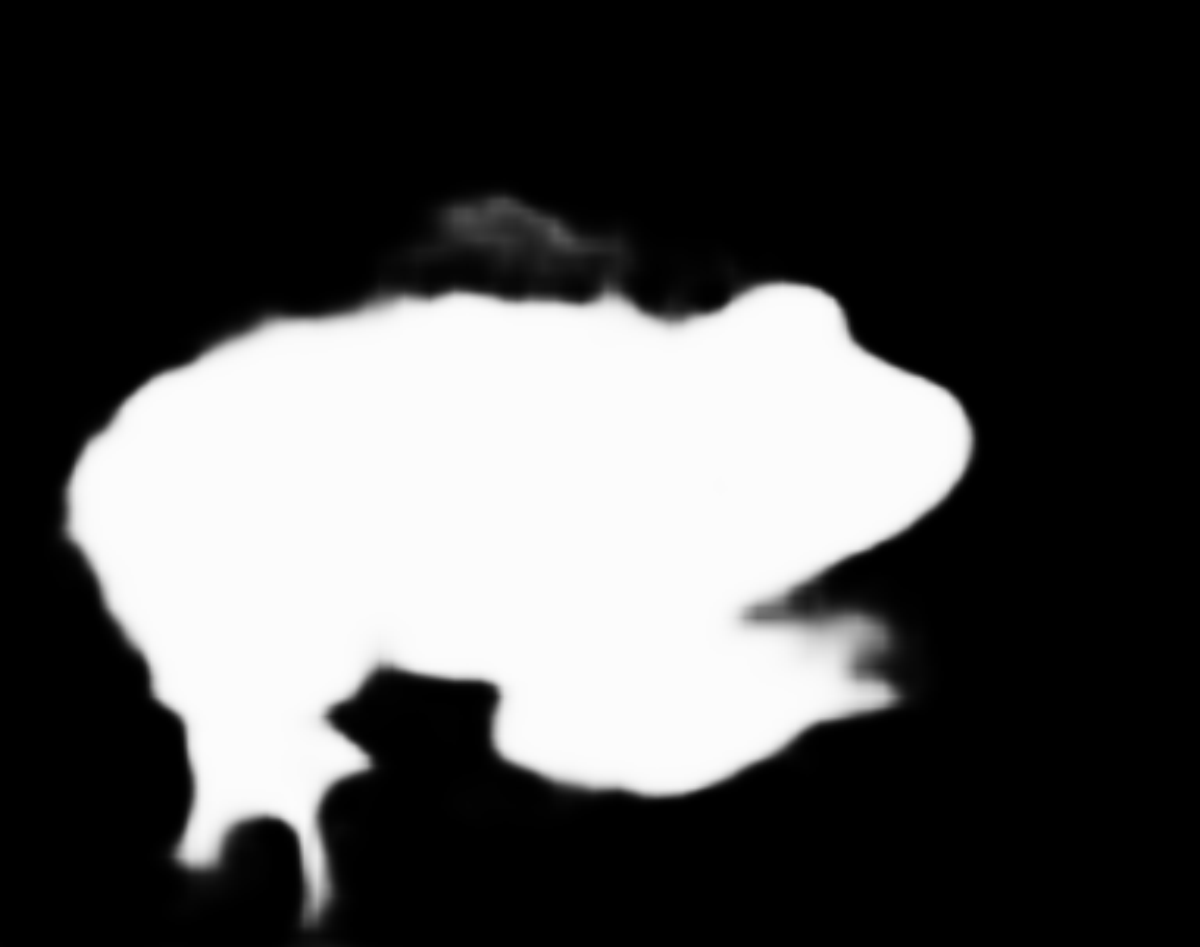}\vspace*{-1mm}
	\end{subfigure}
	
	\ \\
	\vspace*{1mm}
	\begin{subfigure}{0.18\textwidth} 
		\includegraphics[width=\textwidth, height=\textwidth]{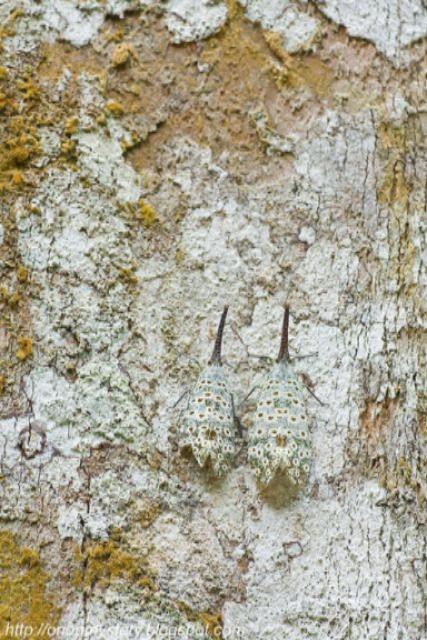}\vspace*{-1mm}
	\end{subfigure}
	\begin{subfigure}{0.18\textwidth}
		\includegraphics[width=\textwidth, height=\textwidth]{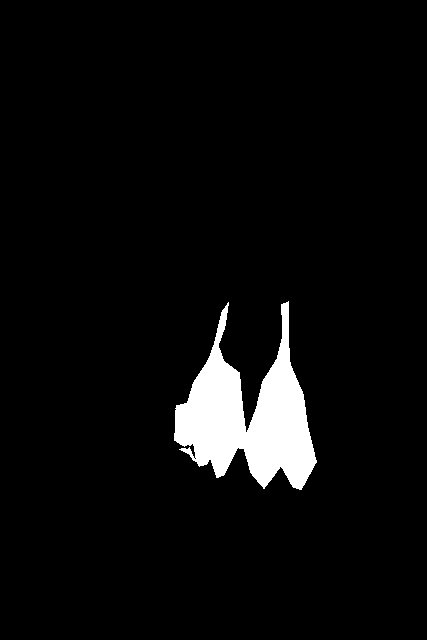}\vspace*{-1mm}
	\end{subfigure}
	\begin{subfigure}{0.18\textwidth}
		\includegraphics[width=\textwidth, height=\textwidth]{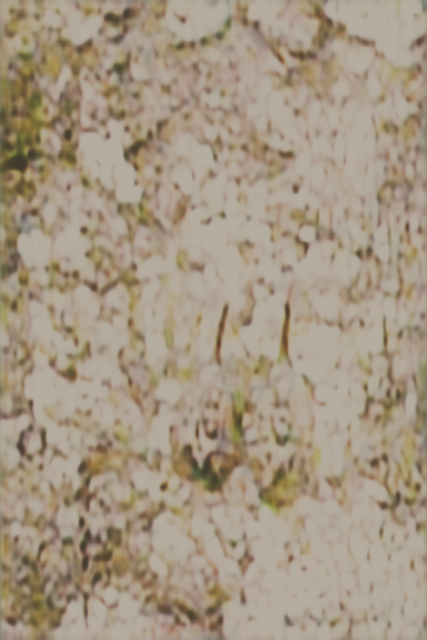}\vspace*{-1mm}
	\end{subfigure}
	\begin{subfigure}{0.18\textwidth}
		\includegraphics[width=\textwidth, height=\textwidth]{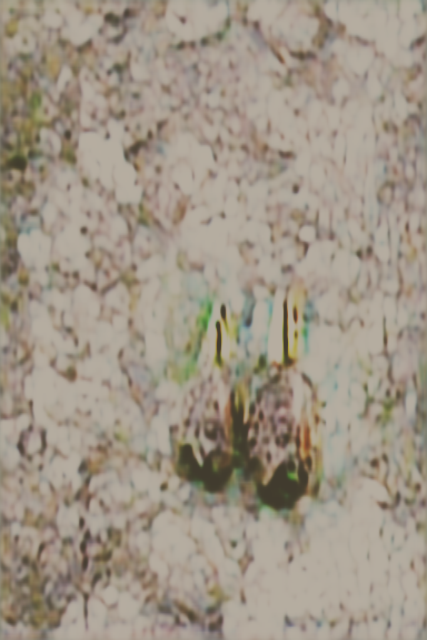}\vspace*{-1mm}
	\end{subfigure}
	\begin{subfigure}{0.18\textwidth}
		\includegraphics[width=\textwidth, height=\textwidth]{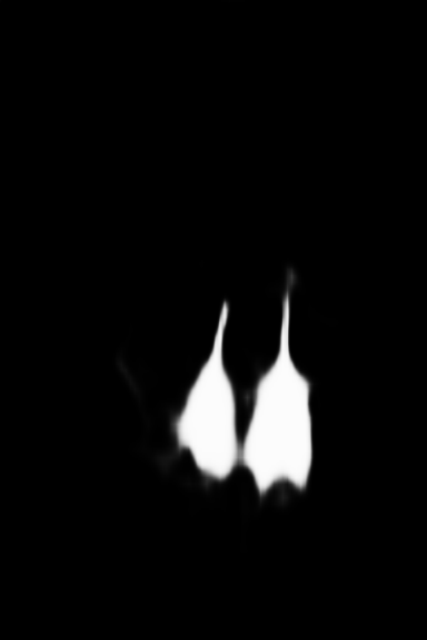}\vspace*{-1mm}
	\end{subfigure}
	
	\ \\
	\vspace*{1mm}
	\begin{subfigure}{0.18\textwidth}
		\includegraphics[width=\textwidth, height=\textwidth]{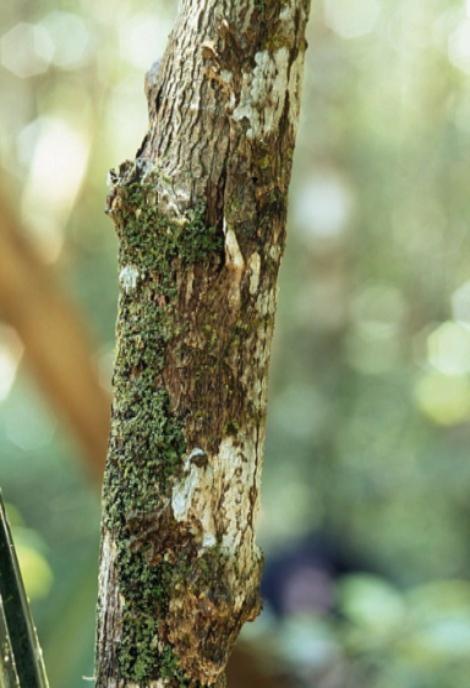}\vspace*{-1mm}\caption{input images}
	\end{subfigure}
	\begin{subfigure}{0.18\textwidth}
		\includegraphics[width=\textwidth, height=\textwidth]{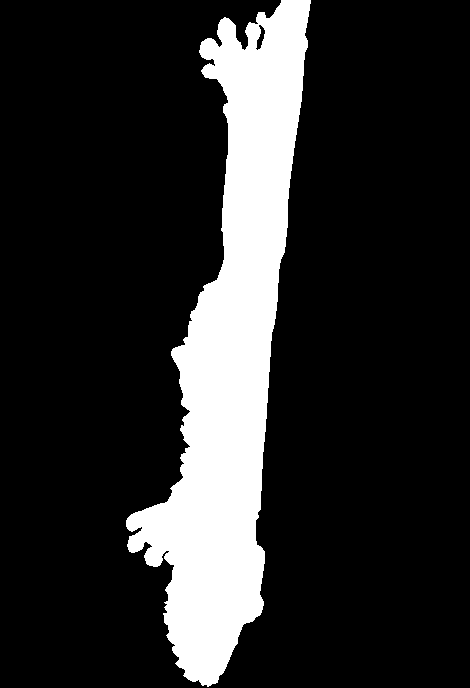}\vspace*{-1mm}\caption{{ ground truths}}
	\end{subfigure}
	\begin{subfigure}{0.18\textwidth}
		\includegraphics[width=\textwidth, height=\textwidth]{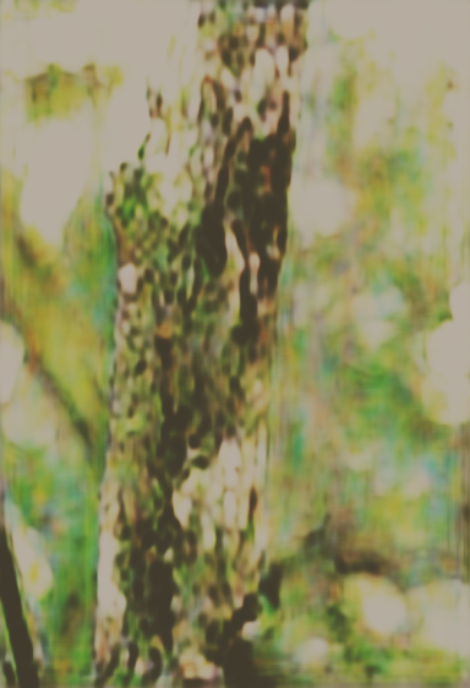}\vspace*{-1mm}\caption{convolutional feature}
	\end{subfigure}
	\begin{subfigure}{0.18\textwidth}
		\includegraphics[width=\textwidth, height=\textwidth]{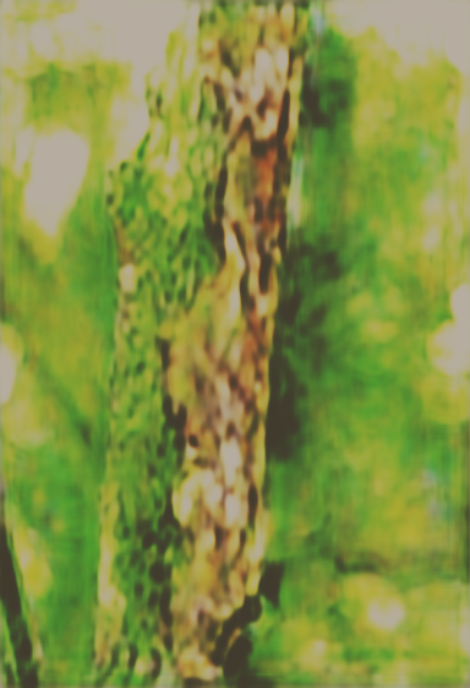}\vspace*{-1mm}\caption{texture-aware feature}
	\end{subfigure}
		\begin{subfigure}{0.18\textwidth}
		\includegraphics[width=\textwidth,height=\textwidth]{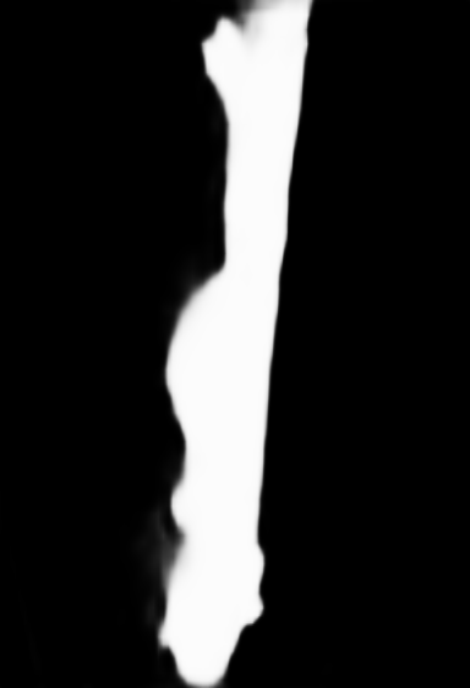}\vspace*{-1mm}\caption{our results}
	\end{subfigure}
	\caption{Visualization results of the reconstructed images from the original convolutional features and texture-aware features produced from our TARM.}

	\label{fig:trmref}
\end{figure*}

\paragraph{Boundary-Consistency Loss.}
The convolutional features contain highly semantic features but tend to produce blurry boundaries between the camouflaged objects and the background due to the small resolutions of the parameter maps.
To solve this issue, we present a boundary-consistency loss $L_{edge}$ to improve the boundary quality by revisiting the prediction results across boundary regions:
\begin{equation}
L_{edge} \ = \ \sum_{\forall i, b_i \in \mathcal{B}}{ \sum_{m, n \in b_i} d(A^{h}_{m,n}, A^{gt}_{m,n})} \ ,
\end{equation}
where $b_i$ is the $i$-th image patch in $\mathcal{B}$, which contains all the image patches that across boundary, as shown in Figure~\ref{fig:edge}.
These image patches are selected when they contain the pixels that belong to different categories.
%
Note that unlike affinity loss, the $L_{edge}$ is performed on parameter maps without downsampling operations and the parameter maps with higher resolutions help to provide more detailed information for boundary prediction.
The additional memory consumption during training is reasonable since we only consider affinity relationship within small patches and no extra computational time is introduced in testing process.
The comparison results in Figure~\ref{fig:edgeref} show that our method with the boundary-consistency loss better preserves the detailed structures of camouflaged objects.

\subsection{Training and Testing Strategies}
\label{subsec:train_test}
The overall loss function was defined as 
\begin{equation}
L  \ = \ {{\lambda}_0L_{seg}+\lambda}_1L_{aff} \ + \ {\lambda}_2L_{edge} \ ,
\end{equation}
where $L_{seg}$ was the binary cross entropy loss used for camouflaged segmentation. 
We empirically set $\lambda_0 = \lambda_1 = 1$ and $\lambda_2 = 10$ to balance the numerical magnitude.
%
%
We implemented our network in Pytorch and used ResNeXt50~\cite{xie2017aggregated} as the backbone network, which was pre-trained on ImageNet~\cite{russakovsky2015imagenet}.
We used stochastic gradient descent (SGD) with the poly learning strategy \cite{liu2015parsenet} to optimize the network by setting the initial learning rate as $0.001$ and decay power as $0.9$. 
The training process was stopped after $30$ epochs.
It took about two hours to train the network on a 1080Ti GPU with the batch size of $16$. 
The input images were resized as $384 \times 384$.
In testing, we adopted the prediction mask with the highest resolution as the final result. 
We used 0.03s to process an image with the size of $384 \times 384$.

\subsection{Visualization}
\label{subsec:visualization}
To visualize the learned texture-aware features, we adopt a decoder~\cite{kim2019u} to reconstruct the images using the feature maps before and after refined by texture-aware refinement module (TARM); see Section~\ref{subsec:TARM}. 
Note that during the visualization process, the weights in our TANet are fixed and we only trained the newly added decoder to reconstruct the original images.   
Figures~\ref{fig:fish}\&\ref{fig:trmref} show the visualization results, where after adopting our TARM module to refine features, the texture differences between camouflaged objects and background are clearly amplified, which helps to improve the overall performance of camouflaged object detection.



\section{Experimental Results}
\label{sec:experiments}


\subsection{Benchmark Protocols}
\begin{table*}[htbp]
	\begin{center}
        \caption{Comparison with state-of-the-art methods for camouflaged object detection on three benchmark datasets.}
		\label{table:state-of-the-art_SD}
		\resizebox{0.96\textwidth}{!}{%
			\begin{tabular}{c|c|c c c c|c c c c|c c c c}
			    \hline
			    \hline
				\multirow{2}{*}{Method} & \multirow{2}{*}{Year} &
				\multicolumn{4}{c|}{CHAMELEON} &  \multicolumn{4}{c|}{CAMO-Test} & \multicolumn{4}{c}{COD10K-Test} \\
				\cline{3-14}
				& & {$S_\alpha \uparrow$} & {$E_\phi \uparrow$} &{$F_{\beta}^w \uparrow$} & {M $\downarrow$} & 	{$S_\alpha \uparrow$} & {$E_\phi \uparrow$}
				 &{$F_{\beta}^w \uparrow$} & {M $\downarrow$} & {$S_\alpha \uparrow$} & {$E_\phi \uparrow$} &{$F_{\beta}^w \uparrow$} & {M $\downarrow$}\\
				 \cline{3-14}
				\hline
				 FPN \cite{lin2017feature} & 2017  & 0.794 & 0.783 & 0.590& 0.075 & 0.684 & 0.677 & 0.483 & 0.131 & 0.697 & 0.691 & 0.411 & 0.075 \\
				\hline
				 MaskRCNN \cite{he2017mask}  & 2017 & 0.643 & 0.778 & 0.518& 0.099 & 0.574 & 0.715 & 0.430 & 0.151 & 0.613 & 0.748 & 0.402 & 0.080 \\
				\hline
				PSPNet \cite{zhao2017pyramid} & 2017  & 0.773 & 0.758 & 0.555& 0.085 & 0.663 & 0.659 & 0.455 & 0.139 & 0.678 & 0.680 & 0.377 & 0.080 \\
				\hline
				 UNet++ \cite{zhou2018unet++} & 2018  & 0.695 & 0.762 & 0.501& 0.094 & 0.599 & 0.653 & 0.392 & 0.149 & 0.623 & 0.672 & 0.350 & 0.086 \\
				\hline
				PiCANet \cite{liu2018picanet} & 2018  & 0.769 & 0.749 & 0.536& 0.085 & 0.609 & 0.584 & 0.356 & 0.156 & 0.649 & 0.643 & 0.322 & 0.090 \\
				\hline
				MSRCNN \cite{huang2019mask} & 2019  & 0.637 & 0.686 & 0.443& 0.091 & 0.617 & 0.669 & 0.454 & 0.133 & 0.641 & 0.706 & 0.419 & 0.073 \\
				\hline
				BASNet \cite{qin2019basnet} & 2019  & 0.687 & 0.721 & 0.474& 0.118 & 0.618 & 0.661 & 0.413 & 0.159 & 0.634 & 0.678 & 0.365 & 0.105 \\
				\hline
				PFANet \cite{zhao2019pyramid} & 2019  & 0.679 & 0.648 & 0.378& 0.144 & 0.659 & 0.622 & 0.391 & 0.172 & 0.636 & 0.618 & 0.286 & 0.128 \\
				\hline
				 CPD \cite{wu2019cascaded} & 2019  & 0.853 & 0.866 & 0.706& 0.052 & 0.726 & 0.729 & 0.550 & 0.115 & 0.747 & 0.770 & 0.508 & 0.059 \\
				\hline
				 HTC \cite{chen2019hybrid} & 2019 & 0.517 & 0.489 & 0.204& 0.129 & 0.476 & 0.442 & 0.174 & 0.172 & 0.548 & 0.520 & 0.221 & 0.088 \\
				\hline
				EGNet \cite{zhao2019egnet} & 2019 & 0.848 & 0.870 & 0.702& 0.050 & 0.732 & 0.768 & 0.583 & 0.104 & 0.737 & 0.779 & 0.509 & 0.056 \\ %
				\hline
			    ANet-SRM \cite{le2019anabranch} & 2019 & $\ddagger$ &$\ddagger$ & $\ddagger$ & $\ddagger$ & 0.682 & 0.685 & 0.484 & 0.126 &  $\ddagger$ &$\ddagger$ & $\ddagger$ & $\ddagger$   \\ %
				\hline
				SINet \cite{fan2020camouflaged} & 2020 & 0.869 & 0.891 & 0.740& 0.044 & 0.751 & 0.771 & 0.606 & 0.100 & 0.771 & 0.806 & 0.551 & 0.051 \\
				\hline
				TANet\_v1 & - & 0.881 & 0.907 & 0.773& 0.039 & 0.778 & 0.813 & 0.659 & 0.089 & 0.794 & 0.838 & 0.613 & 0.043 \\
				\hline
				{TANet (ours)} & - &  \textbf{0.888} & \textbf{0.911} & \textbf{0.786} & \textbf{0.036} & \textbf{0.793}  & \textbf{0.834} & \textbf{0.690} & \textbf{0.083} & \textbf{0.803} & \textbf{0.848} & \textbf{0.629} & \textbf{0.041}\\
				\hline
				\hline
		\end{tabular}
        }
	    \vspace{1mm}
		
	\end{center}
    \vspace{-8mm}
\end{table*}

\begin{figure*}[t]
	\centering
	\vspace*{0.5mm}
	\begin{subfigure}{0.105\textwidth}
		\includegraphics[width=\textwidth]{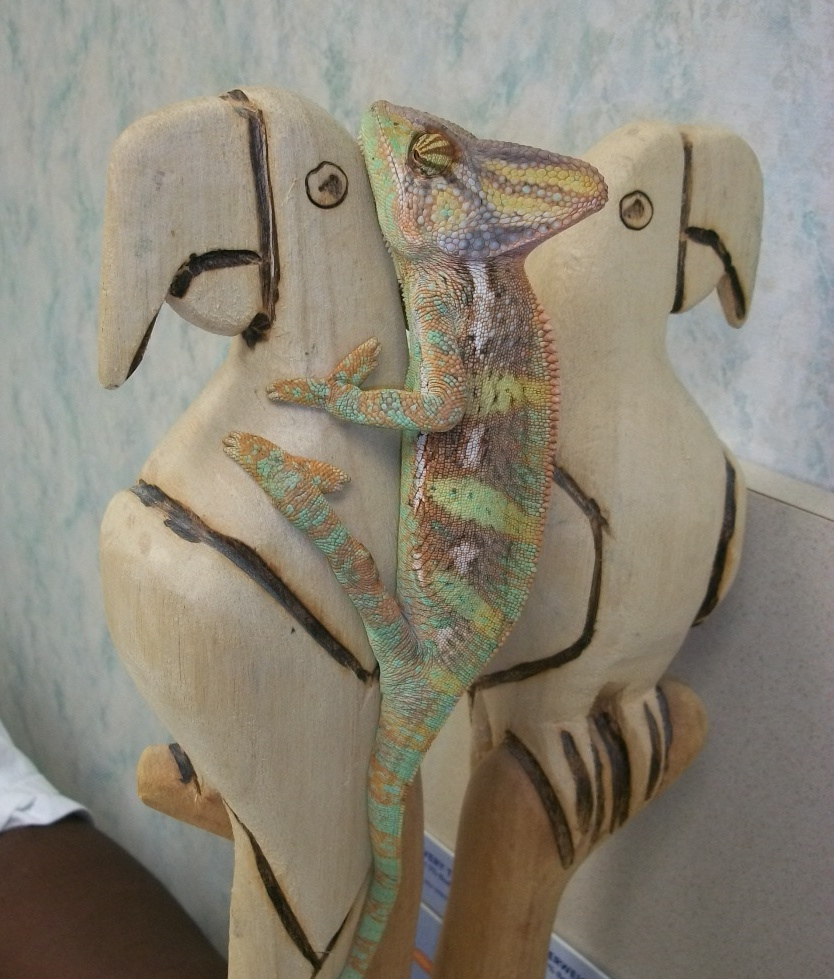}
	\end{subfigure}
	\begin{subfigure}{0.105\textwidth}
		\includegraphics[width=\textwidth]{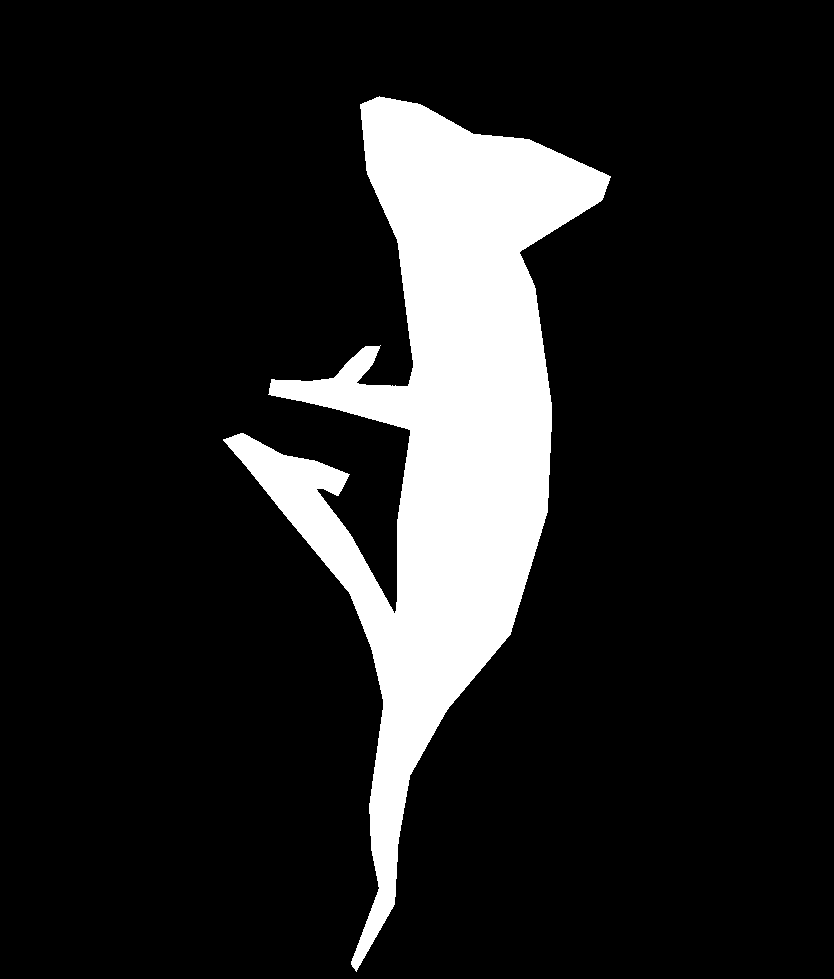}
	\end{subfigure}
	\begin{subfigure}{0.105\textwidth}
		\includegraphics[width=\textwidth]{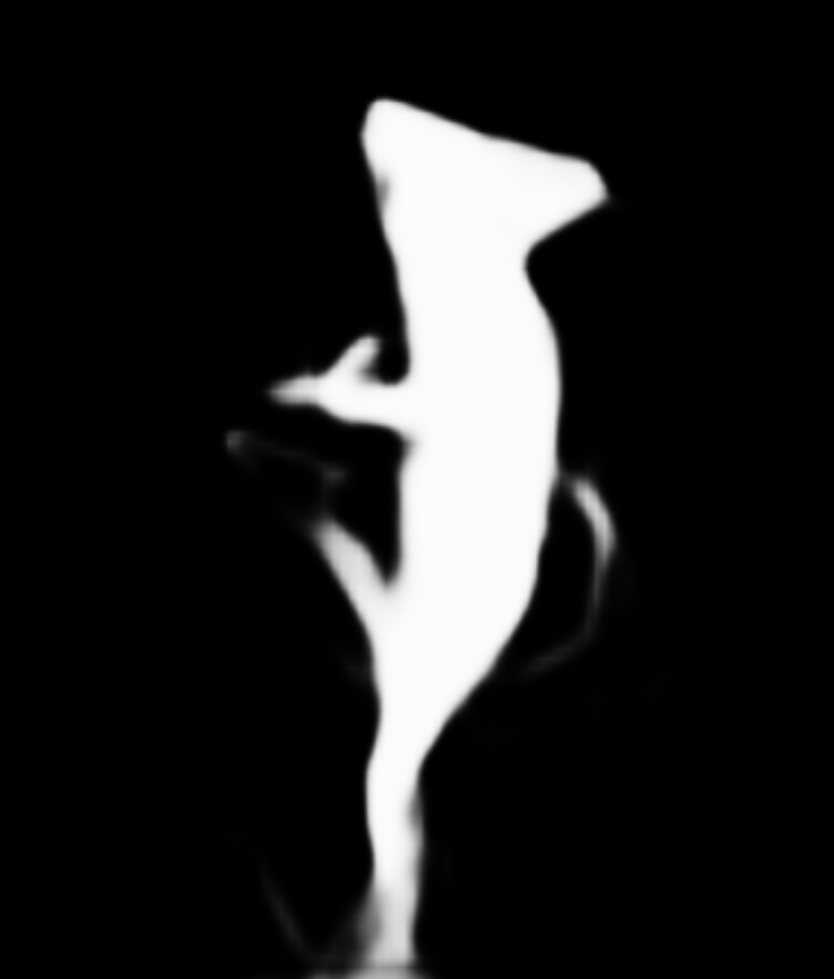}
	\end{subfigure}
    \begin{subfigure}{0.105\textwidth}
		\includegraphics[width=\textwidth]{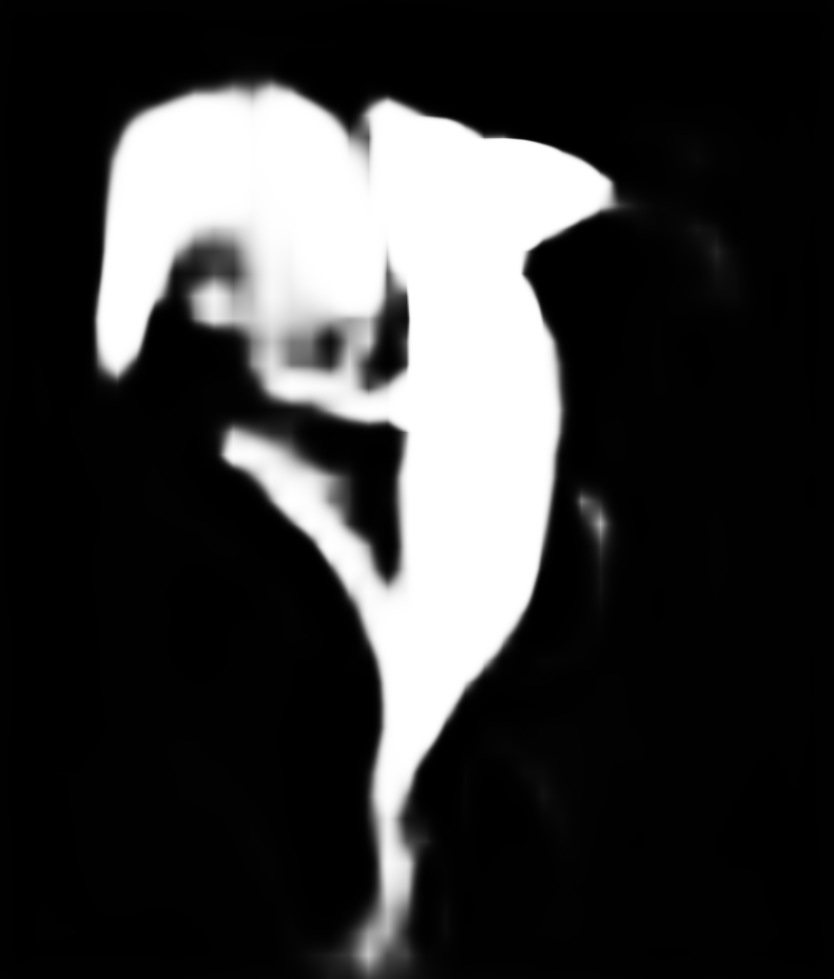}
	\end{subfigure}
    \begin{subfigure}{0.105\textwidth}
		\includegraphics[width=\textwidth]{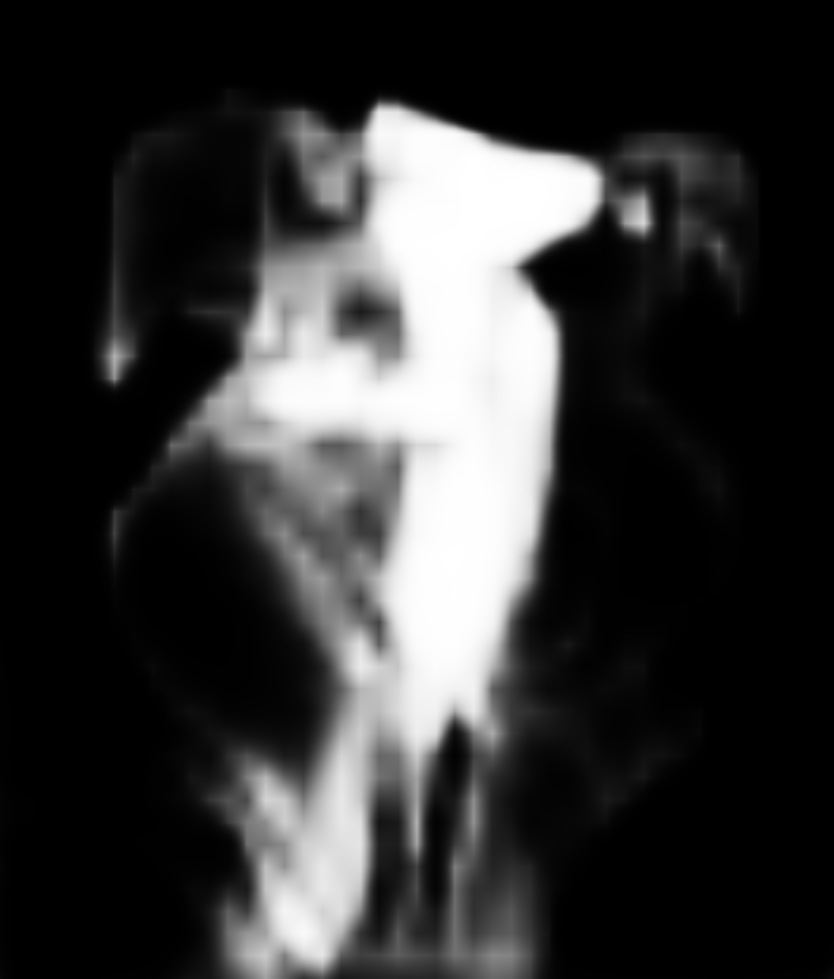}
	\end{subfigure}
    \begin{subfigure}{0.105\textwidth}
		\includegraphics[width=\textwidth]{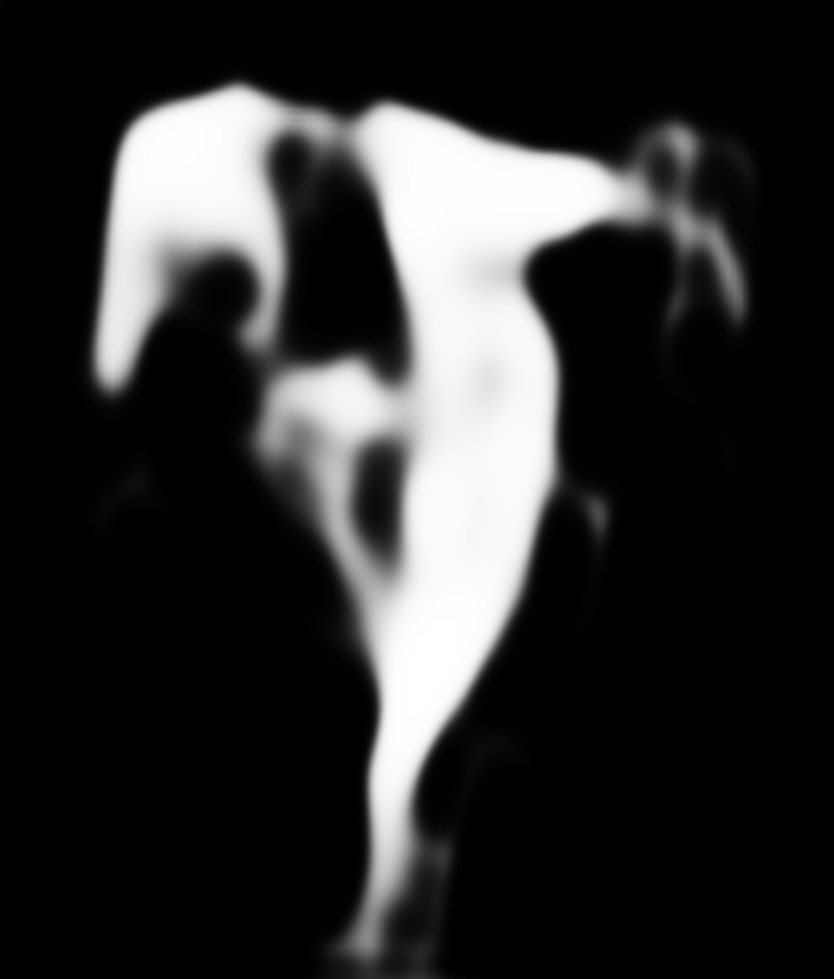}
	\end{subfigure}
	\begin{subfigure}{0.105\textwidth}
		\includegraphics[width=\textwidth]{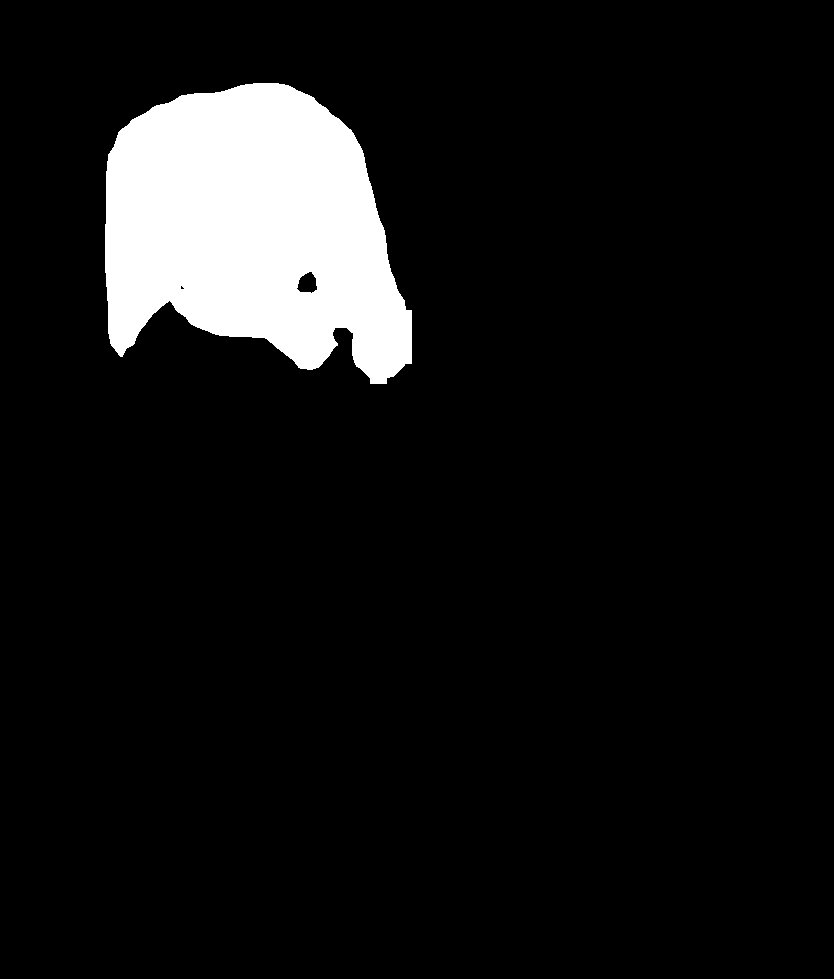}
	\end{subfigure}
	\begin{subfigure}{0.105\textwidth}
		\includegraphics[width=\textwidth]{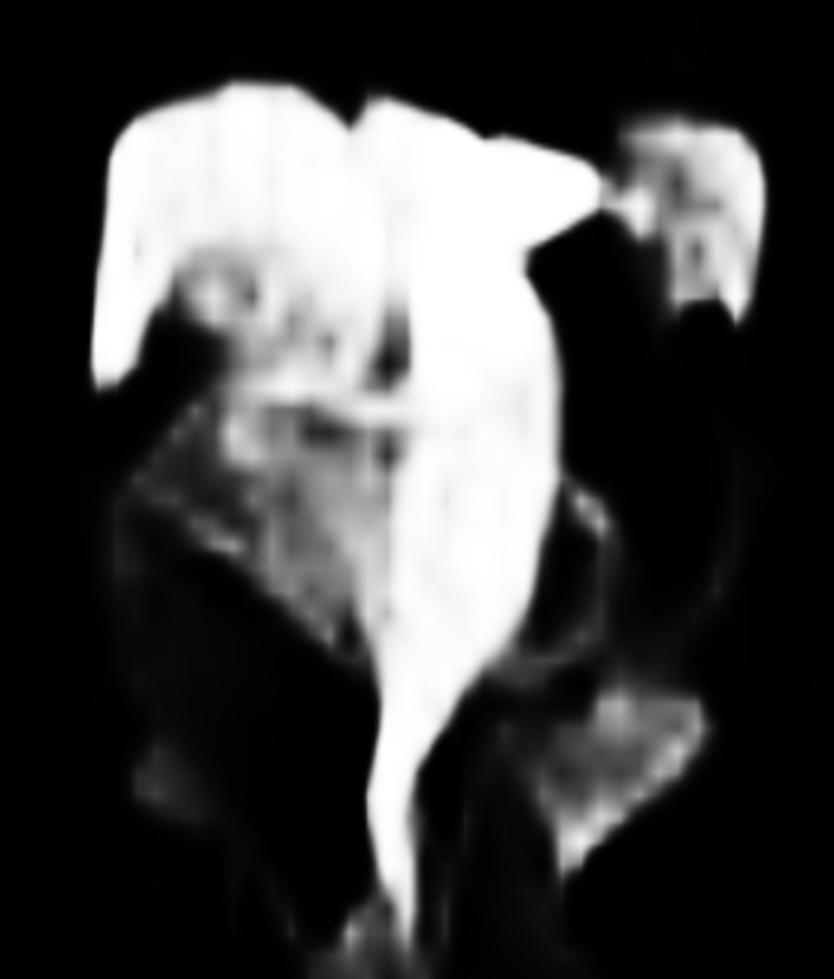}
	\end{subfigure}
    \begin{subfigure}{0.105\textwidth}
		\includegraphics[width=\textwidth]{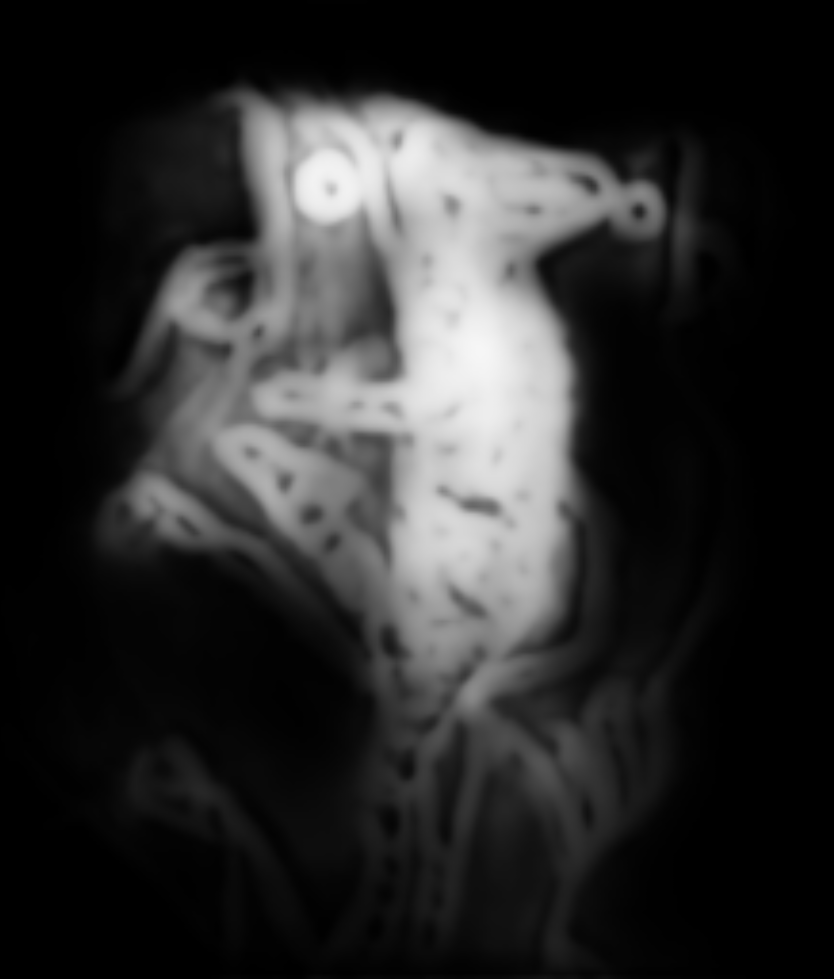}
	\end{subfigure}
	\ \\
	\vspace*{0.5mm}
	\begin{subfigure}{0.105\textwidth}
		\includegraphics[width=\textwidth]{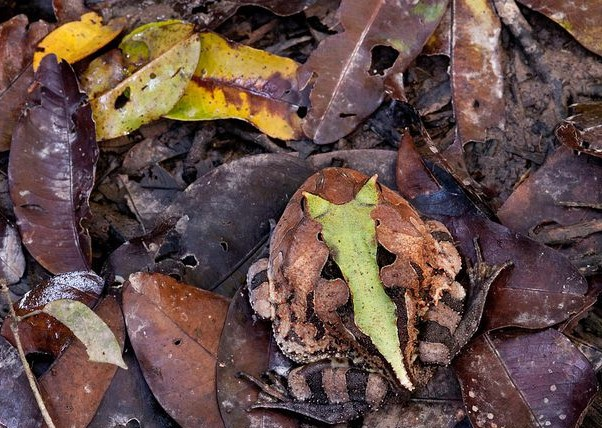}
	\end{subfigure}
	\begin{subfigure}{0.105\textwidth}
		\includegraphics[width=\textwidth]{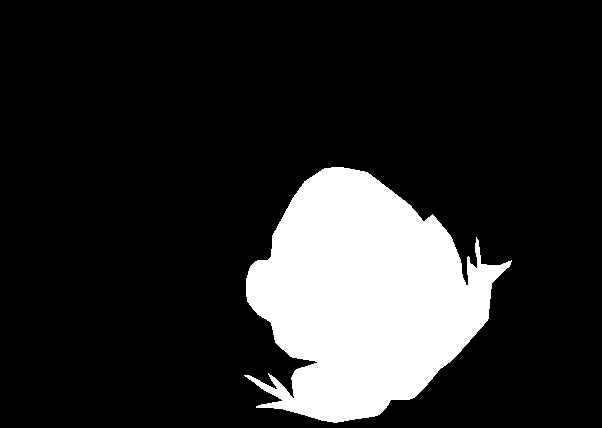}
	\end{subfigure}
	\begin{subfigure}{0.105\textwidth}
		\includegraphics[width=\textwidth]{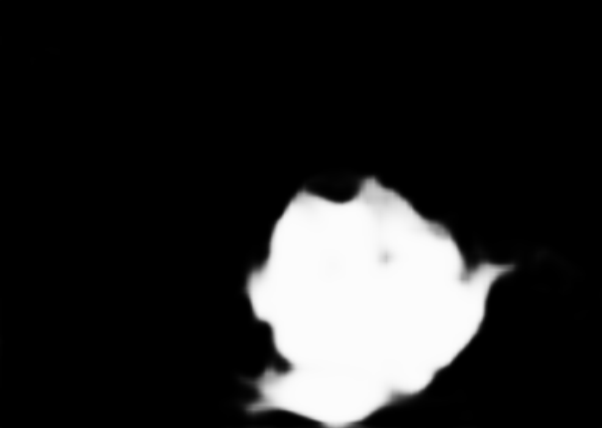}
	\end{subfigure}
    \begin{subfigure}{0.105\textwidth}
		\includegraphics[width=\textwidth]{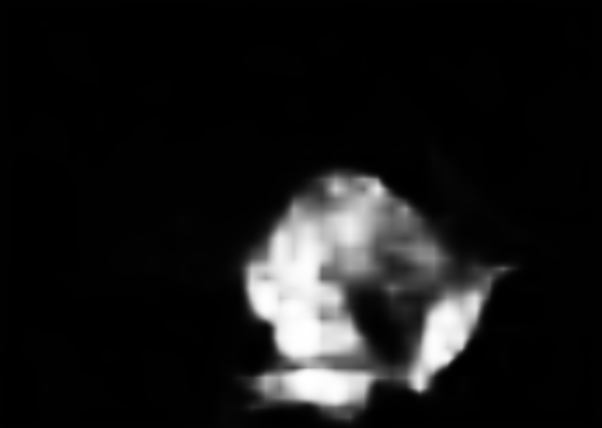}
	\end{subfigure}
    \begin{subfigure}{0.105\textwidth}
		\includegraphics[width=\textwidth]{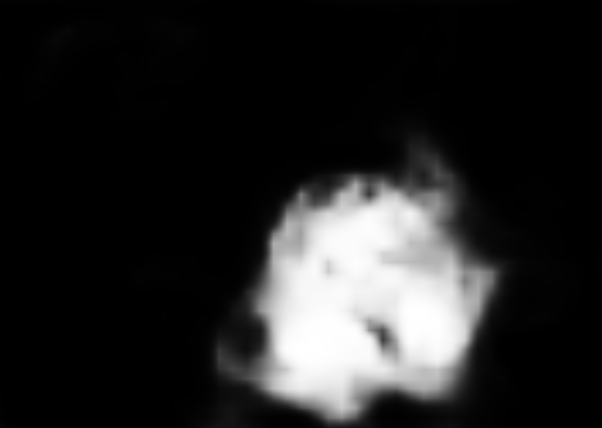}
	\end{subfigure}
    \begin{subfigure}{0.105\textwidth}
		\includegraphics[width=\textwidth]{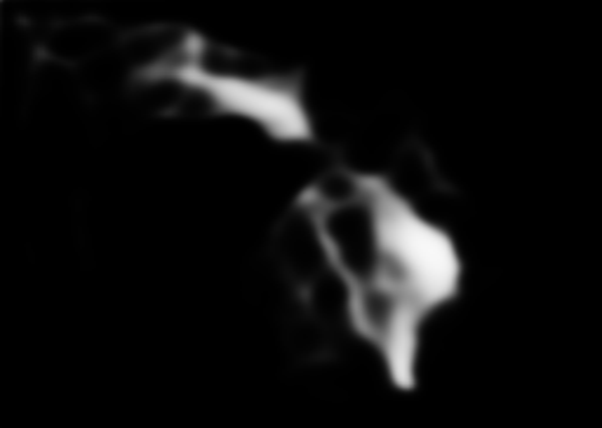}
	\end{subfigure}
	\begin{subfigure}{0.105\textwidth}
		\includegraphics[width=\textwidth]{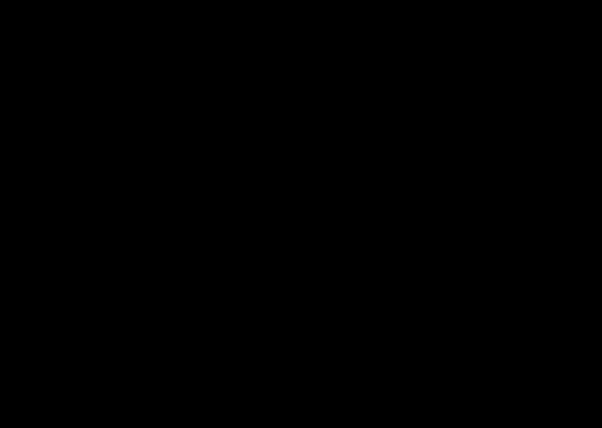}
	\end{subfigure}
	\begin{subfigure}{0.105\textwidth}
		\includegraphics[width=\textwidth]{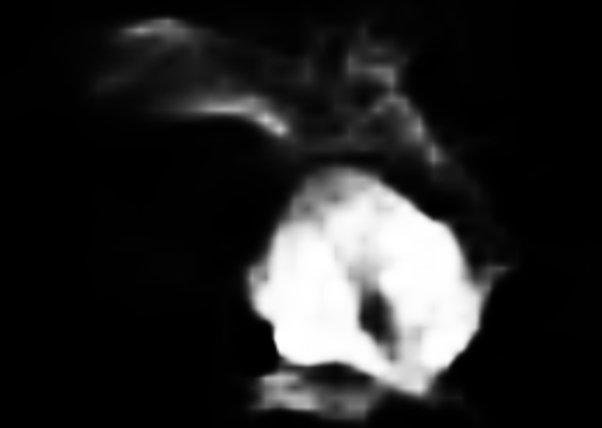}
	\end{subfigure}
    \begin{subfigure}{0.105\textwidth}
		\includegraphics[width=\textwidth]{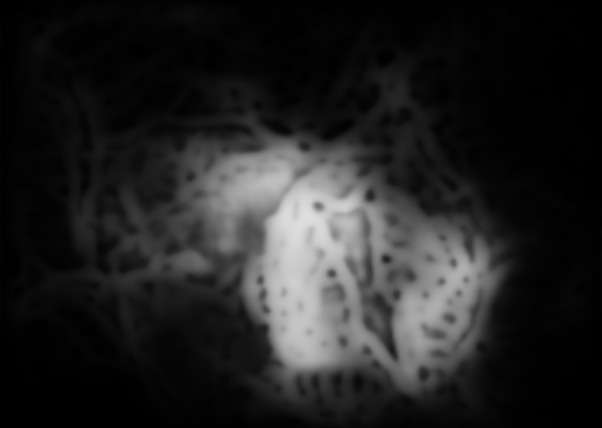}
	\end{subfigure}
	\ \\
	\vspace*{0.5mm}
	\begin{subfigure}{0.105\textwidth}
		\includegraphics[width=\textwidth]{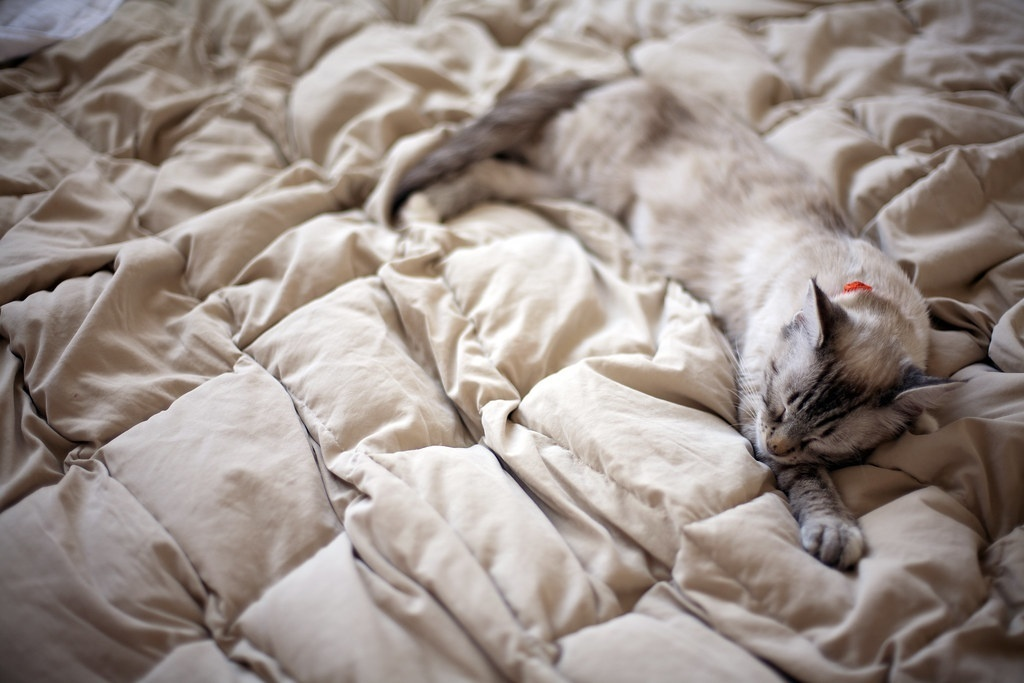}
	\end{subfigure}
	\begin{subfigure}{0.105\textwidth}
		\includegraphics[width=\textwidth]{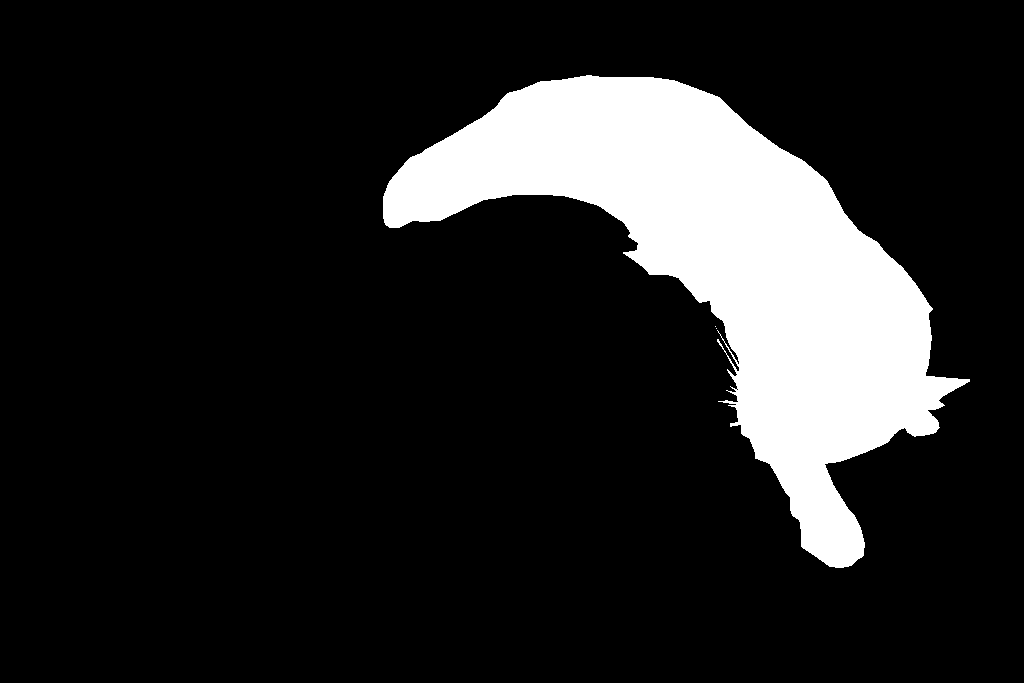}
	\end{subfigure}
	\begin{subfigure}{0.105\textwidth}
		\includegraphics[width=\textwidth]{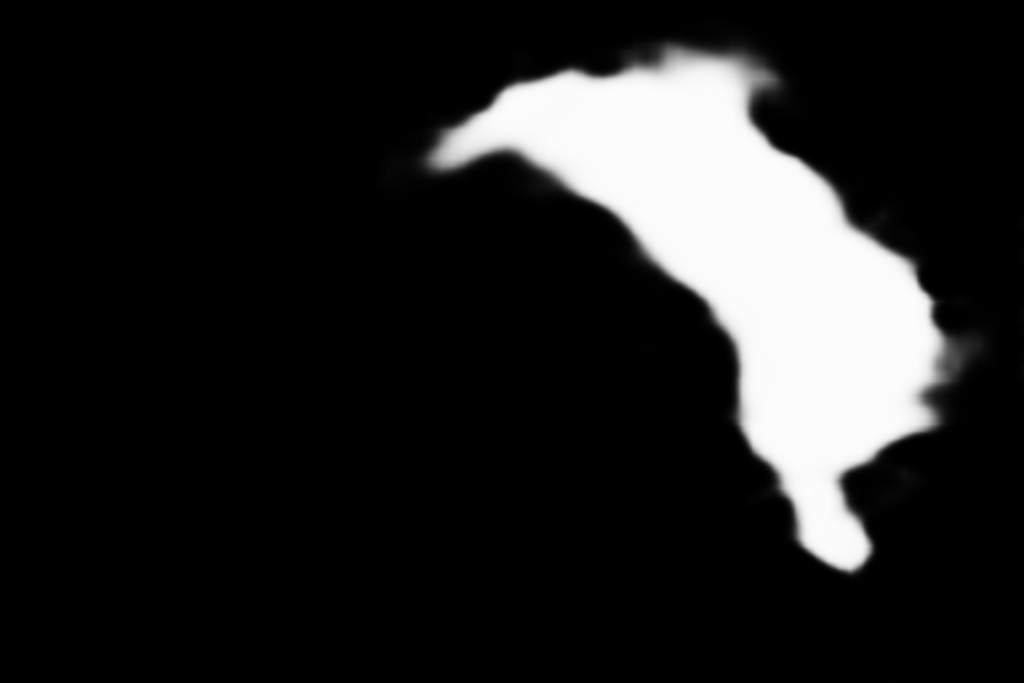}
	\end{subfigure}
    \begin{subfigure}{0.105\textwidth}
		\includegraphics[width=\textwidth]{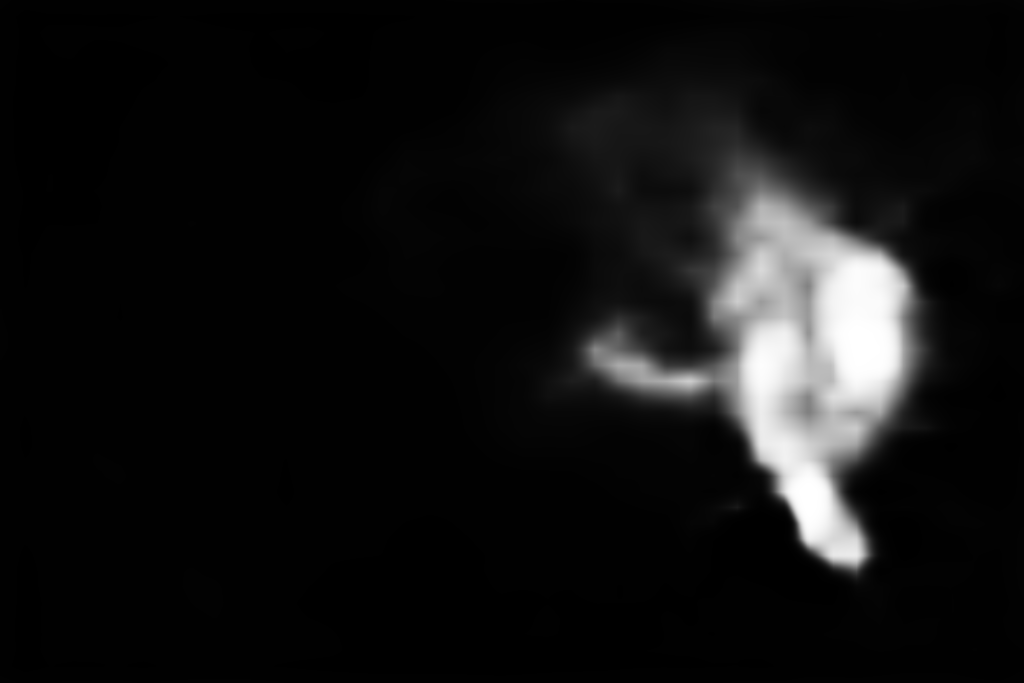}
	\end{subfigure}
    \begin{subfigure}{0.105\textwidth}
		\includegraphics[width=\textwidth]{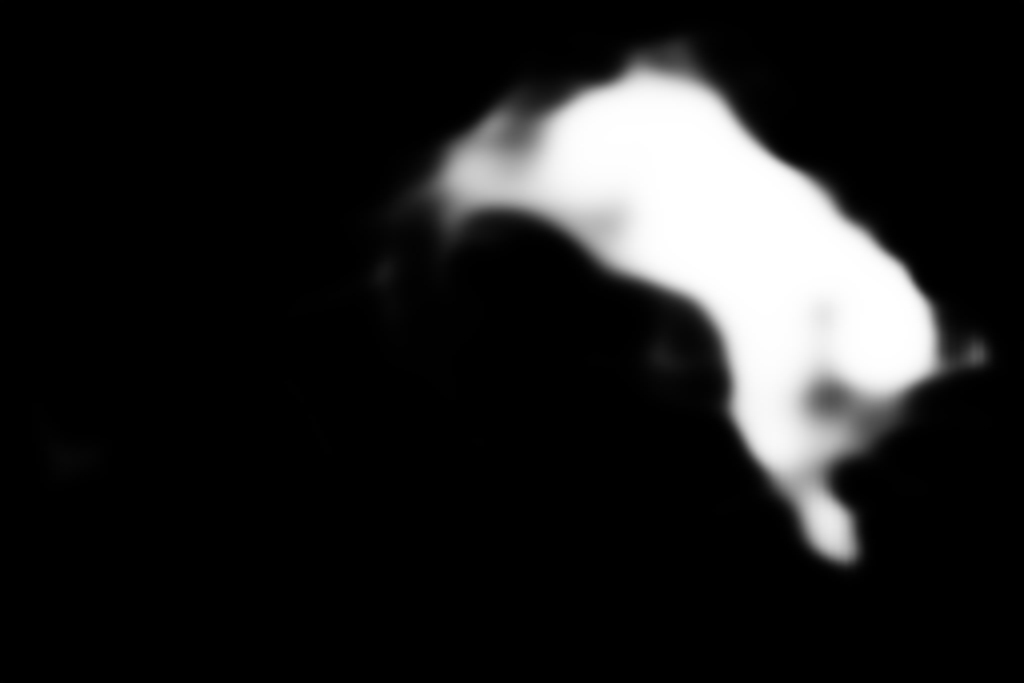}
	\end{subfigure}
    \begin{subfigure}{0.105\textwidth}
		\includegraphics[width=\textwidth]{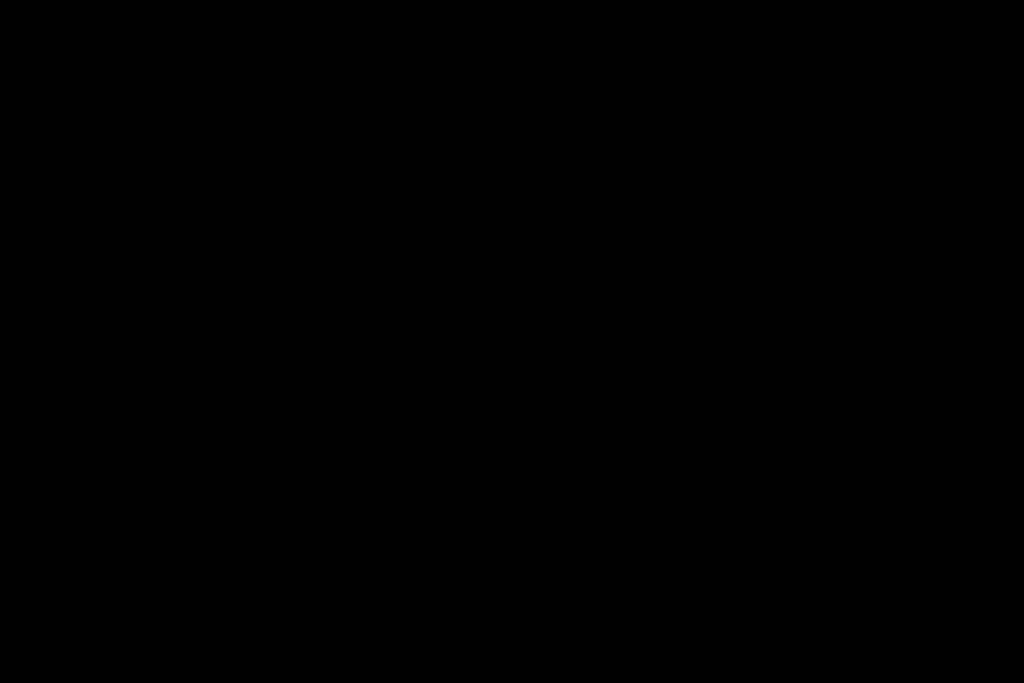}
	\end{subfigure}
	\begin{subfigure}{0.105\textwidth}
		\includegraphics[width=\textwidth]{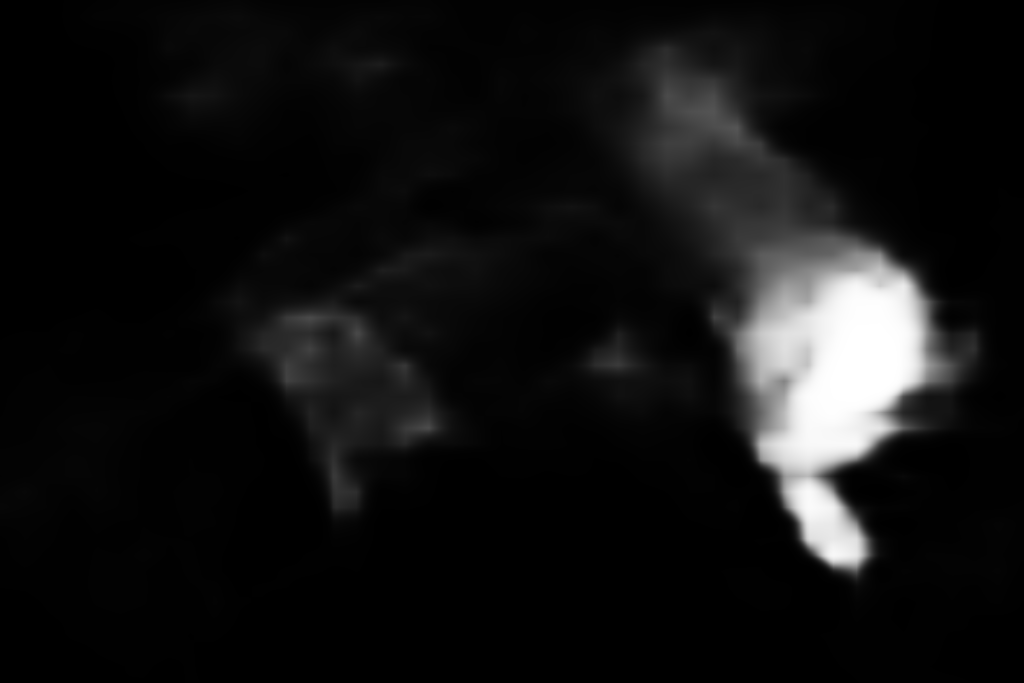}
	\end{subfigure}
	\begin{subfigure}{0.105\textwidth}
		\includegraphics[width=\textwidth]{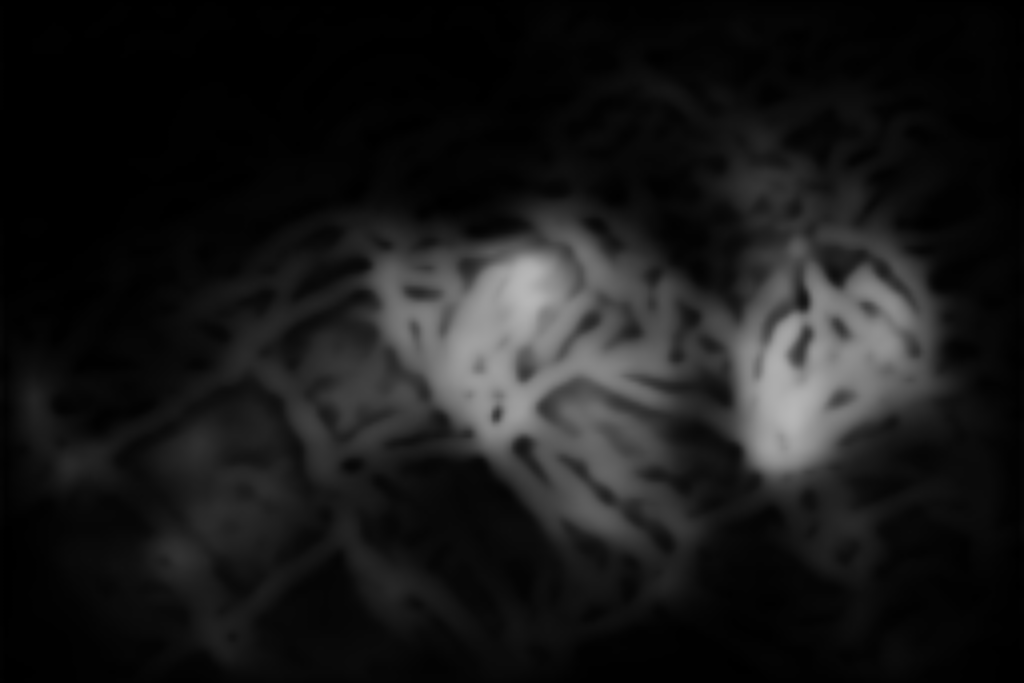}
	\end{subfigure}
    \begin{subfigure}{0.105\textwidth}
		\includegraphics[width=\textwidth]{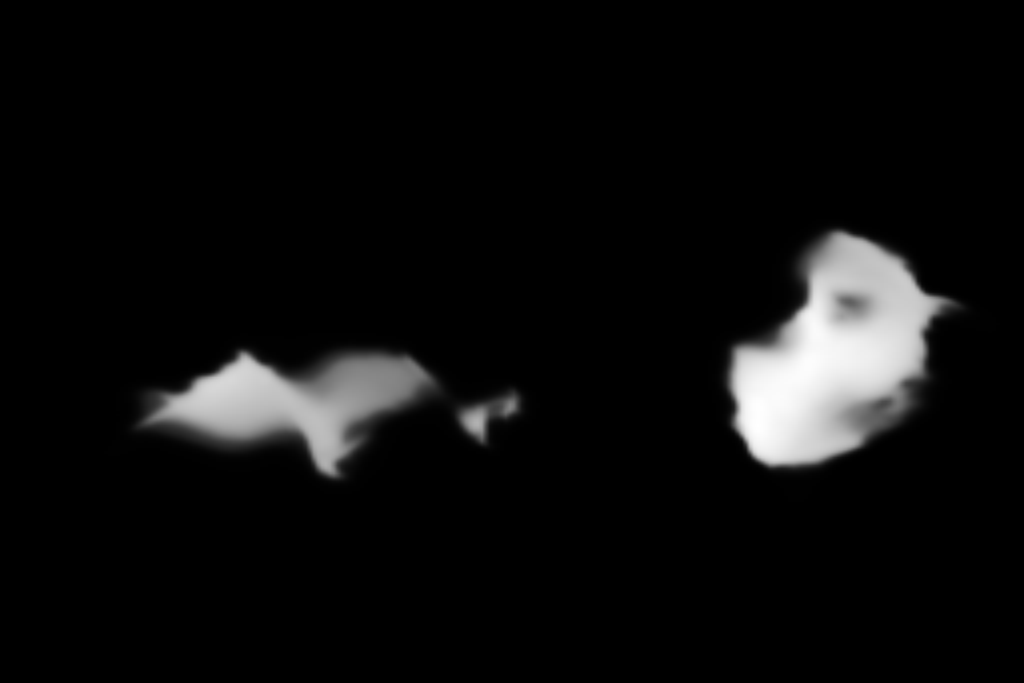}
	\end{subfigure}
	\ \\
	
	\vspace*{0.5mm}
	\begin{subfigure}{0.105\textwidth}
		\includegraphics[width=\textwidth]{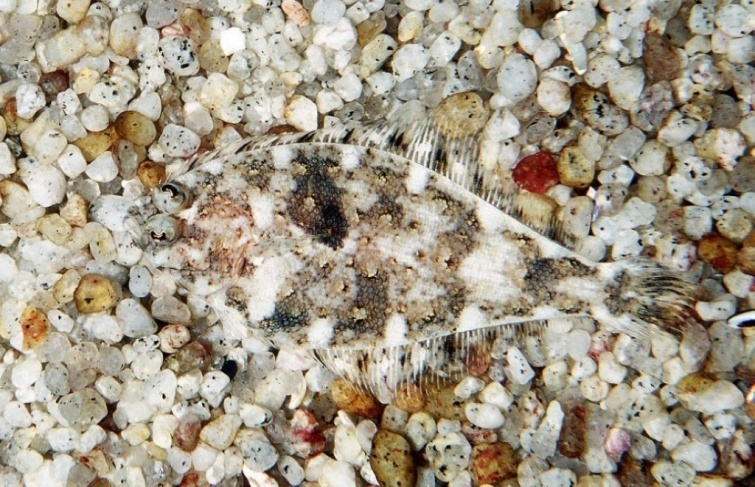}
	\end{subfigure}
	\begin{subfigure}{0.105\textwidth}
		\includegraphics[width=\textwidth]{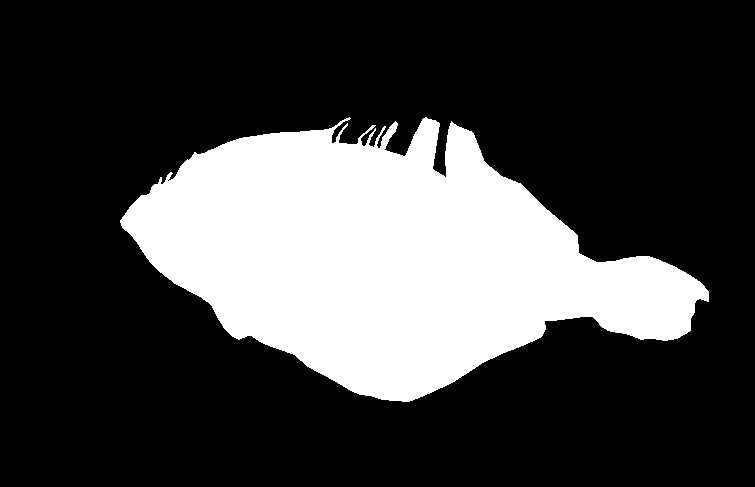}
	\end{subfigure}
	\begin{subfigure}{0.105\textwidth}
		\includegraphics[width=\textwidth]{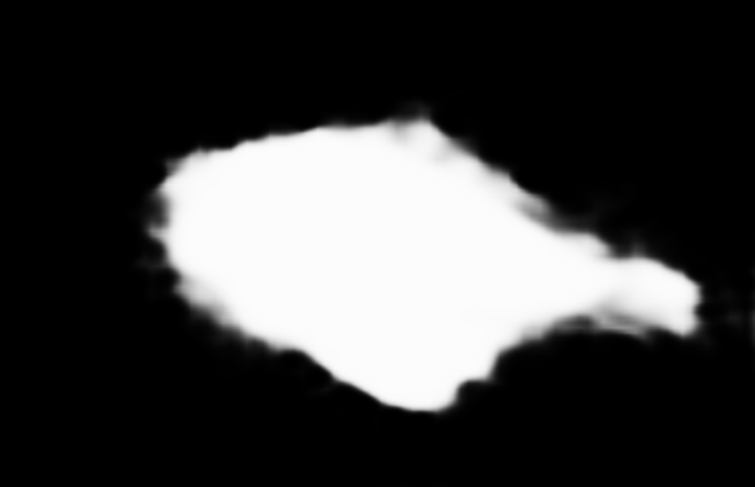}
	\end{subfigure}
    \begin{subfigure}{0.105\textwidth}
		\includegraphics[width=\textwidth]{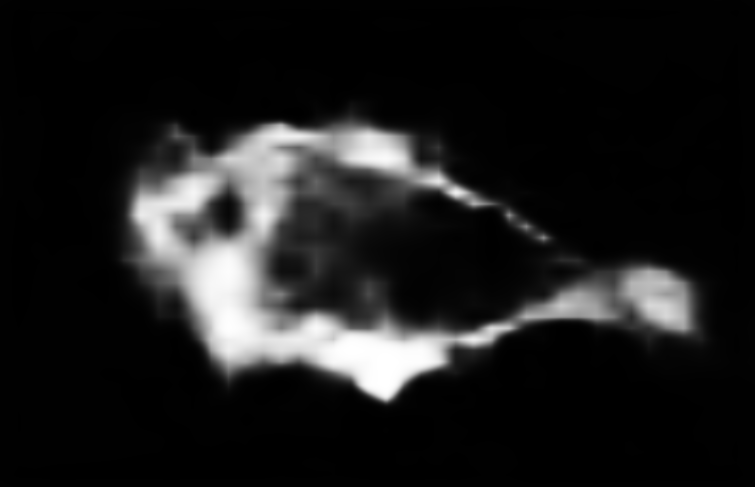}
	\end{subfigure}
    \begin{subfigure}{0.105\textwidth}
		\includegraphics[width=\textwidth]{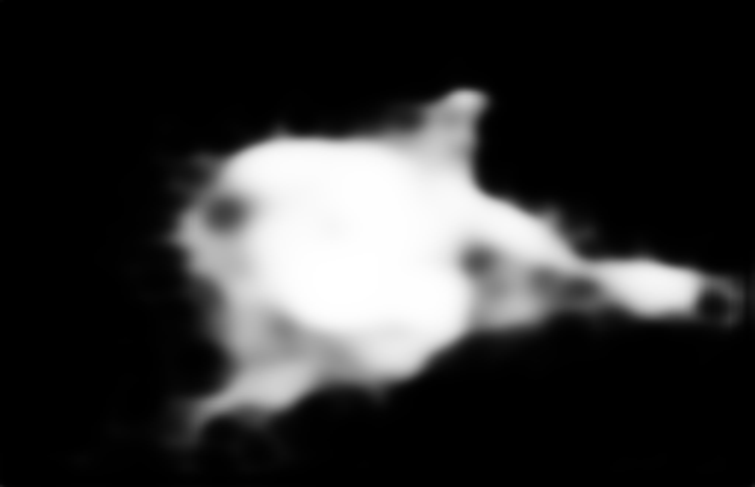}
	\end{subfigure}
    \begin{subfigure}{0.105\textwidth}
		\includegraphics[width=\textwidth]{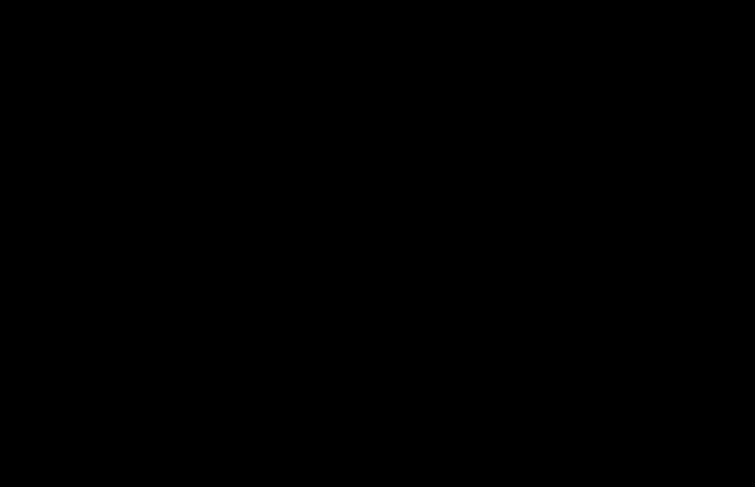}
	\end{subfigure}
	\begin{subfigure}{0.105\textwidth}
		\includegraphics[width=\textwidth]{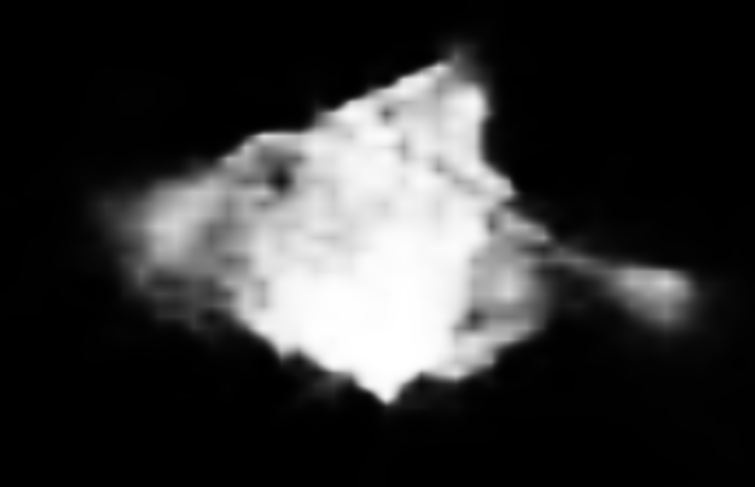}
	\end{subfigure}
	\begin{subfigure}{0.105\textwidth}
		\includegraphics[width=\textwidth]{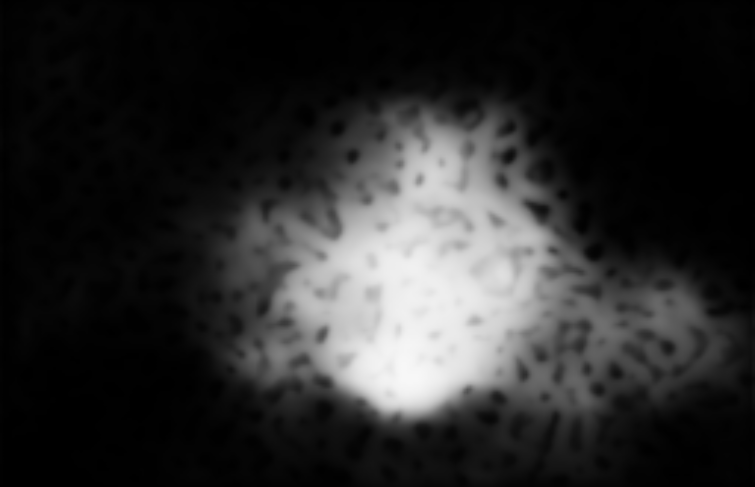}
	\end{subfigure}
    \begin{subfigure}{0.105\textwidth}
		\includegraphics[width=\textwidth]{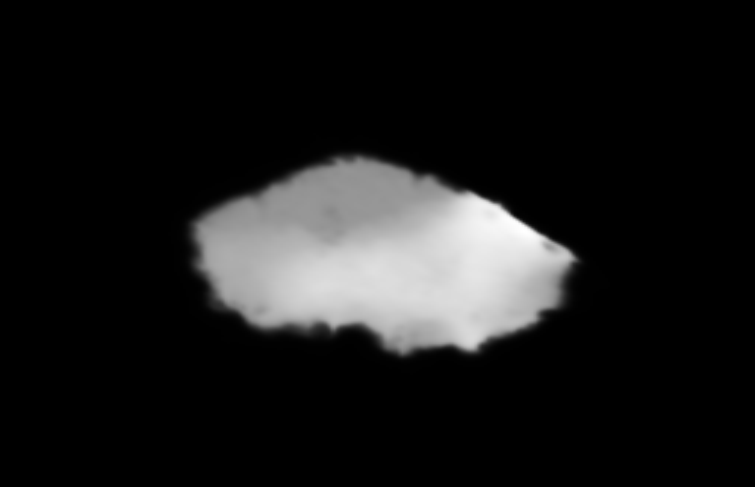}
	\end{subfigure}
	\ \\
	\vspace*{0.5mm}
	\begin{subfigure}{0.105\textwidth}
		\includegraphics[width=\textwidth]{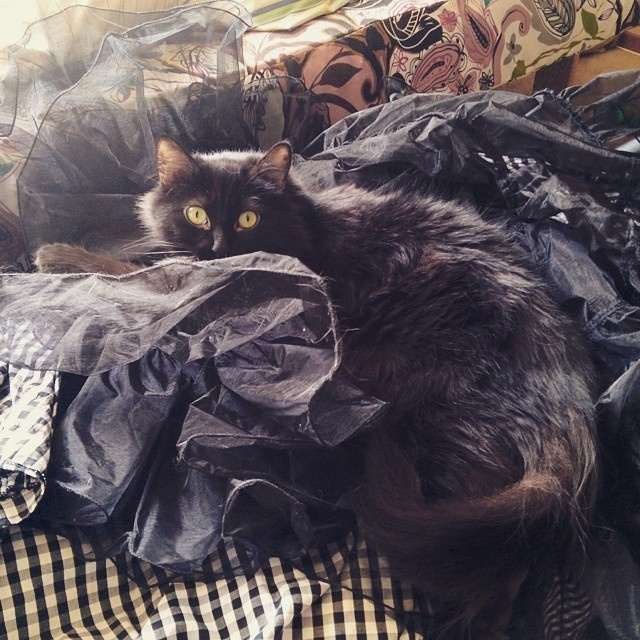}
	\end{subfigure}
	\begin{subfigure}{0.105\textwidth}
		\includegraphics[width=\textwidth]{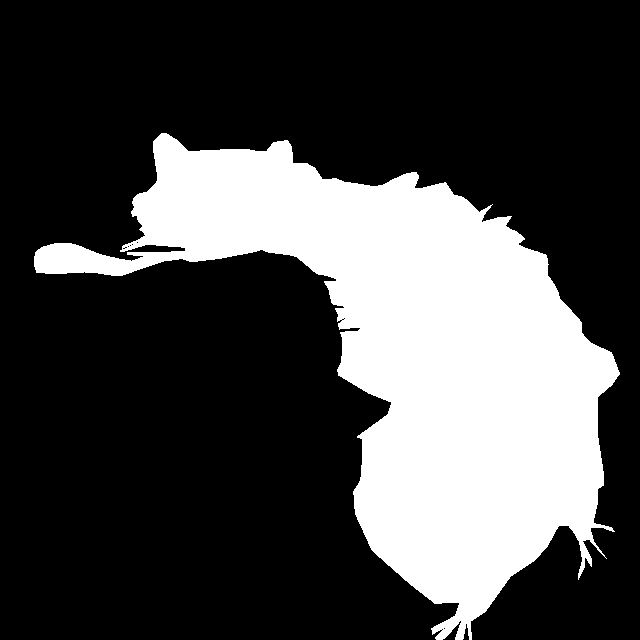}
	\end{subfigure}
	\begin{subfigure}{0.105\textwidth}
		\includegraphics[width=\textwidth]{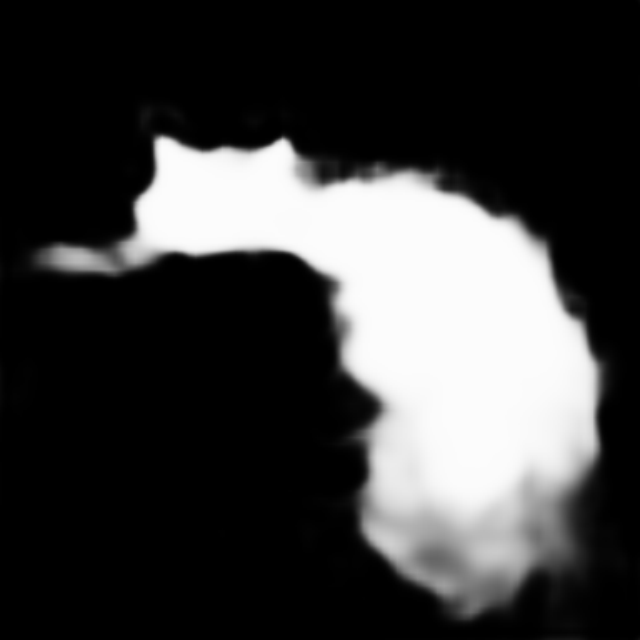}
	\end{subfigure}
    \begin{subfigure}{0.105\textwidth}
		\includegraphics[width=\textwidth]{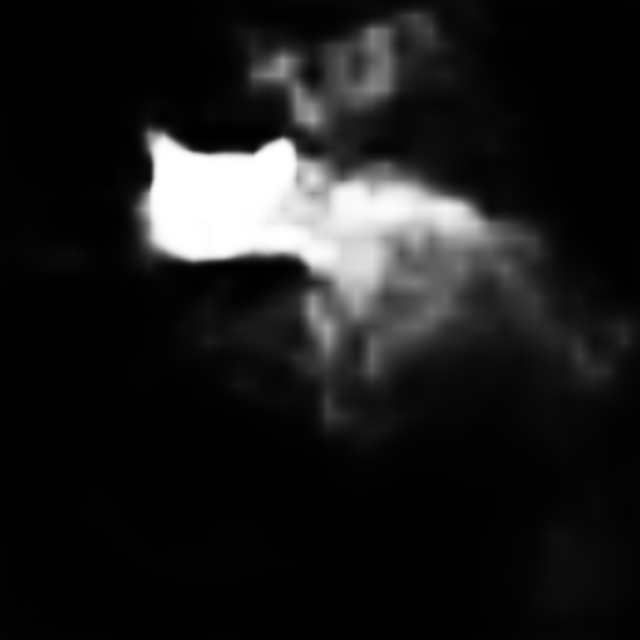}
	\end{subfigure}
    \begin{subfigure}{0.105\textwidth}
		\includegraphics[width=\textwidth]{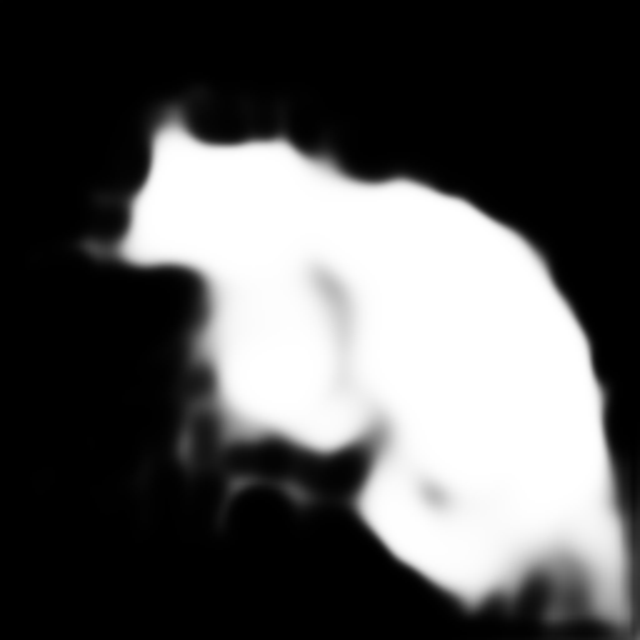}
	\end{subfigure}
    \begin{subfigure}{0.105\textwidth}
		\includegraphics[width=\textwidth]{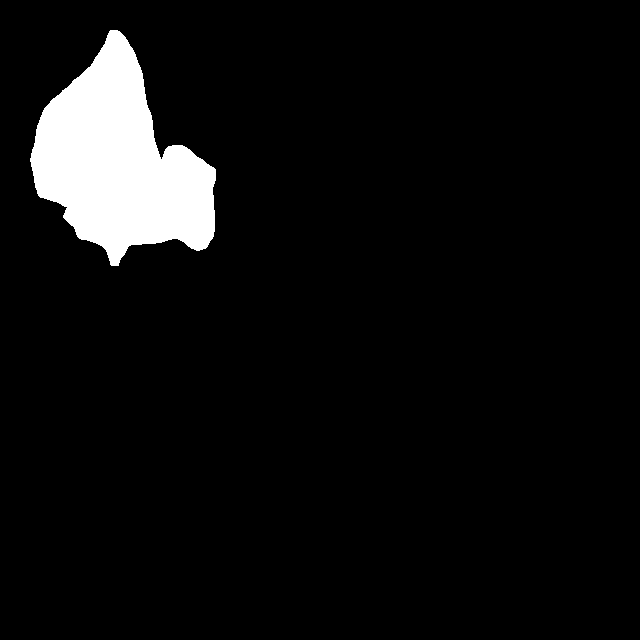}
	\end{subfigure}
	\begin{subfigure}{0.105\textwidth}
		\includegraphics[width=\textwidth]{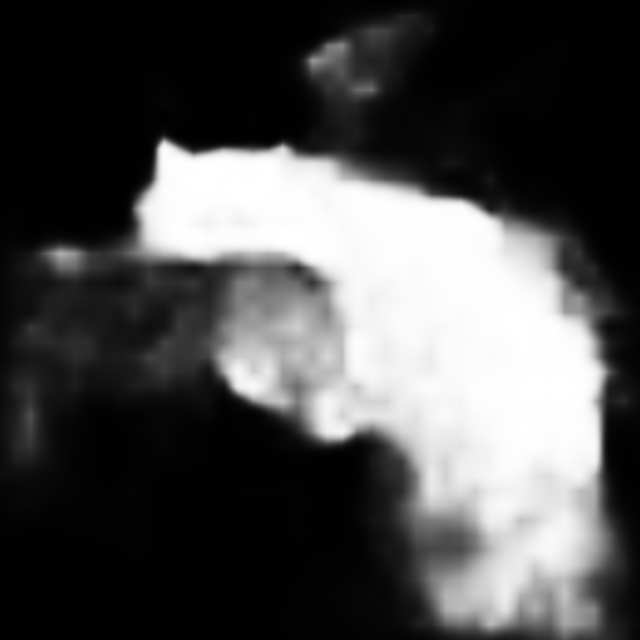}
	\end{subfigure}
	\begin{subfigure}{0.105\textwidth}
		\includegraphics[width=\textwidth]{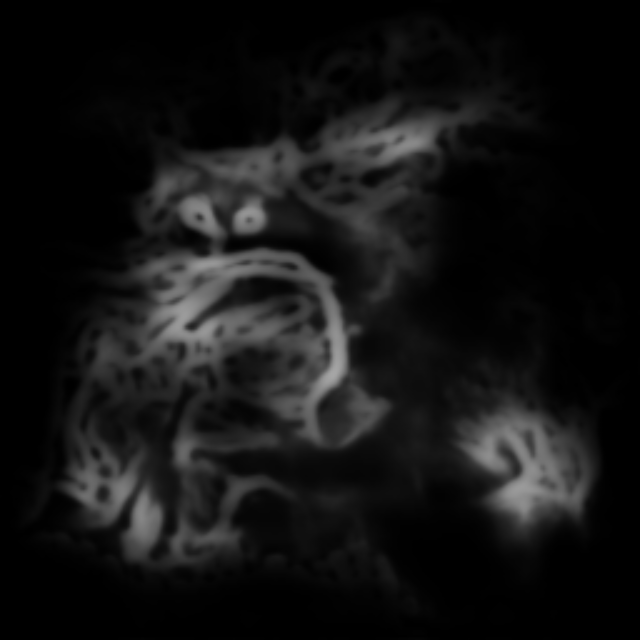}
	\end{subfigure}
    \begin{subfigure}{0.105\textwidth}
		\includegraphics[width=\textwidth]{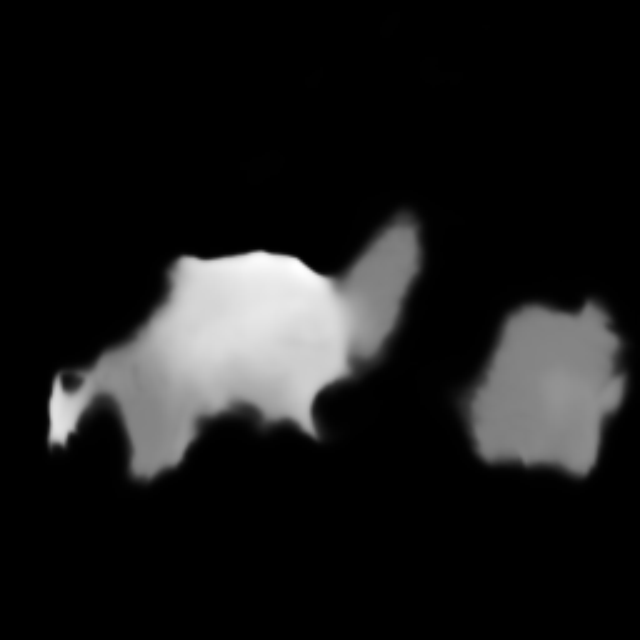}
	\end{subfigure}
	\ \\
	\vspace*{0.5mm}
	\begin{subfigure}{0.105\textwidth}
		\includegraphics[width=\textwidth]{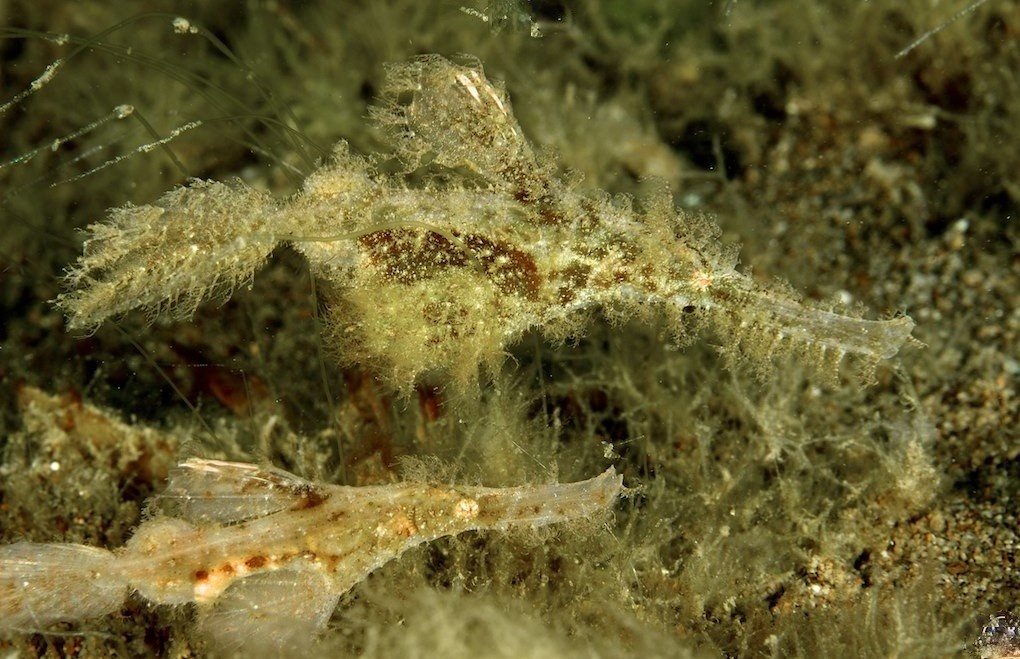}
	\end{subfigure}
	\begin{subfigure}{0.105\textwidth}
		\includegraphics[width=\textwidth]{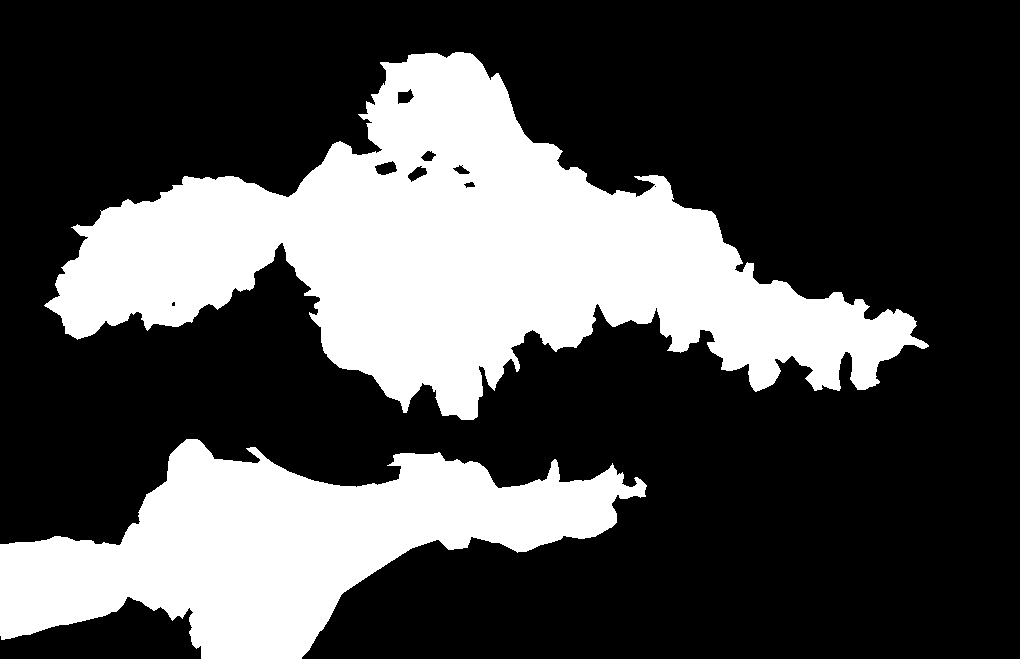}
	\end{subfigure}
	\begin{subfigure}{0.105\textwidth}
		\includegraphics[width=\textwidth]{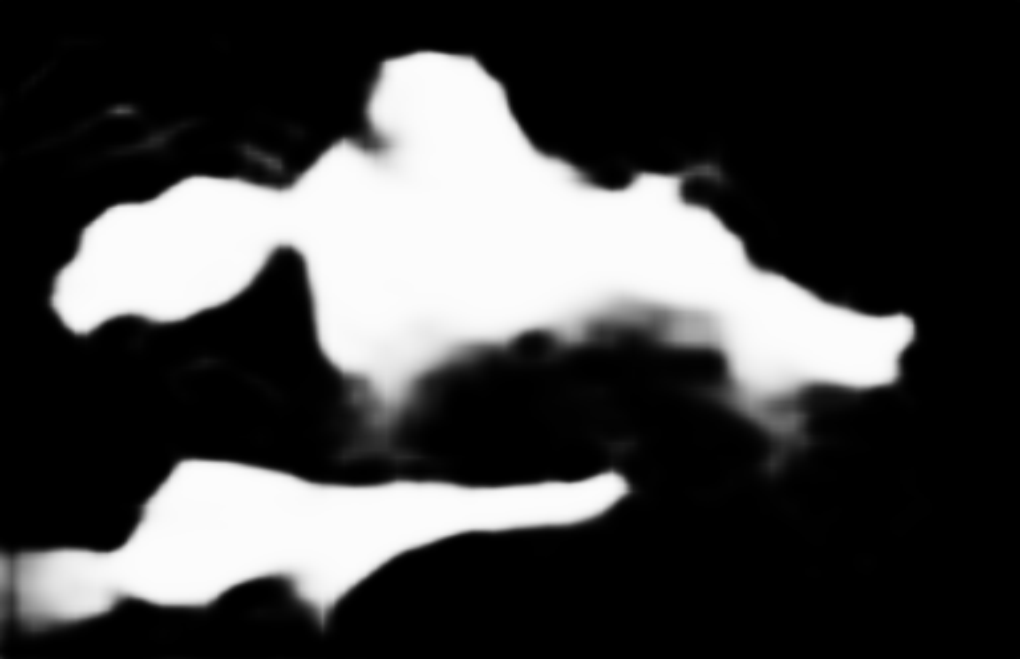}
	\end{subfigure}
    \begin{subfigure}{0.105\textwidth}
		\includegraphics[width=\textwidth]{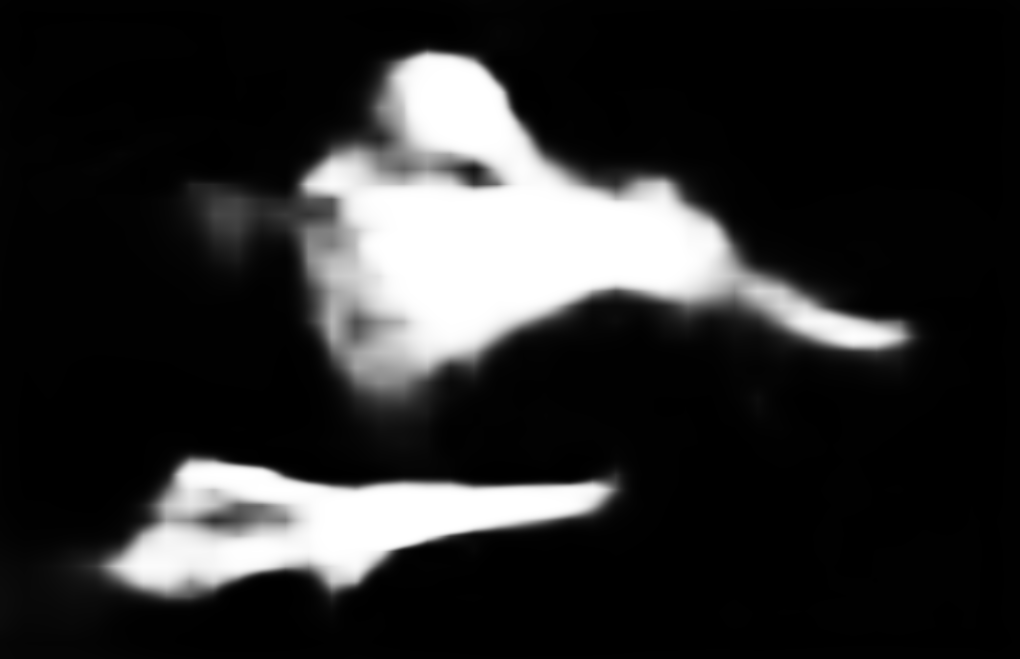}
	\end{subfigure}
    \begin{subfigure}{0.105\textwidth}
		\includegraphics[width=\textwidth]{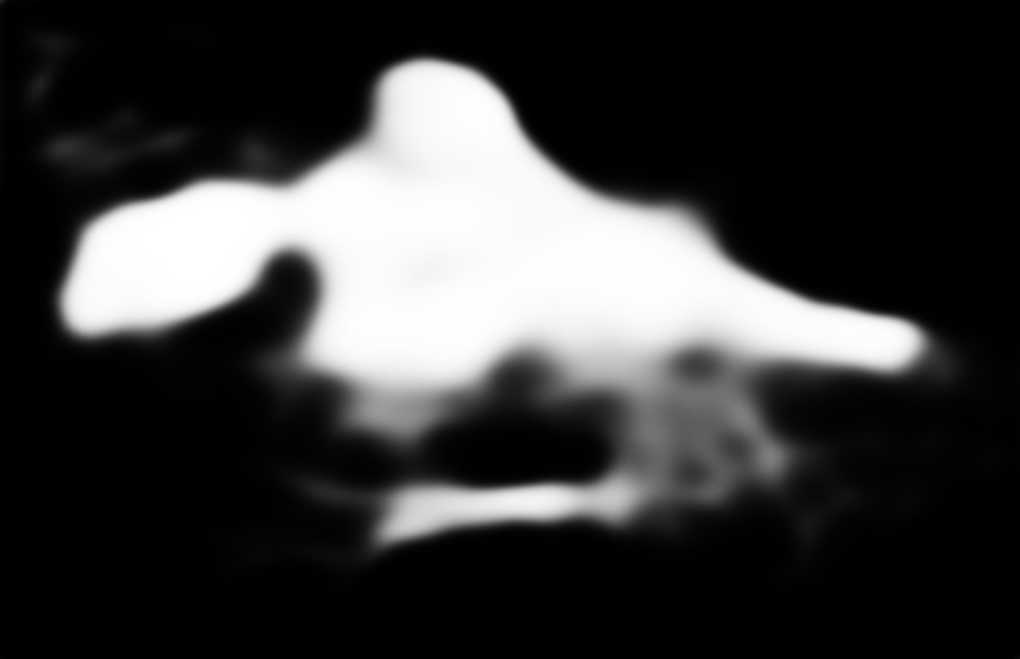}
	\end{subfigure}
    \begin{subfigure}{0.105\textwidth}
		\includegraphics[width=\textwidth]{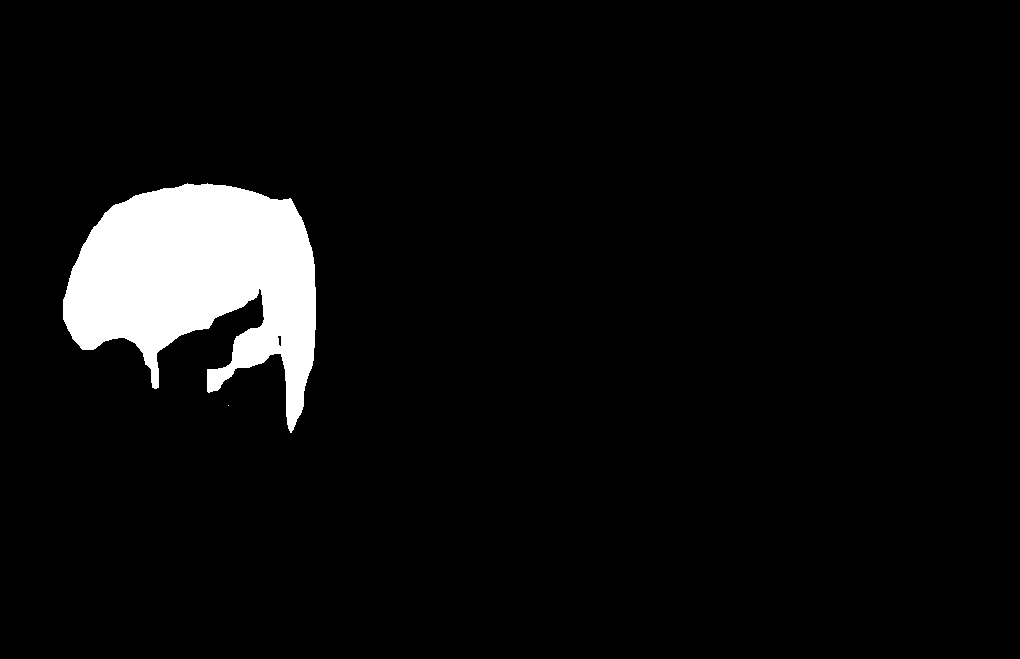}
	\end{subfigure}
	\begin{subfigure}{0.105\textwidth}
		\includegraphics[width=\textwidth]{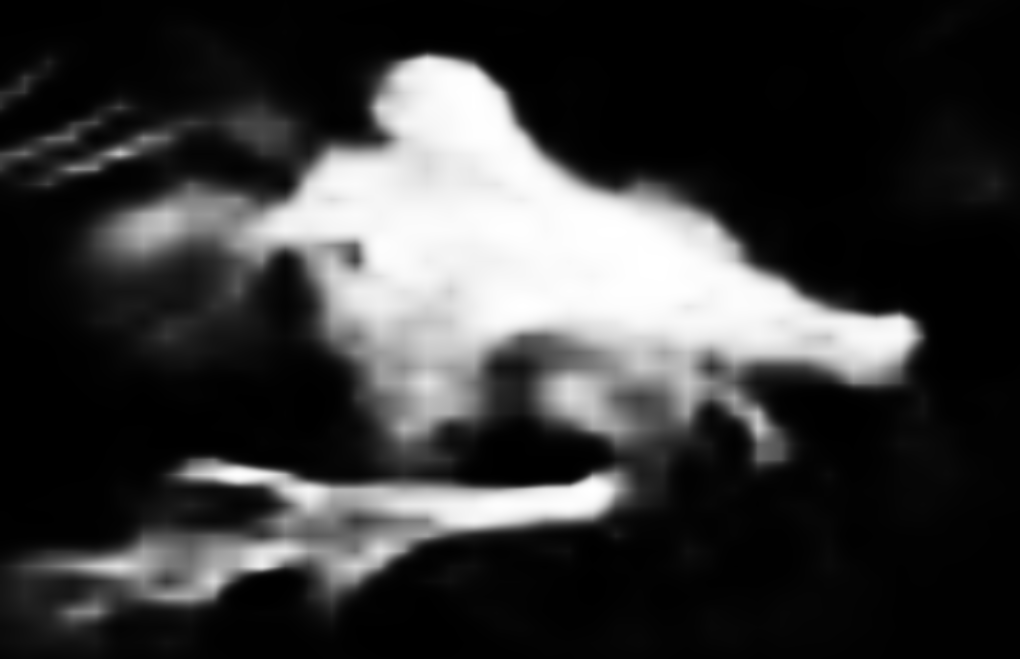}
	\end{subfigure}
	\begin{subfigure}{0.105\textwidth}
		\includegraphics[width=\textwidth]{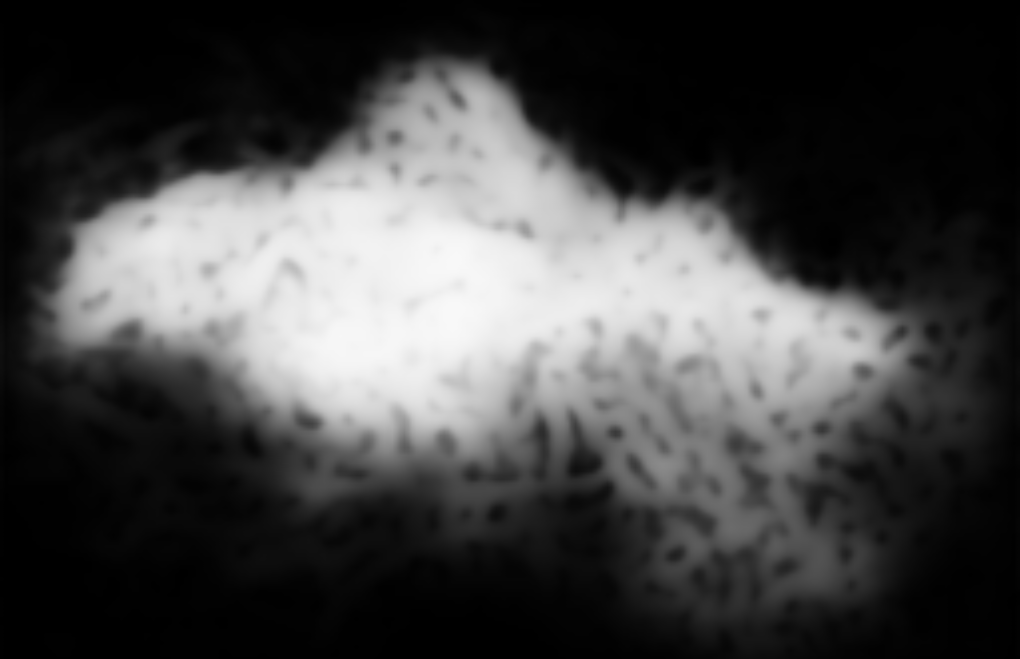}
	\end{subfigure}
    \begin{subfigure}{0.105\textwidth}
		\includegraphics[width=\textwidth]{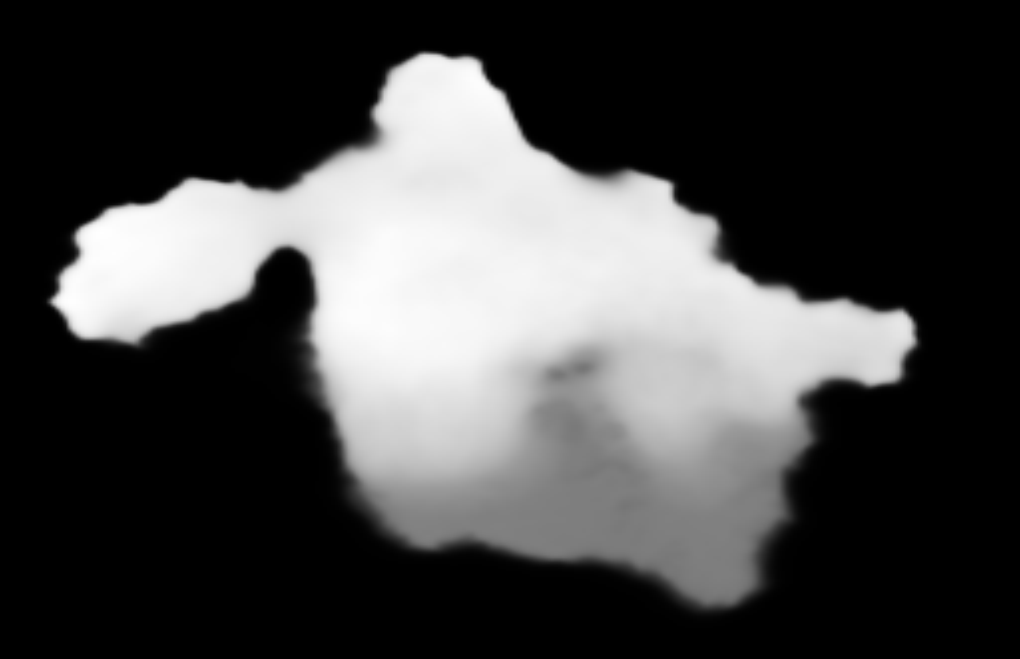}
	\end{subfigure}
	\ \\
	\vspace*{0.5mm}
	\begin{subfigure}{0.105\textwidth}
		\includegraphics[width=\textwidth]{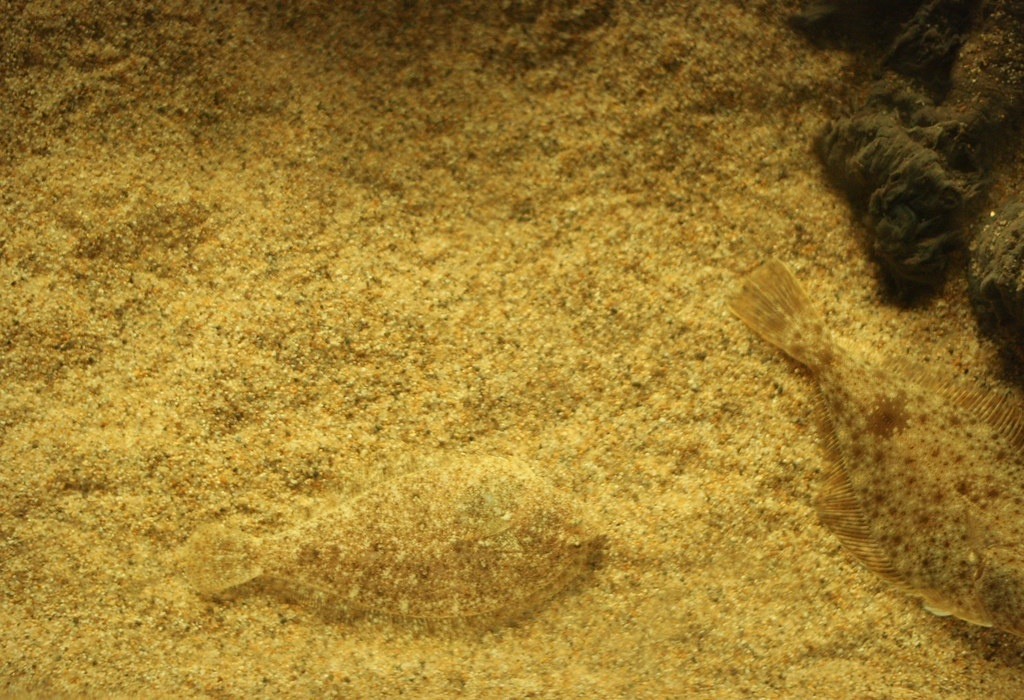}
	\end{subfigure}
	\begin{subfigure}{0.105\textwidth}
		\includegraphics[width=\textwidth]{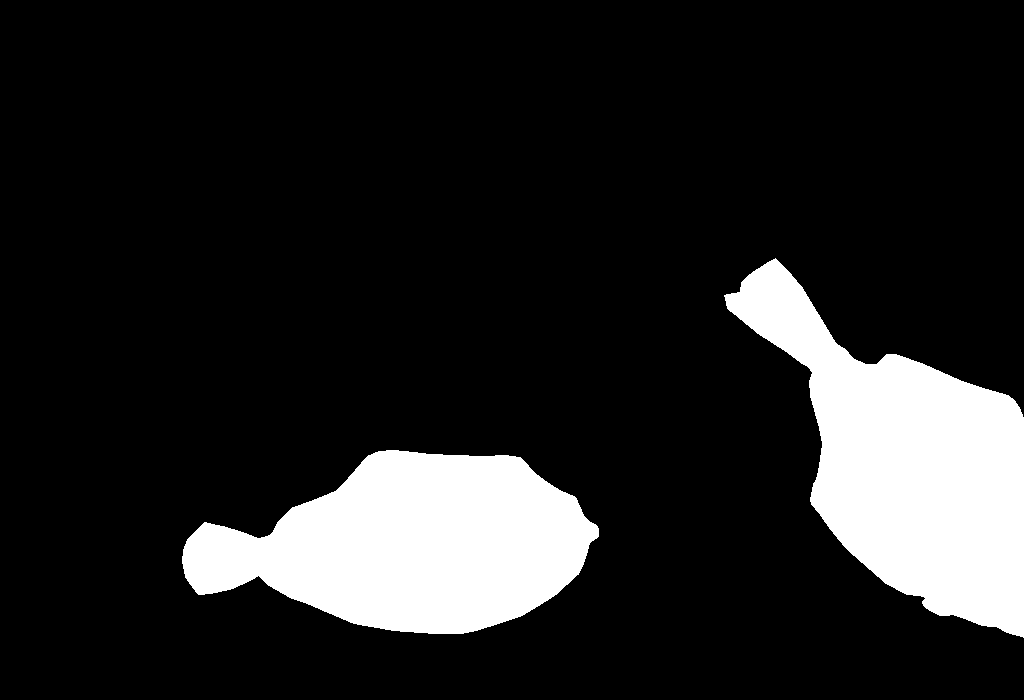}
	\end{subfigure}
	\begin{subfigure}{0.105\textwidth}
		\includegraphics[width=\textwidth]{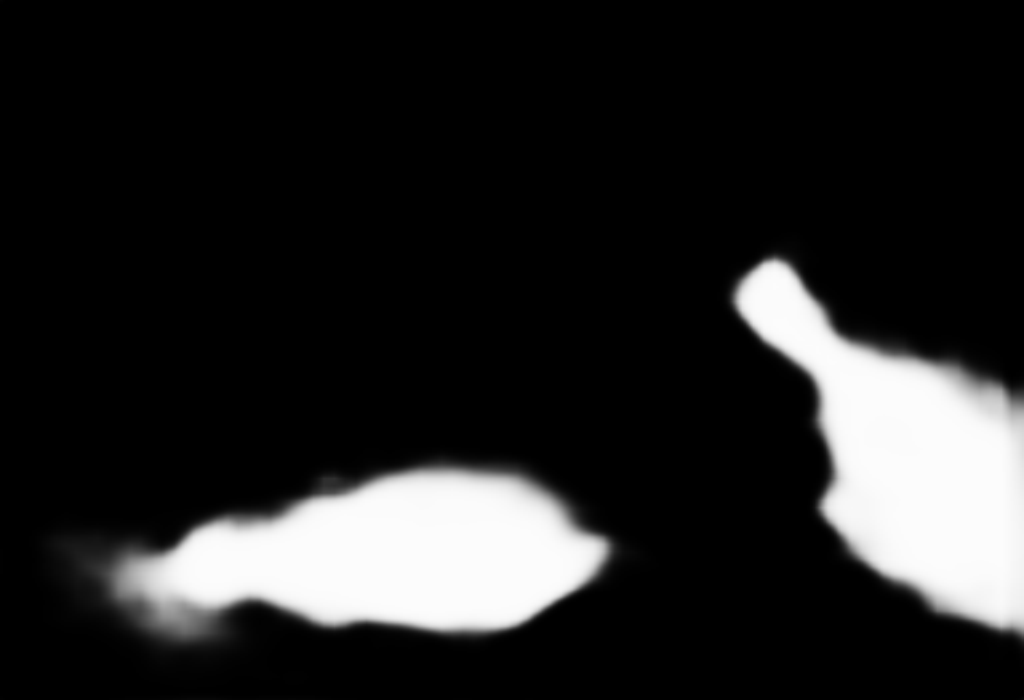}
	\end{subfigure}
    \begin{subfigure}{0.105\textwidth}
		\includegraphics[width=\textwidth]{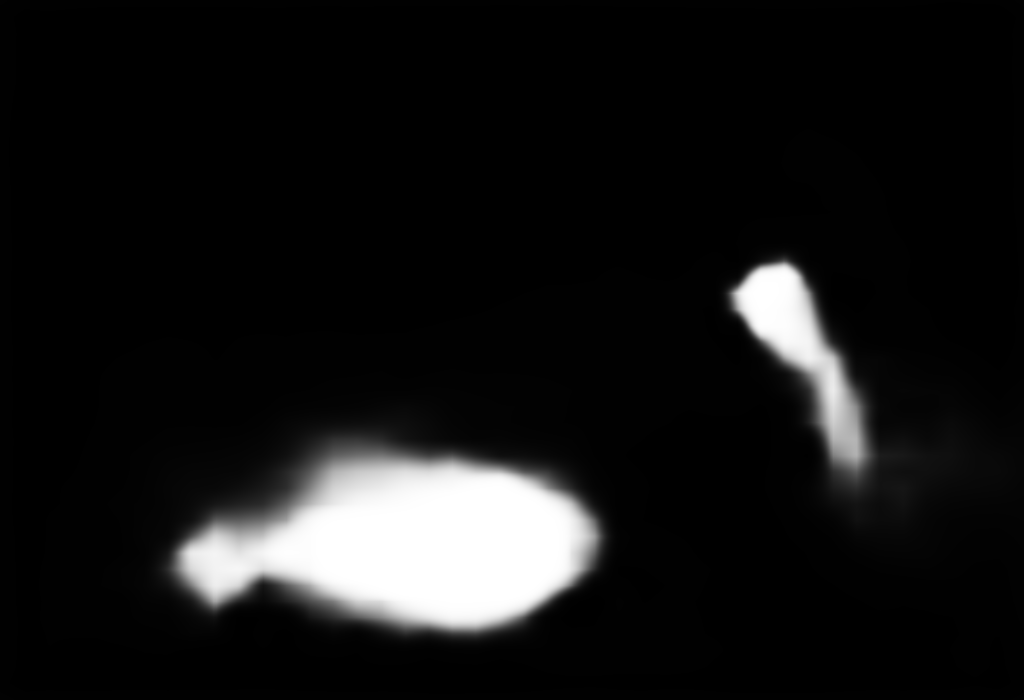}
	\end{subfigure}
    \begin{subfigure}{0.105\textwidth}
		\includegraphics[width=\textwidth]{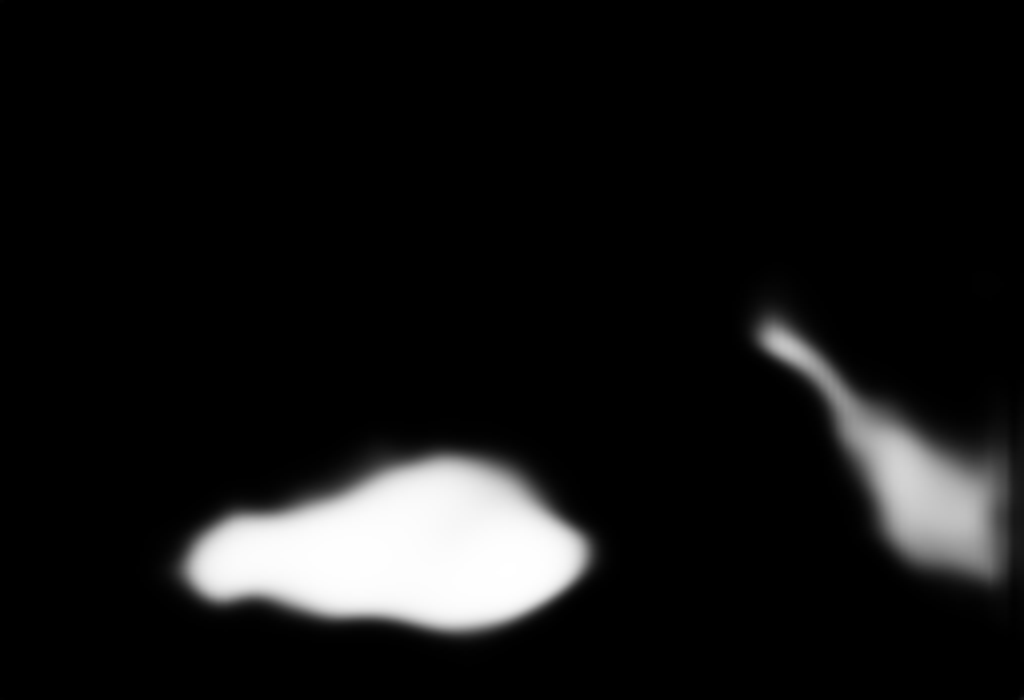}
	\end{subfigure}
    \begin{subfigure}{0.105\textwidth}
		\includegraphics[width=\textwidth]{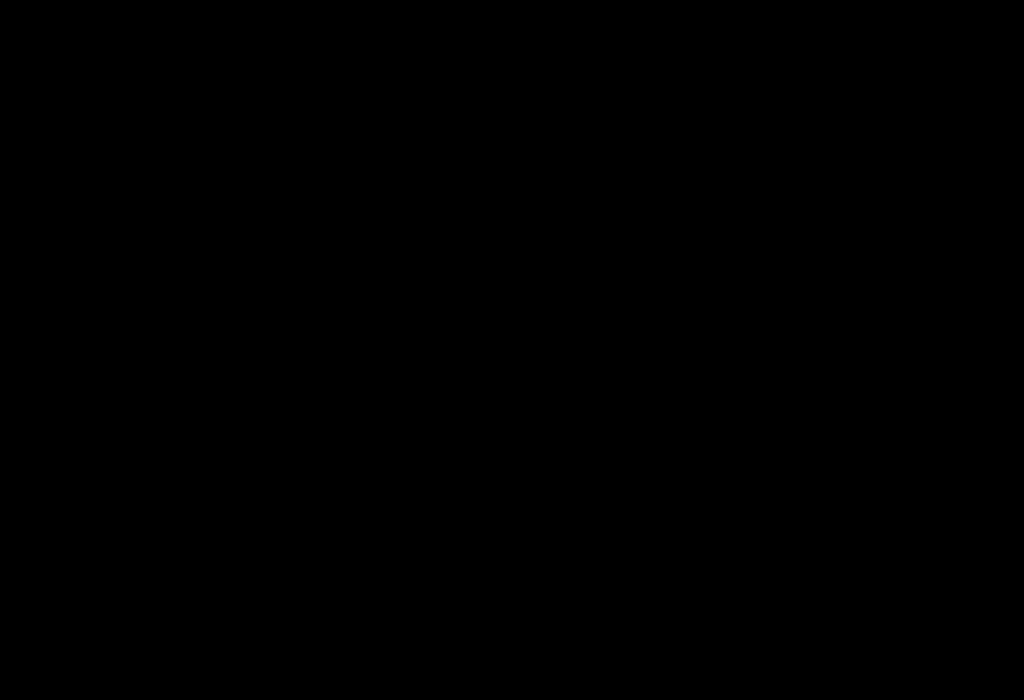}
	\end{subfigure}
	\begin{subfigure}{0.105\textwidth}
		\includegraphics[width=\textwidth]{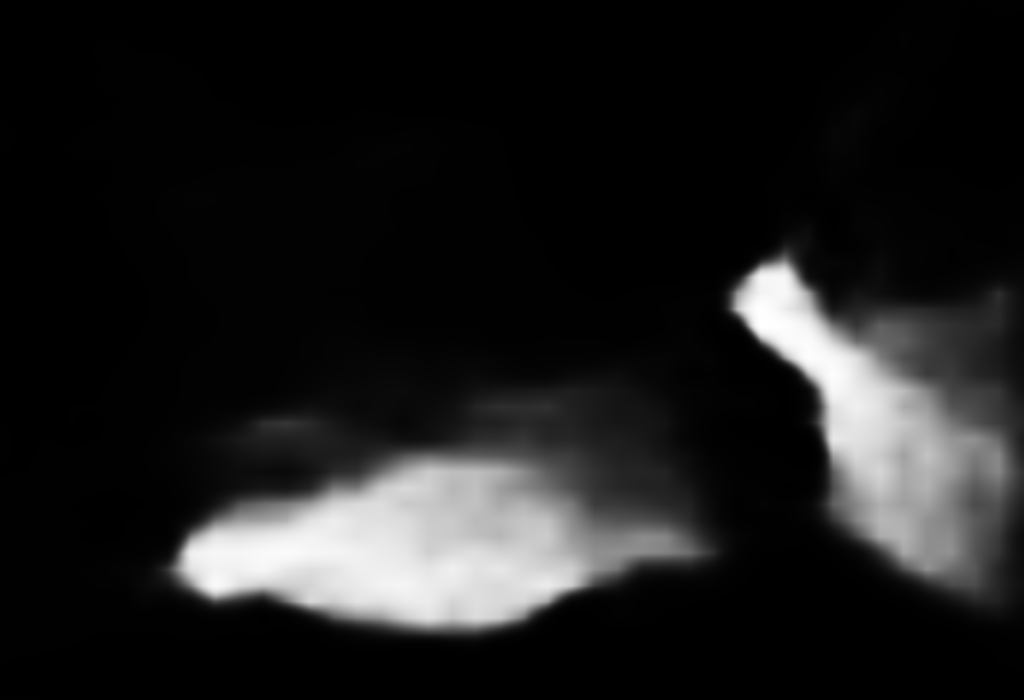}
	\end{subfigure}
	\begin{subfigure}{0.105\textwidth}
		\includegraphics[width=\textwidth]{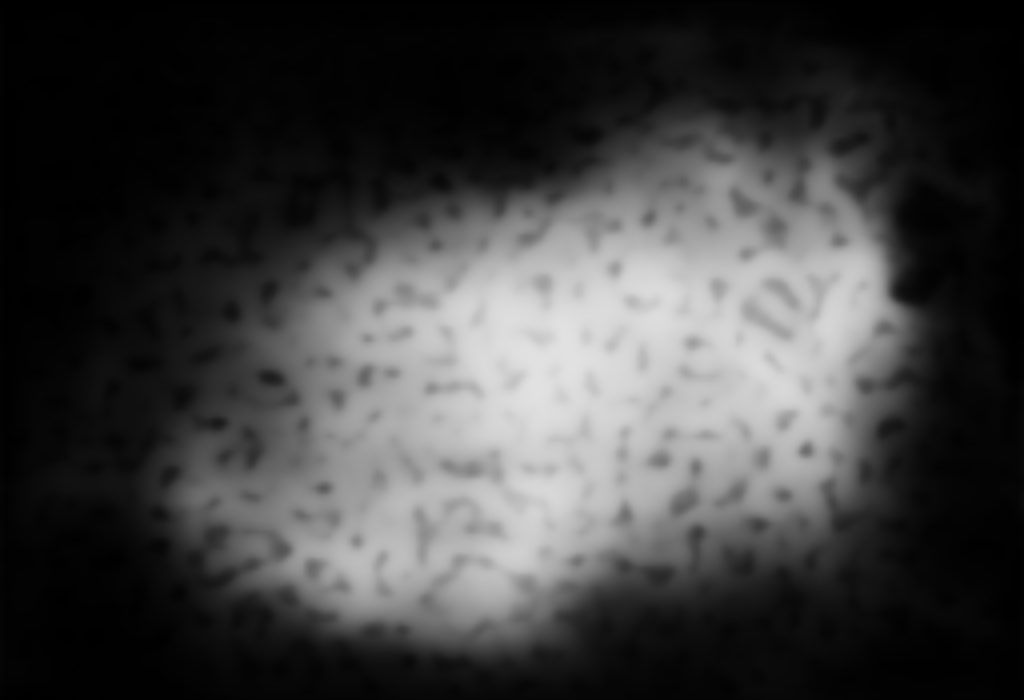}
	\end{subfigure}
    \begin{subfigure}{0.105\textwidth}
		\includegraphics[width=\textwidth]{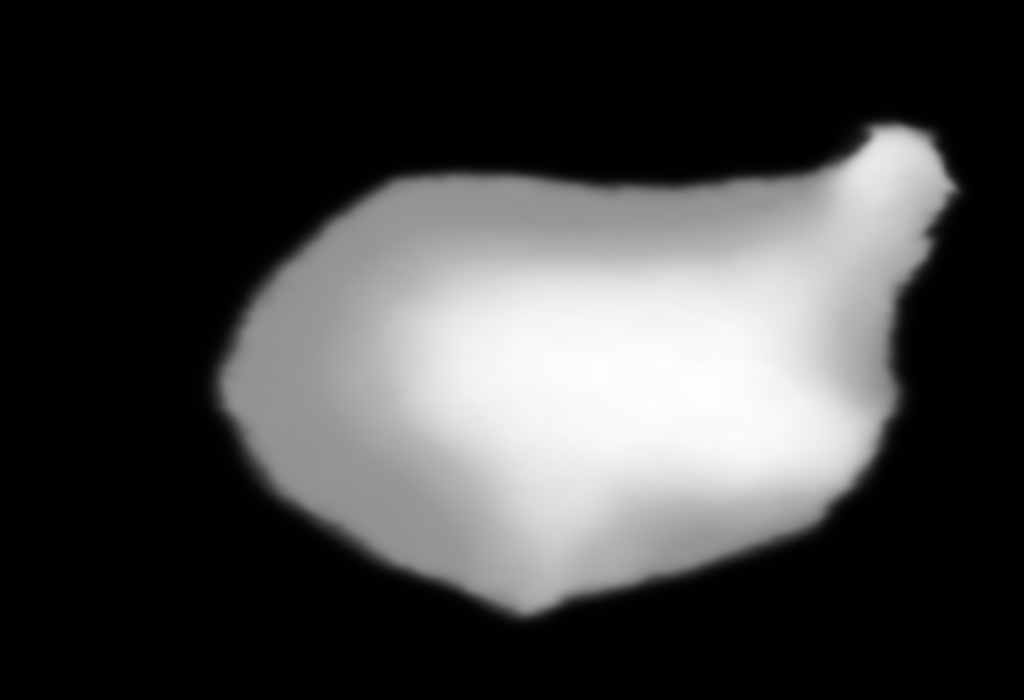}
	\end{subfigure}
	\ \\
	\vspace*{0.5mm}
	\begin{subfigure}{0.105\textwidth}
		\includegraphics[width=\textwidth]{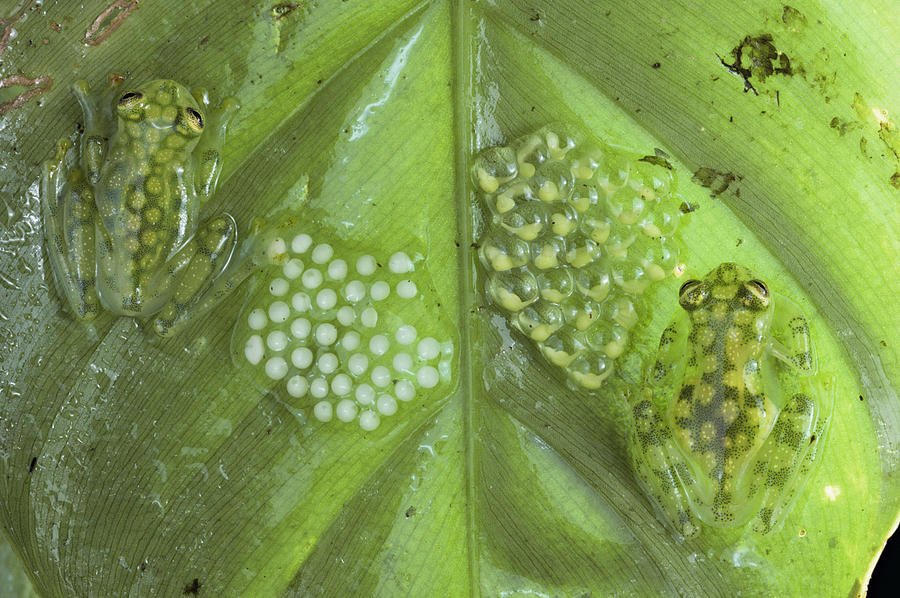}
	\end{subfigure}
	\begin{subfigure}{0.105\textwidth}
		\includegraphics[width=\textwidth]{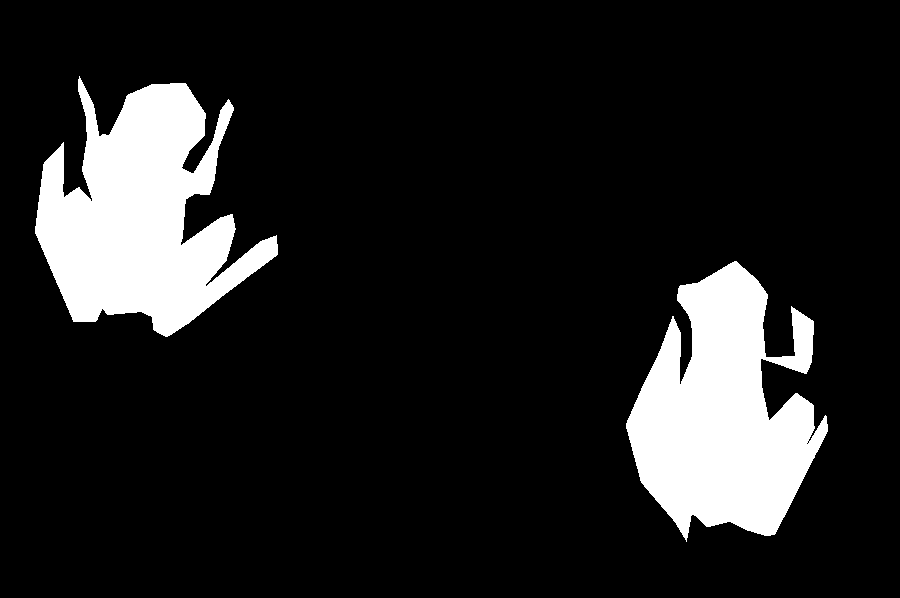}
	\end{subfigure}
	\begin{subfigure}{0.105\textwidth}
		\includegraphics[width=\textwidth]{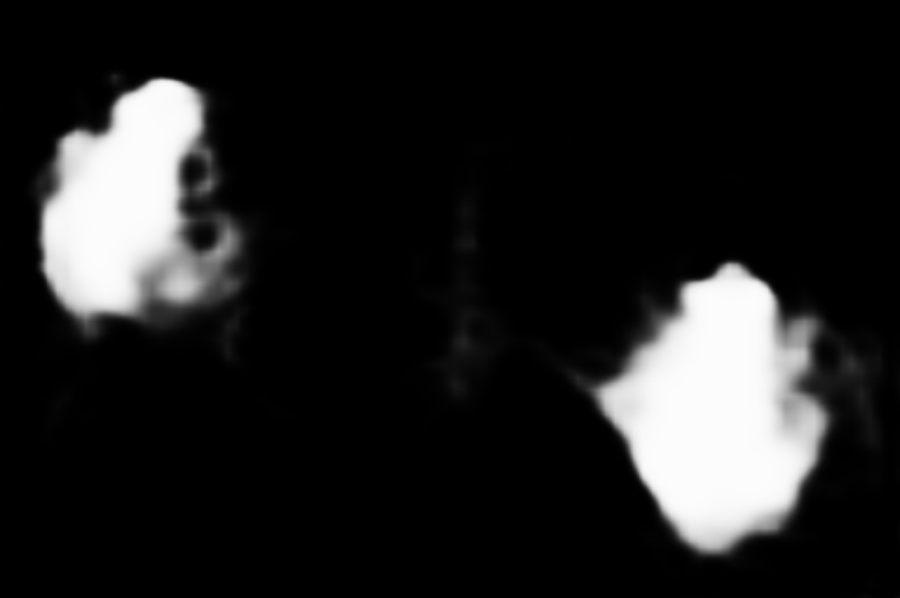}
	\end{subfigure}
    \begin{subfigure}{0.105\textwidth}
		\includegraphics[width=\textwidth]{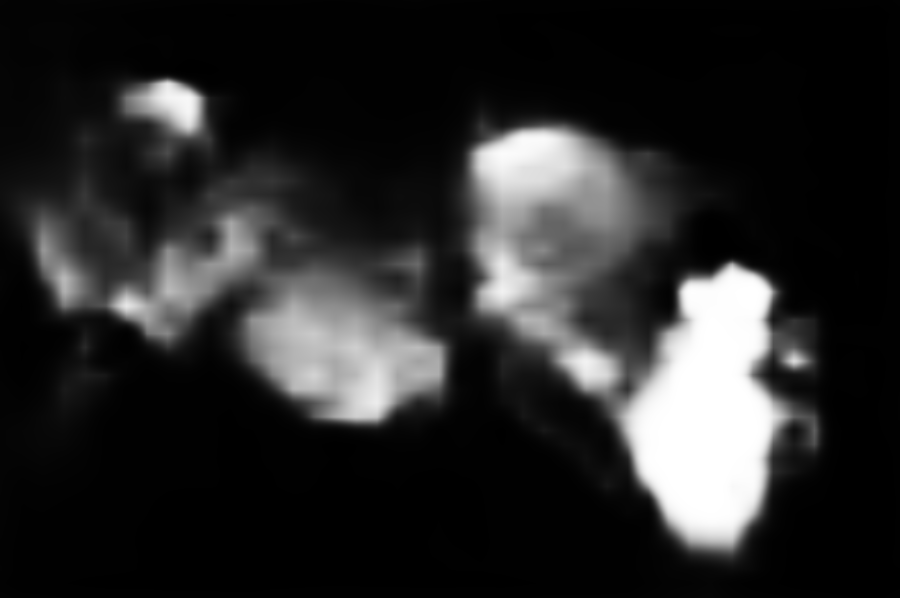}
	\end{subfigure}
    \begin{subfigure}{0.105\textwidth}
		\includegraphics[width=\textwidth]{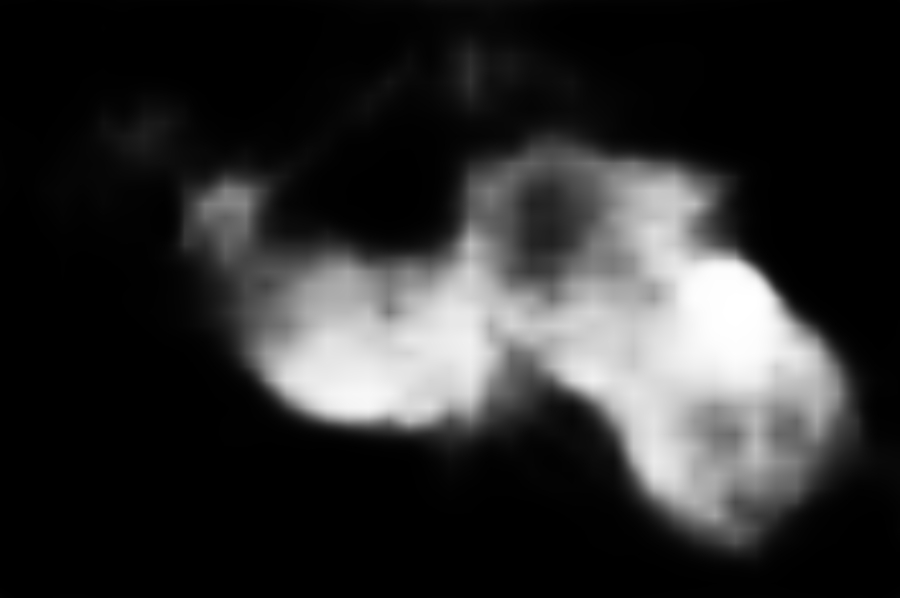}
	\end{subfigure}
    \begin{subfigure}{0.105\textwidth}
		\includegraphics[width=\textwidth]{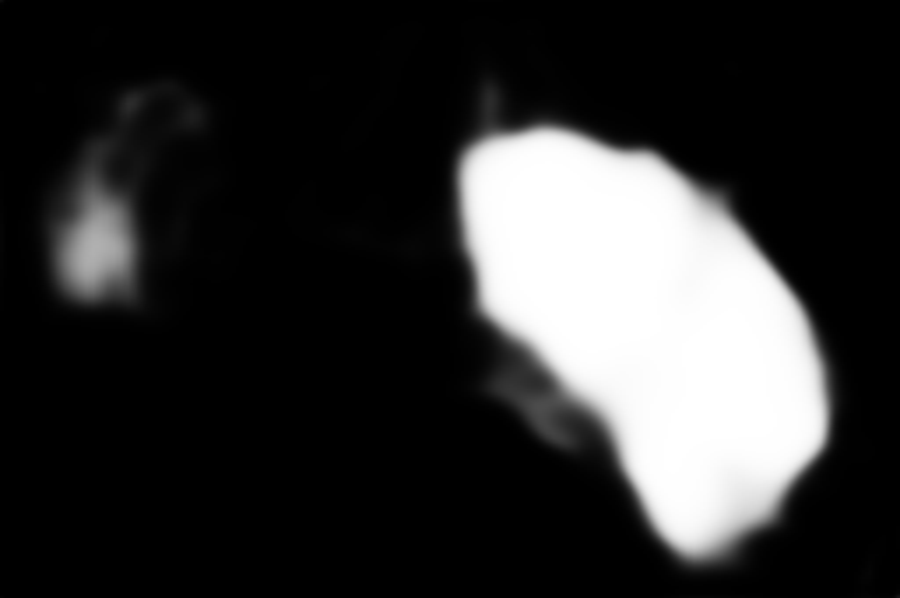}
	\end{subfigure}
	\begin{subfigure}{0.105\textwidth}
		\includegraphics[width=\textwidth]{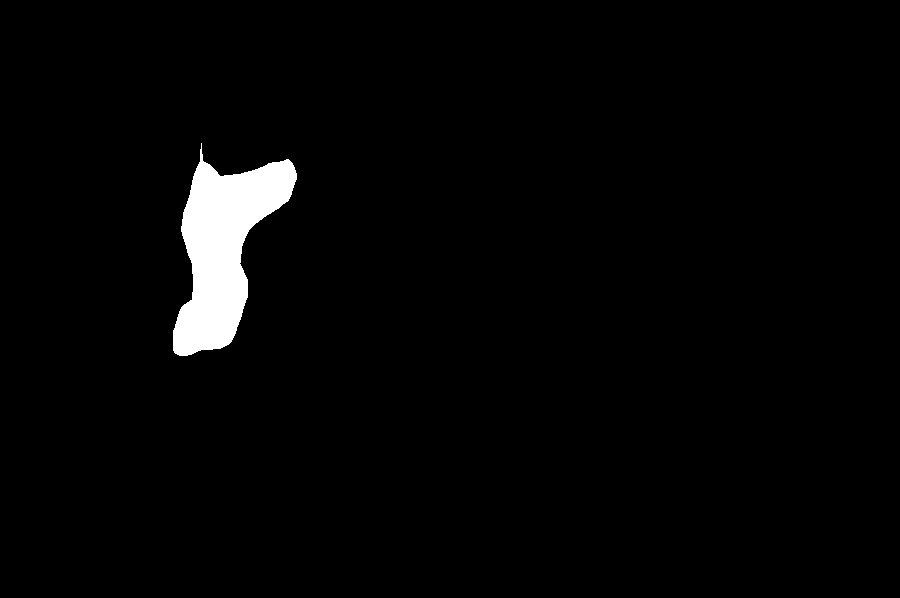}
	\end{subfigure}
	\begin{subfigure}{0.105\textwidth}
		\includegraphics[width=\textwidth]{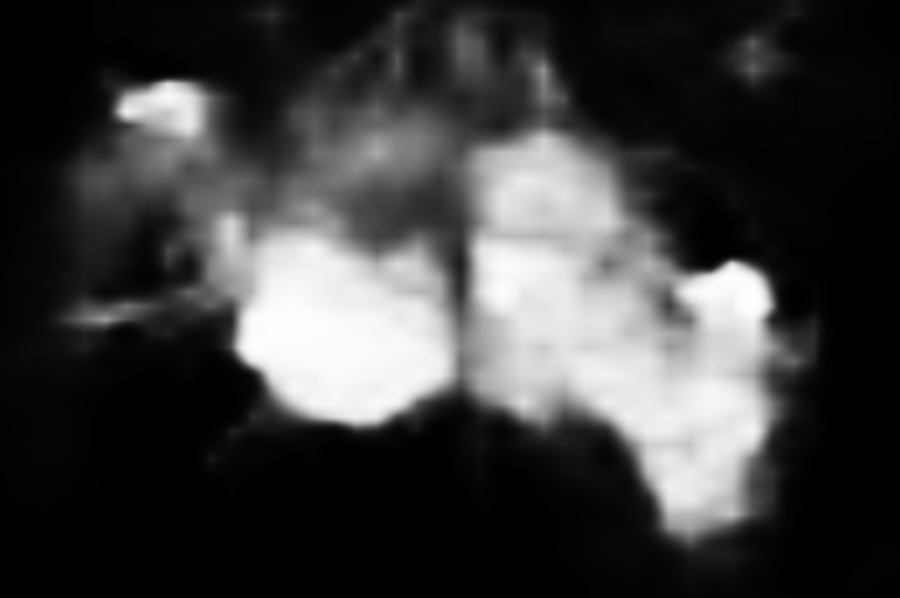}
	\end{subfigure}
    \begin{subfigure}{0.105\textwidth}
		\includegraphics[width=\textwidth]{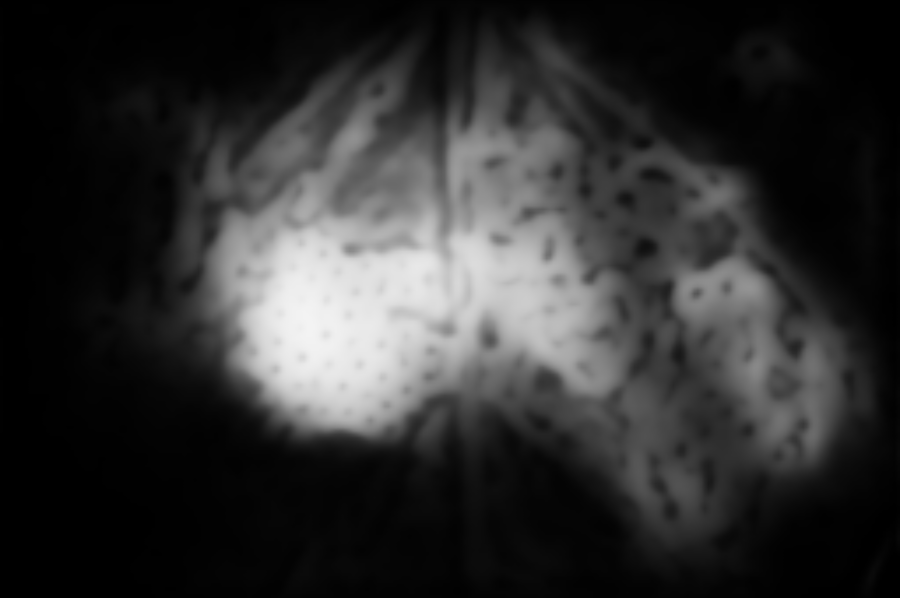}
	\end{subfigure}
	\ \\
	\vspace*{0.5mm}
	\begin{subfigure}{0.105\textwidth}
		\includegraphics[width=\textwidth]{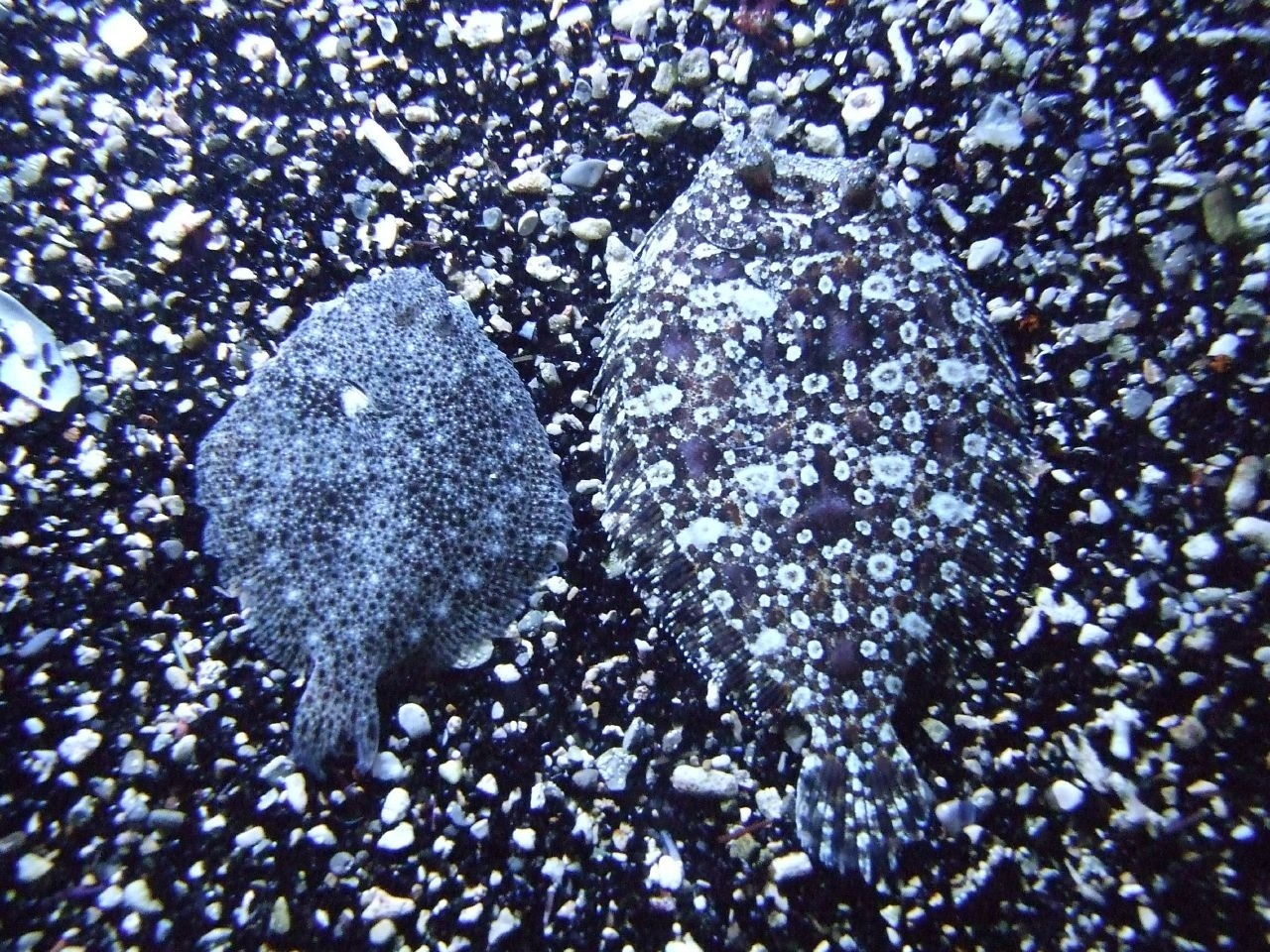}
		\vspace{-5.5mm} \caption{inputs}
	\end{subfigure}
	\begin{subfigure}{0.105\textwidth}
		\includegraphics[width=\textwidth]{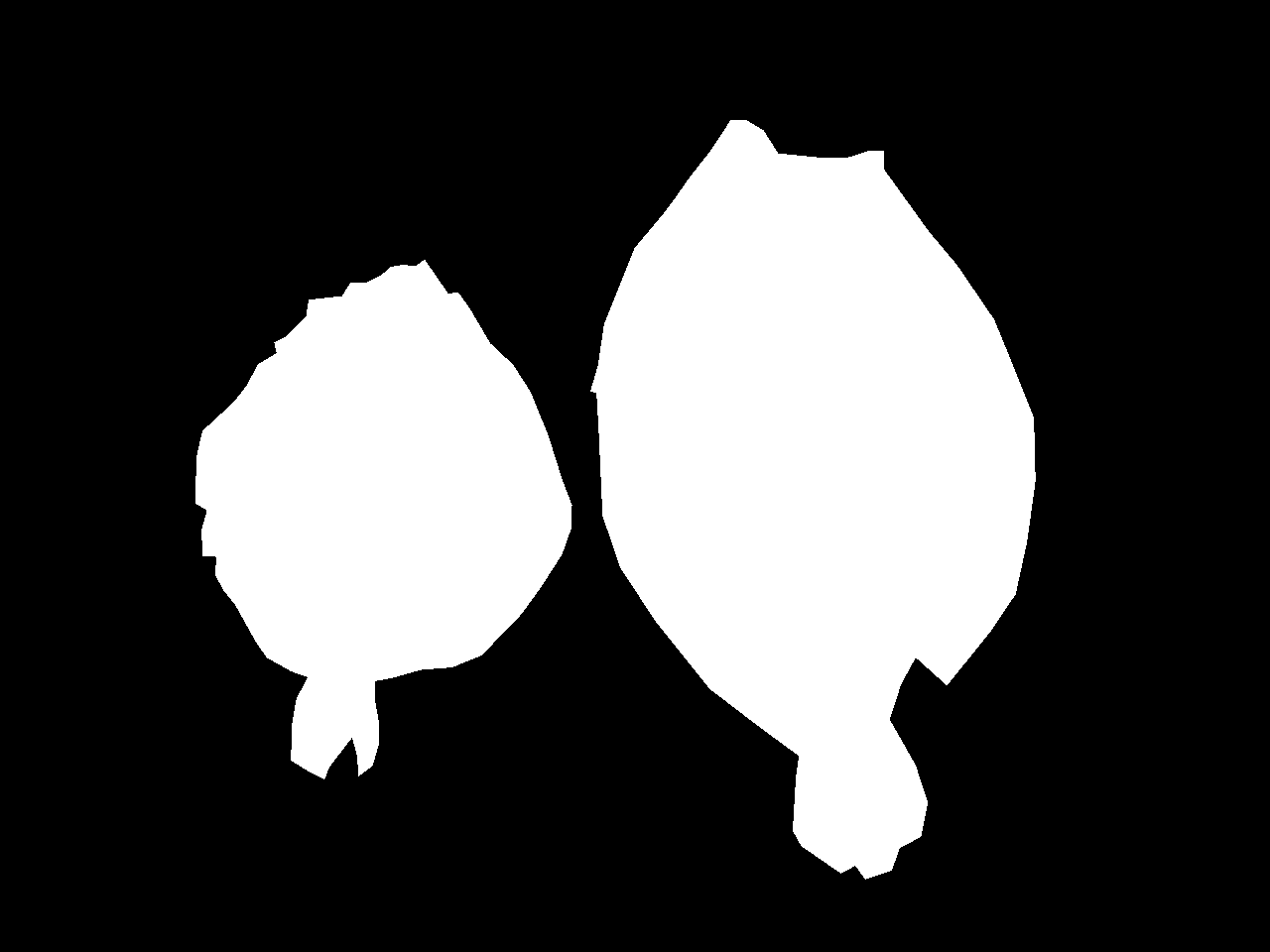}
		\vspace{-5.5mm} \caption{{\footnotesize ground truth}}
	\end{subfigure}
	\begin{subfigure}{0.105\textwidth}
		\includegraphics[width=\textwidth]{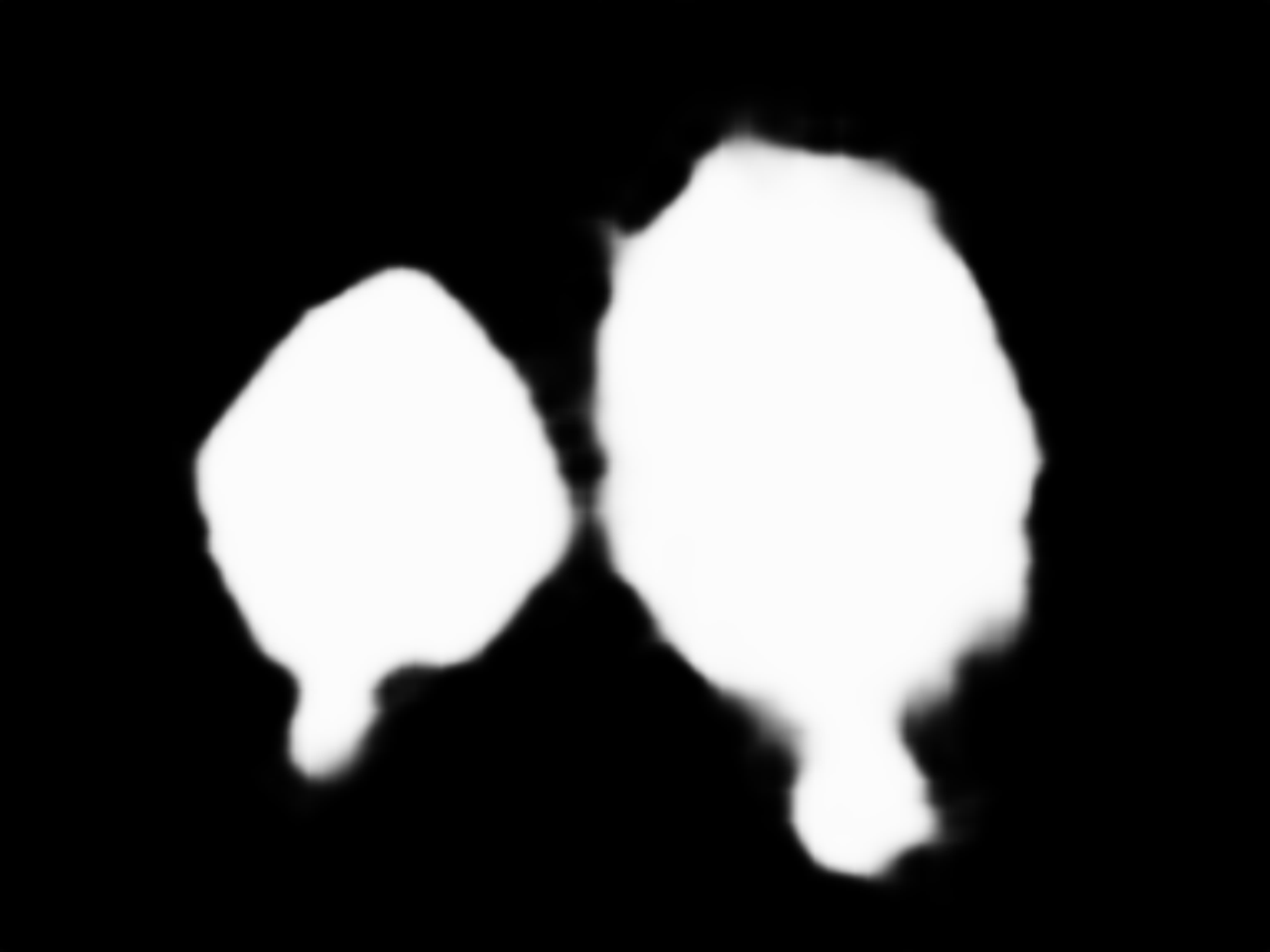}
		\vspace{-5.5mm} \caption{ours}
	\end{subfigure}
    \begin{subfigure}{0.105\textwidth}
		\includegraphics[width=\textwidth]{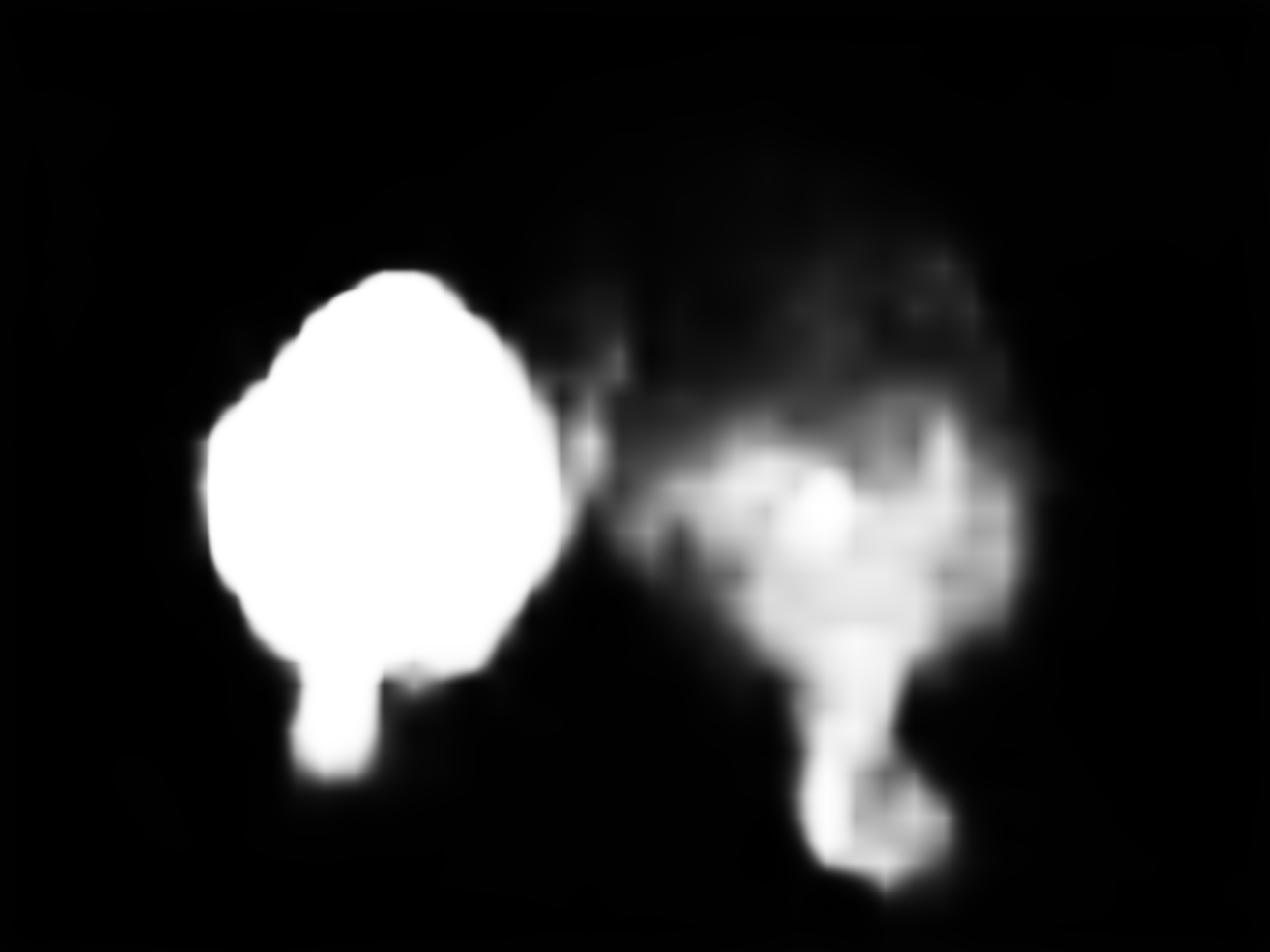}
		\vspace{-5.5mm} \caption{SINet}
	\end{subfigure}
    \begin{subfigure}{0.105\textwidth}
		\includegraphics[width=\textwidth]{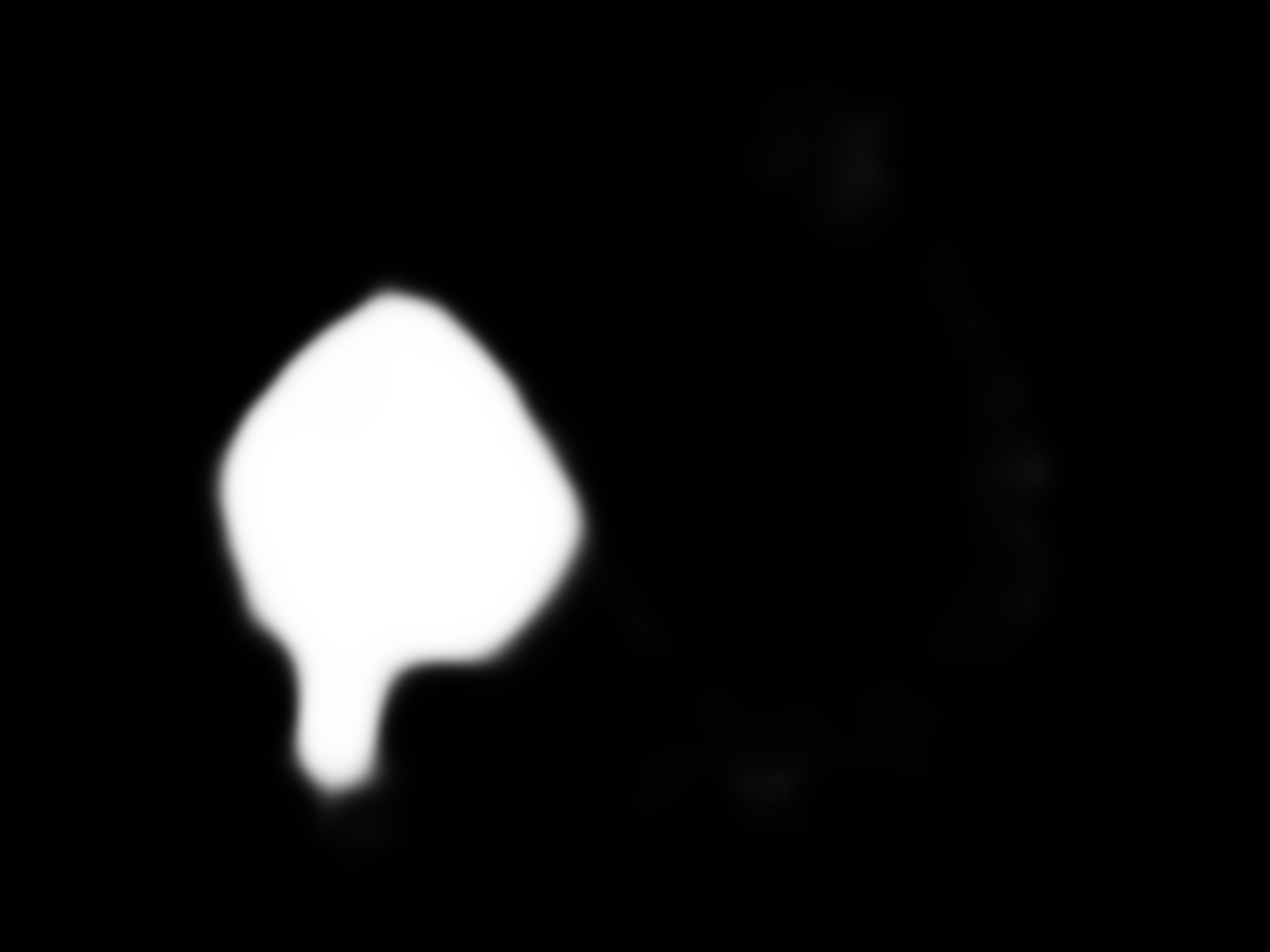}
		\vspace{-5.5mm} \caption{EGNet}
	\end{subfigure}
	\begin{subfigure}{0.105\textwidth}
		\includegraphics[width=\textwidth]{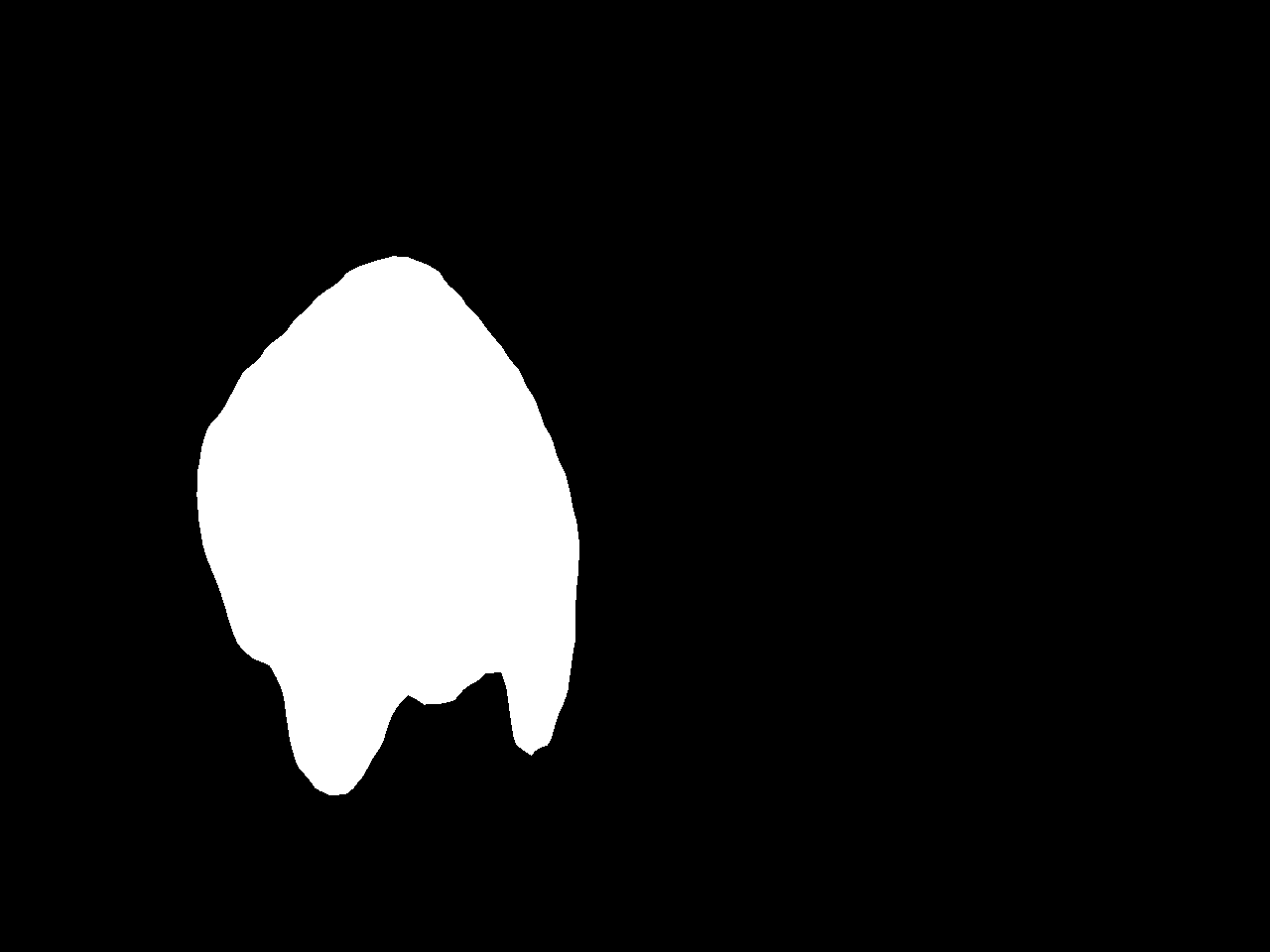}
		\vspace{-5.5mm} \caption{HTC}
	\end{subfigure}
	\begin{subfigure}{0.105\textwidth}
		\includegraphics[width=\textwidth]{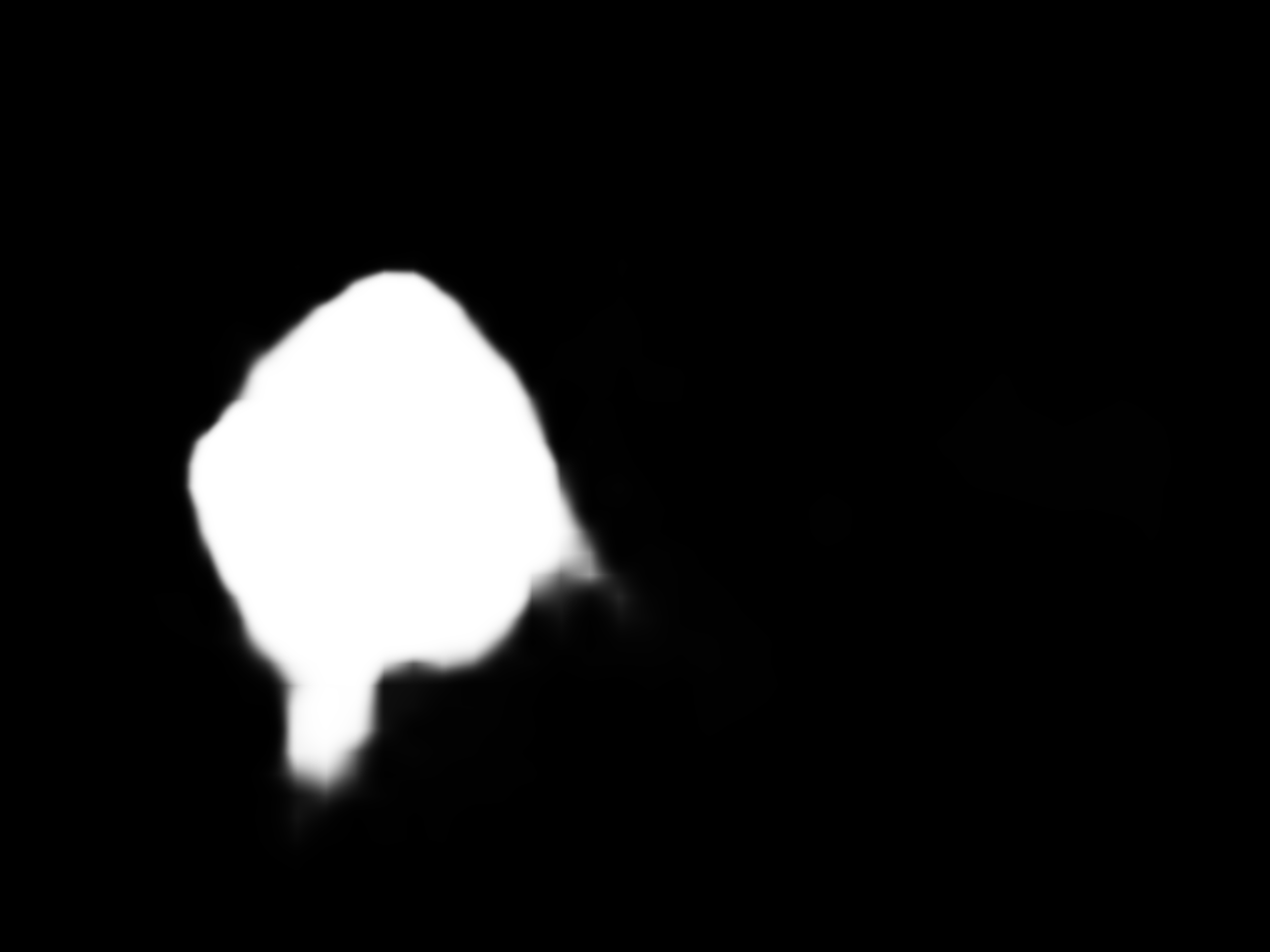}
		\vspace{-5.5mm} \caption{CPD}
	\end{subfigure}
    \begin{subfigure}{0.105\textwidth}
		\includegraphics[width=\textwidth]{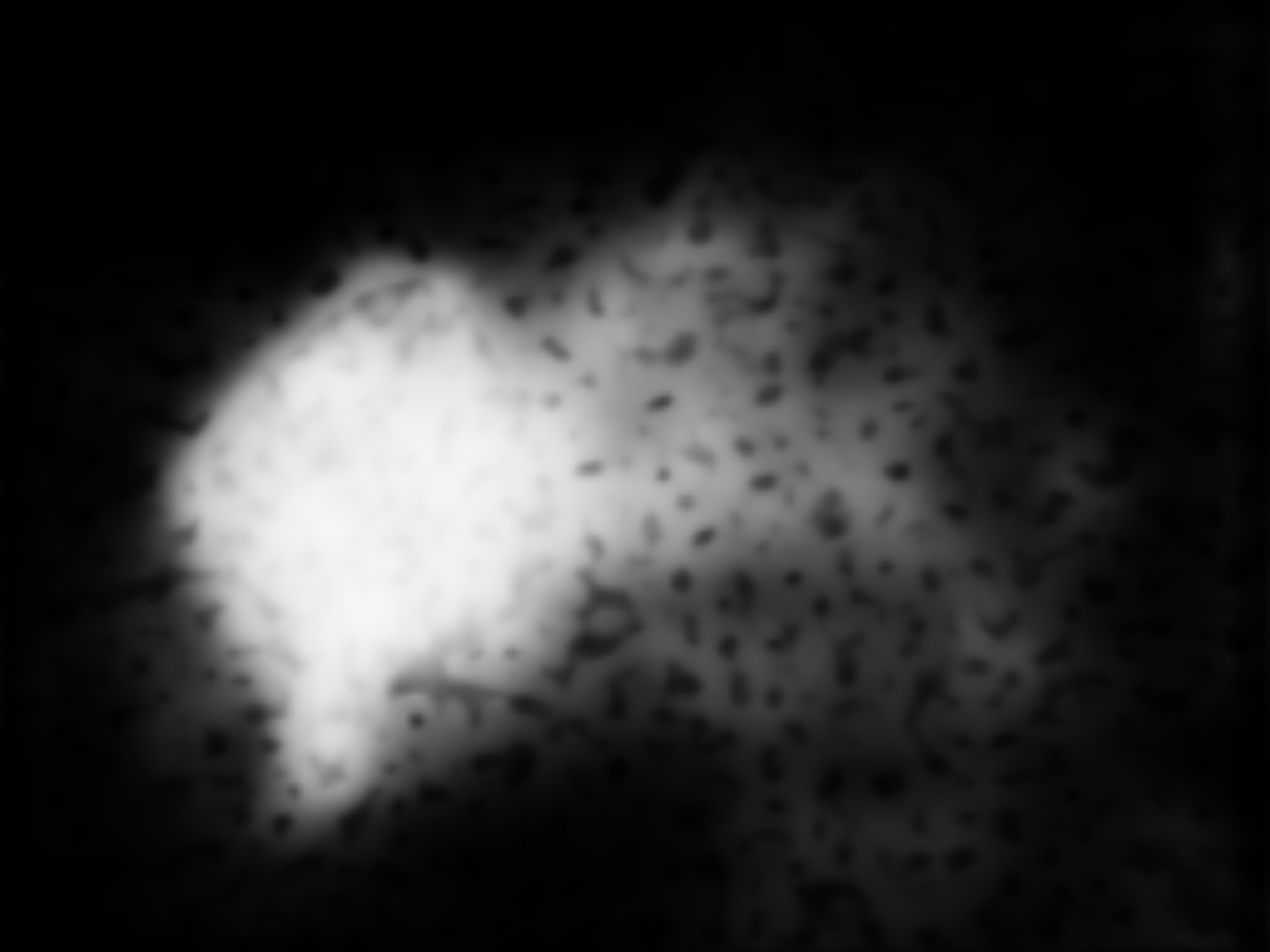}
		\vspace{-5.5mm} \caption{PFANet}
	\end{subfigure}
	\begin{subfigure}{0.105\textwidth}
		\includegraphics[width=\textwidth]{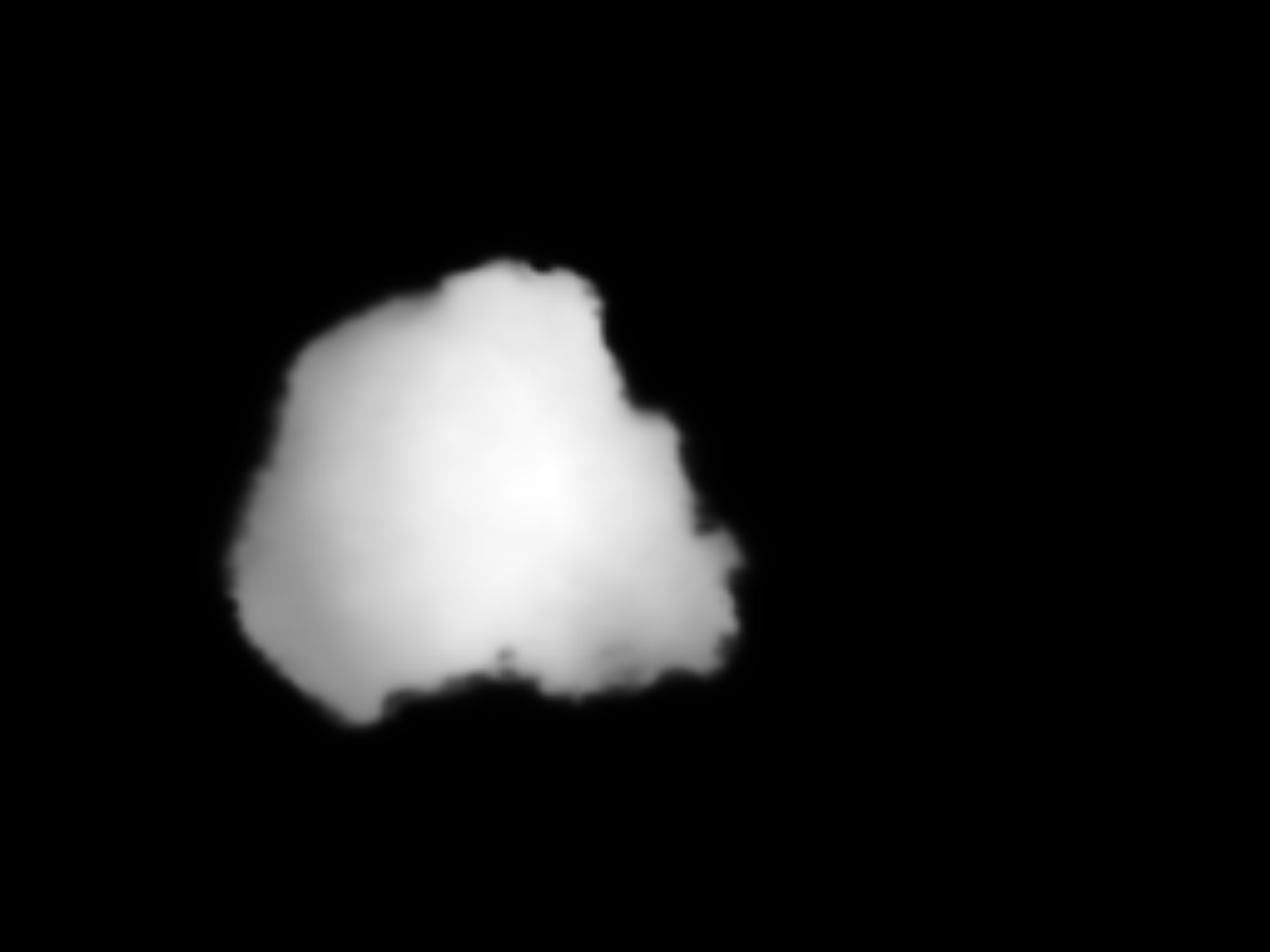}
		\vspace{-5.5mm} \caption{BASNet}
	\end{subfigure}
	\ \\
    \vspace{-1.5mm}
	\caption{Visual comparison of camouflaged object detection maps produced by different methods. (a) input images; (b) ground truths; camouflaged object detection maps produced by (c) our method, (d) SINet~\cite{fan2020camouflaged}, (e) EGNet~\cite{zhao2019egnet}, (f) HTC~\cite{chen2019hybrid}, (g) CPD~\cite{wu2019cascaded}, (h) PFANet~\cite{zhao2019pyramid}, and (i) BASNet~\cite{qin2019basnet}. Apparently, our method can better identify camouflaged objects than all the compared detectors, and our prediction results (c) are more consistent with the ground truth images.}
	\label{fig:comparison_real_photos_part1}
\end{figure*}

\begin{table*}[htbp]
	\begin{center}
        \caption{Ablation study results. Here, we compare the quantitative results of our full pipeline and baseline networks on three benchmark datasets. }
		\label{table:AB}
		\resizebox{\textwidth}{!}{
		\begin{tabular}{c|c |c c c c| c c c c |c c c c}
				\hline
				\hline
				&\multirow{2}{*}{Method} &
				\multicolumn{4}{c|}{CHAMELEON} &  \multicolumn{4}{c|}{CAMO-Test} & \multicolumn{4}{c}{COD10K-Test} \\
				\cline{3-14}
				  & & {$S_\alpha \uparrow$} & {$E_\phi \uparrow$} &{$F_{\beta}^w \uparrow$} & {M $\downarrow$} & 	{$S_\alpha \uparrow$} & {$E_\phi \uparrow$}
				 &{$F_{\beta}^w \uparrow$} & {M $\downarrow$} & {$S_\alpha \uparrow$} & {$E_\phi \uparrow$} &{$F_{\beta}^w \uparrow$} & {M $\downarrow$}\\
			
				 \hline
			   $M_1$ & basic \cite{lin2017feature} & 0.856 & 0.866 & 0.710 & 0.050 & 0.763 & 0.783 & 0.621 & 0.097 & 0.772 & 0.797 & 0.557 & 0.049     \\ 
				 \hline
			    $M_2$ & basic+RRB & 0.862 & 0.885 & 0.733 & 0.046 & 0.774 & 0.807 & 0.649 & 0.091 & 0.782 & 0.813 & 0.583 & 0.046     \\ 
			    \hline
				 $M_3$ & basic+RRB+TARM w/o BCL & 0.879 & 0.909 & 0.767 & 0.039 & 0.782 & 0.820 & 0.675 & 0.087 & 0.796 & 0.840 & 0.618 & 0.042 \\ %
				 \hline
				Ours & TANet & \textbf{0.888} & \textbf{0.911} & \textbf{0.786} & \textbf{0.036} & \textbf{0.793} & \textbf{0.834} & \textbf{0.690} &  \textbf{0.083} & \textbf{0.803}  & \textbf{0.848} & \textbf{0.629} & \textbf{0.041} 	\\
				\hline
				\hline
		\end{tabular}
		}
	    \vspace{1mm}
	\end{center}
    \vspace{-8mm}

\end{table*}

We employ three widely-used COD benchmark datasets to test each COD method.
COD10K~\cite{fan2020camouflaged} is the largest annotated dataset with $3,040$ training images and $2,026$ testing images.
CAMO~\cite{le2019anabranch} includes $1,000$ training images and $250$ testing images while CHAMELEON~\cite{skurowski2018animal} consists of $76$ images.
The same training dataset used by the most recent COD method~\cite{fan2020camouflaged} is employed to train our network for fair comparisons. The training set consists of $3,040$ training images of COD10K (3,040 images), $600$ training images of CPD1K, and $1,000$ training images of CAMO.
We test different COD methods on the testing set (COD10K-Test) of COD10K, the testing set (CAMO-Test) of CAMO, and the whole CHAMELEON for their results.
{\em We shall release our code, the trained network model, and the predicted COD maps of our method upon the publication of this work.}

\subsection{Evaluation Metrics}
To conduct quantitative comparisons, we employ four common metrics to test different COD methods. They are  Structure-measure~\cite{fan2017structure} ($S_{\alpha}$), Enhanced-measure~\cite{fan2018enhanced} ($E_{\phi}$),  weight-F-measure~\cite{margolin2014evaluate}~($F_{\beta}^{w}$), and MAE~\cite{perazzi2012saliency}; see SINet \cite{fan2020camouflaged} for the definitions of the four evaluation metrics.
Overall, a better COD method has larger $S_{\alpha}$, $E_{\phi}$, and $E_{\phi}$ scores, but a smaller MAE score.

\begin{figure*}[t]
	\centering
		\vspace*{0.5mm}
	\begin{subfigure}{0.12\textwidth}
		\includegraphics[width=\textwidth]{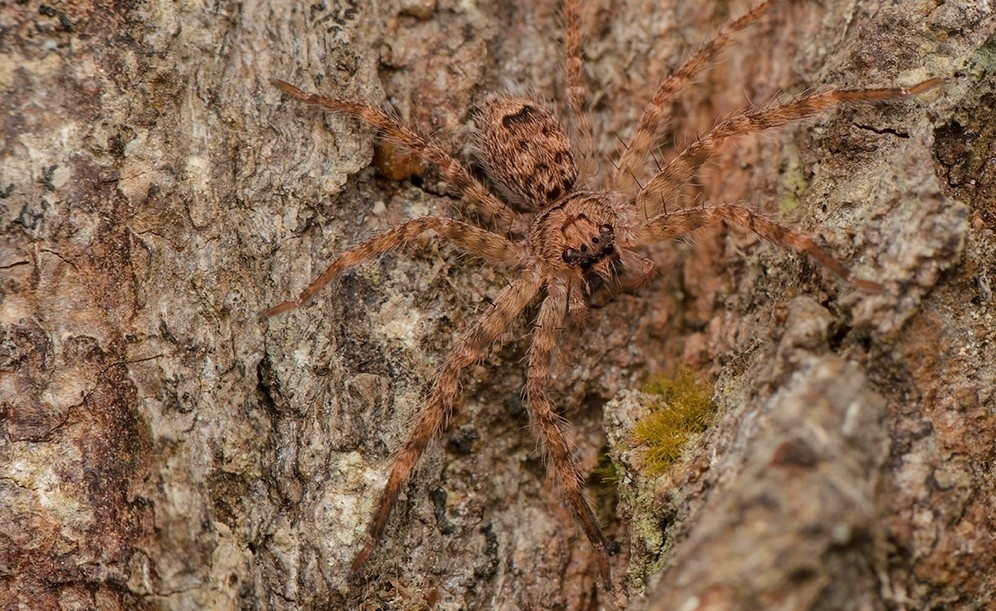}
	\end{subfigure}
	\begin{subfigure}{0.12\textwidth}
		\includegraphics[width=\textwidth]{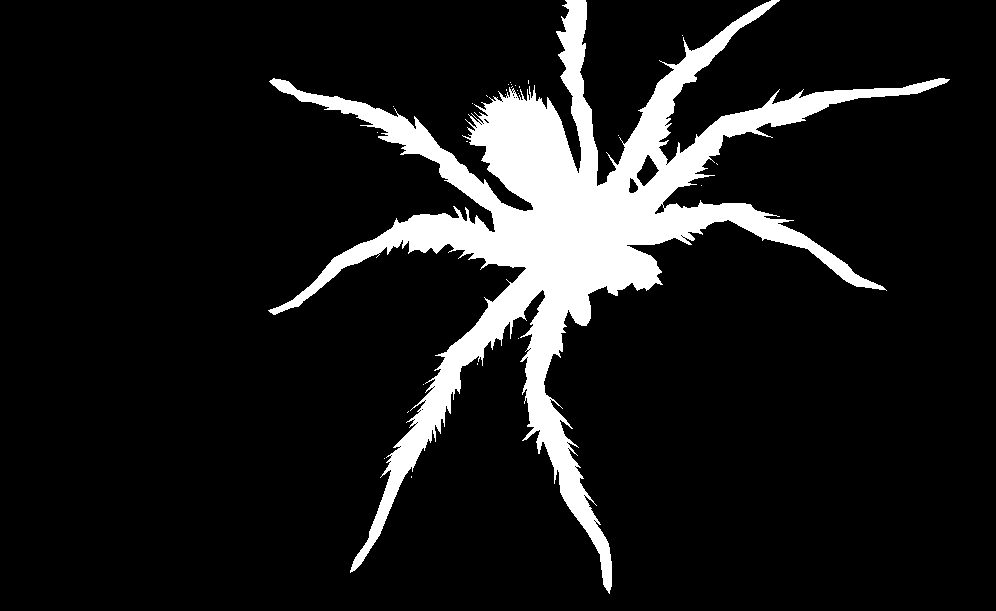}
	\end{subfigure}
	\begin{subfigure}{0.12\textwidth}
		\includegraphics[width=\textwidth]{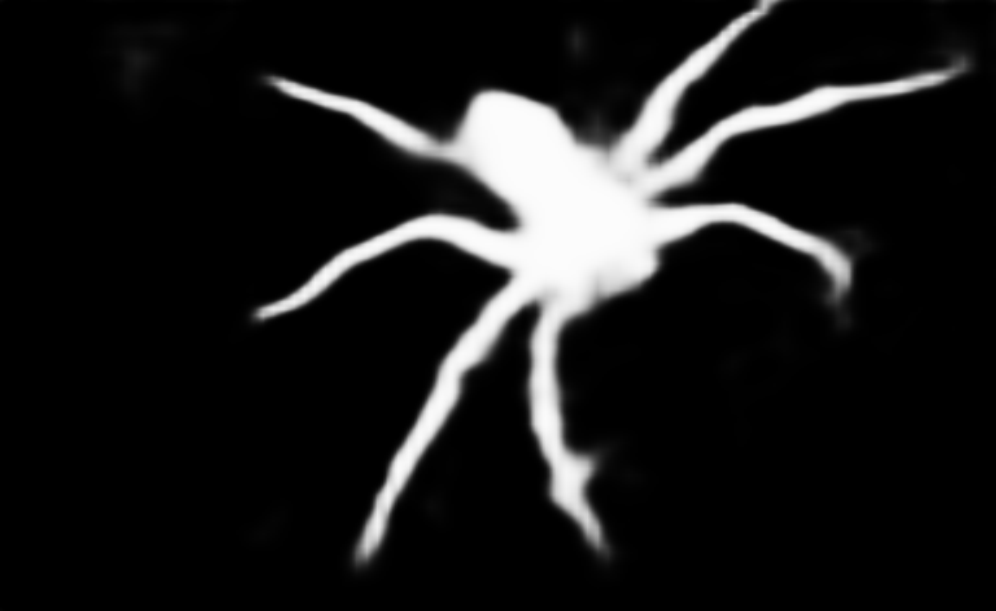}
	\end{subfigure}
    \begin{subfigure}{0.12\textwidth}
		\includegraphics[width=\textwidth]{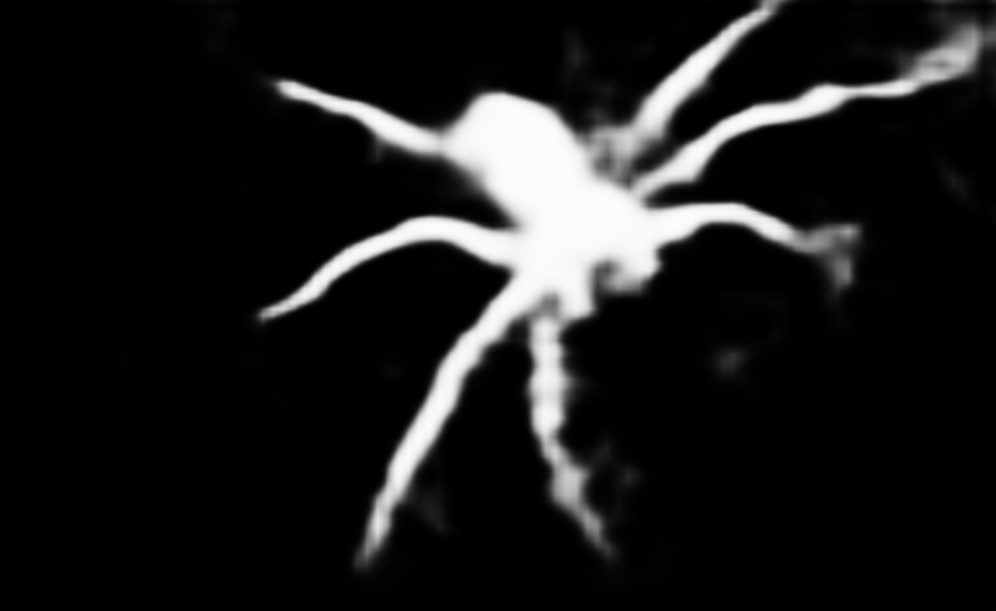}
	\end{subfigure}
    \begin{subfigure}{0.12\textwidth}
		\includegraphics[width=\textwidth]{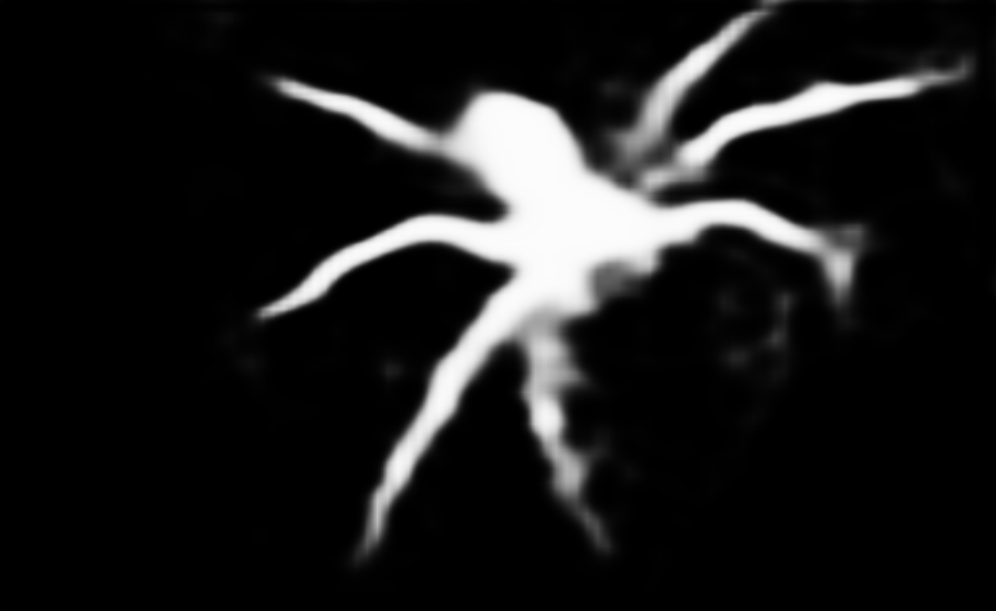}
	\end{subfigure}
    \begin{subfigure}{0.12\textwidth}
		\includegraphics[width=\textwidth]{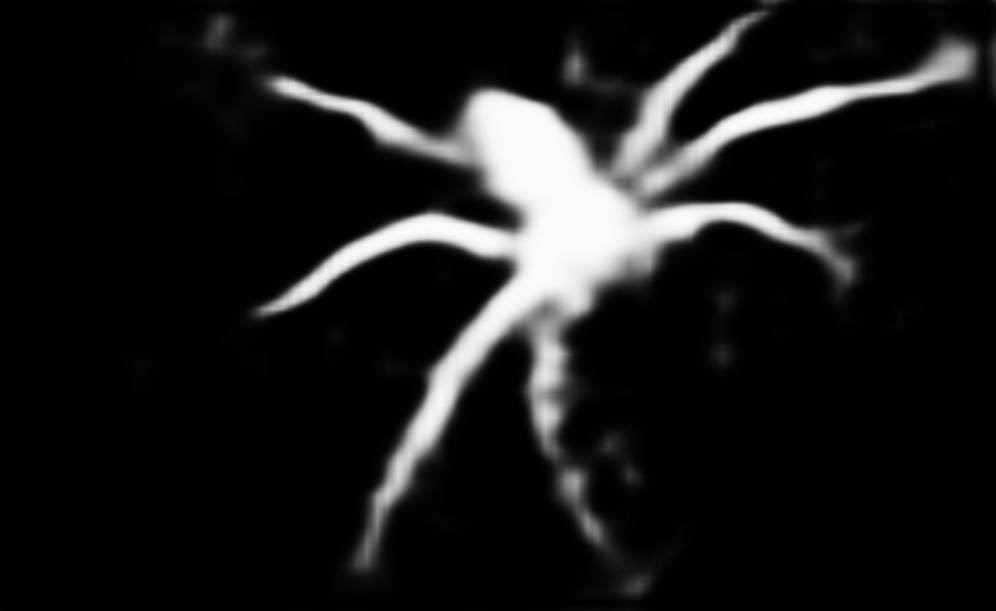}
	\end{subfigure}
    \begin{subfigure}{0.12\textwidth}
		\includegraphics[width=\textwidth]{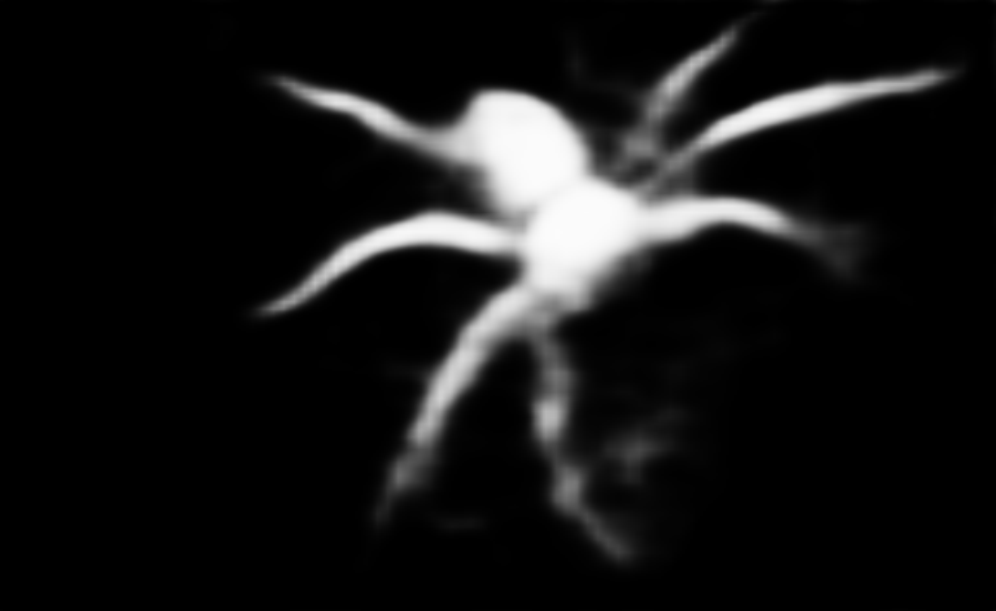}
	\end{subfigure}
	\ \\
	\vspace*{0.5mm}
	\begin{subfigure}{0.12\textwidth}
		\includegraphics[width=\textwidth]{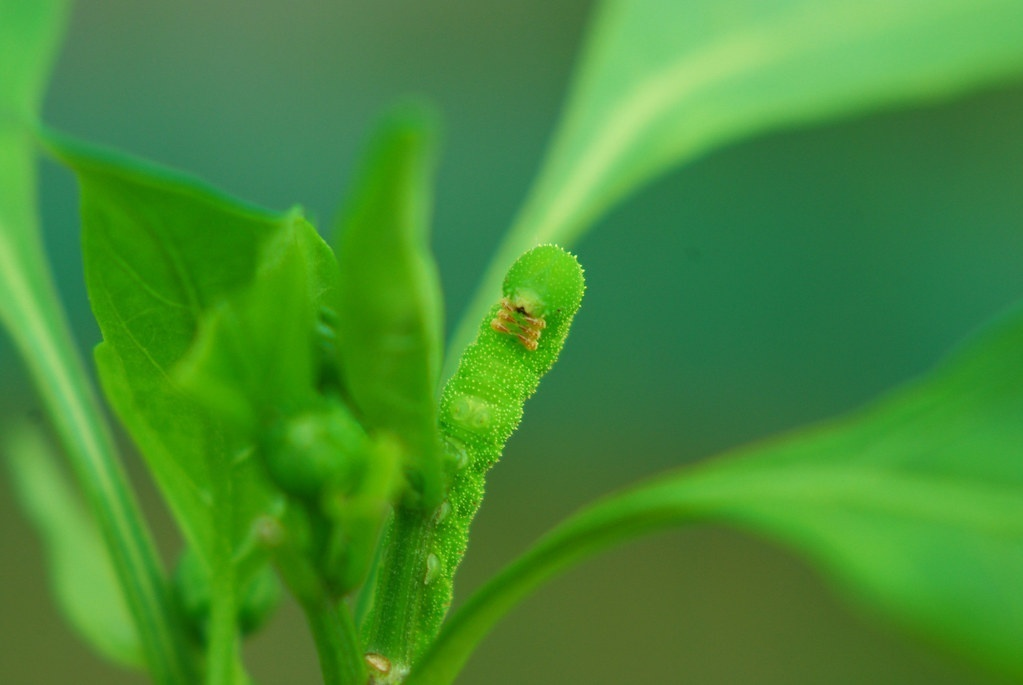}
	\end{subfigure}
	\begin{subfigure}{0.12\textwidth}
		\includegraphics[width=\textwidth]{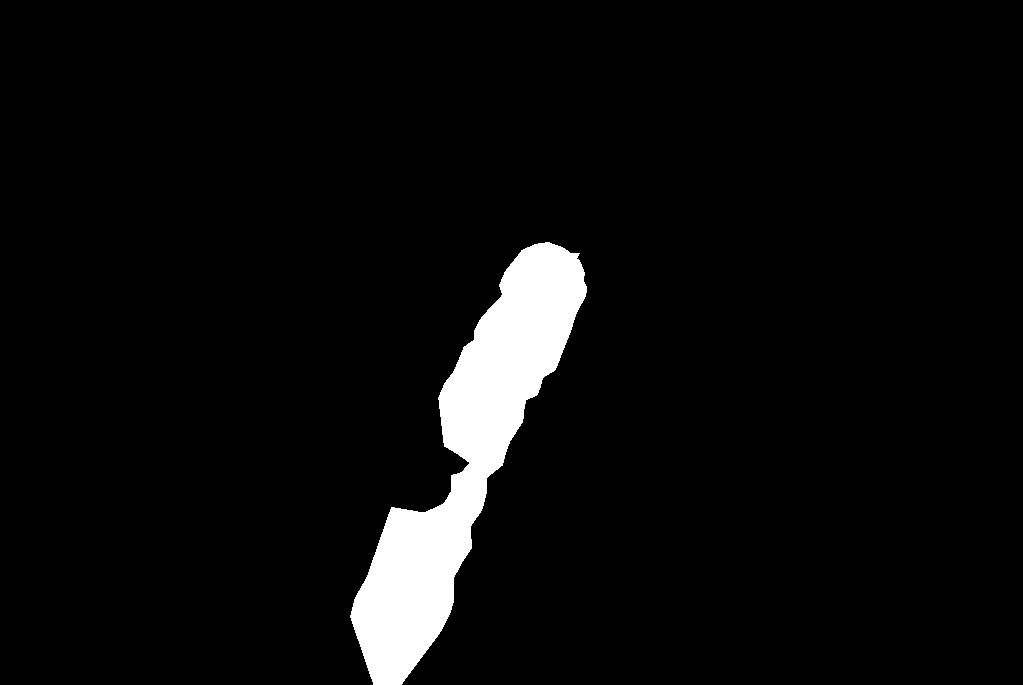}
	\end{subfigure}
	\begin{subfigure}{0.12\textwidth}
		\includegraphics[width=\textwidth]{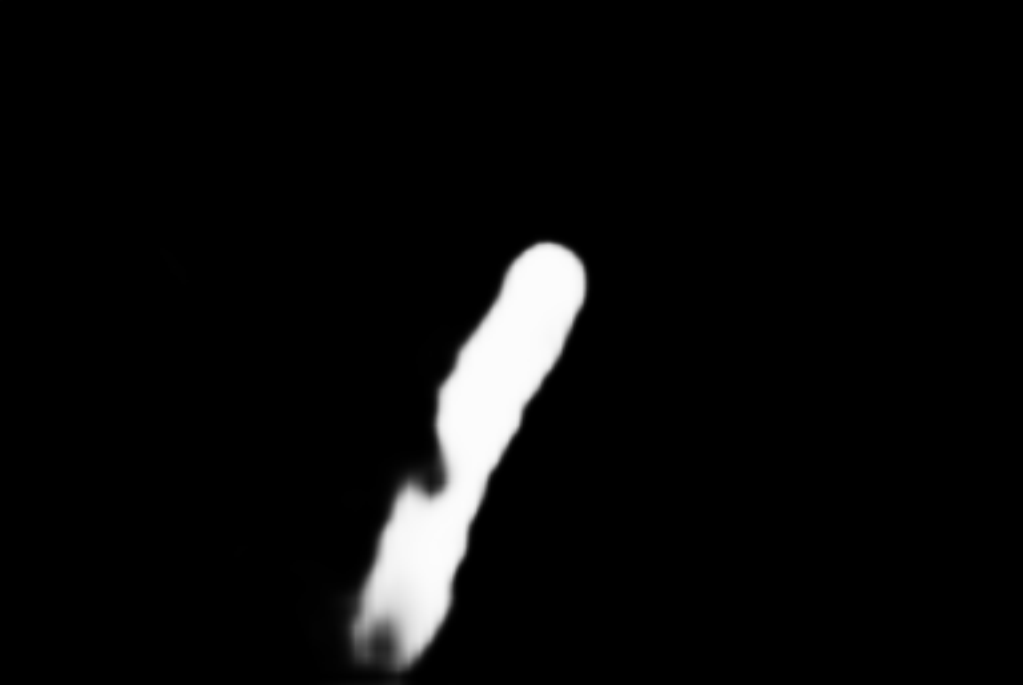}
	\end{subfigure}
    \begin{subfigure}{0.12\textwidth}
		\includegraphics[width=\textwidth]{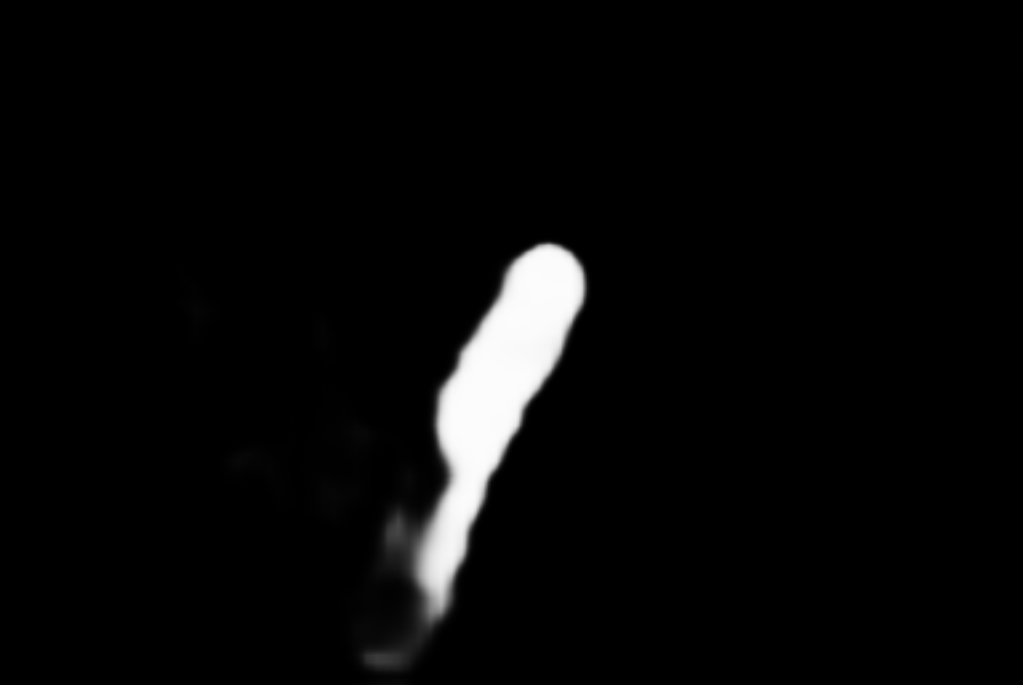}
	\end{subfigure}
    \begin{subfigure}{0.12\textwidth}
		\includegraphics[width=\textwidth]{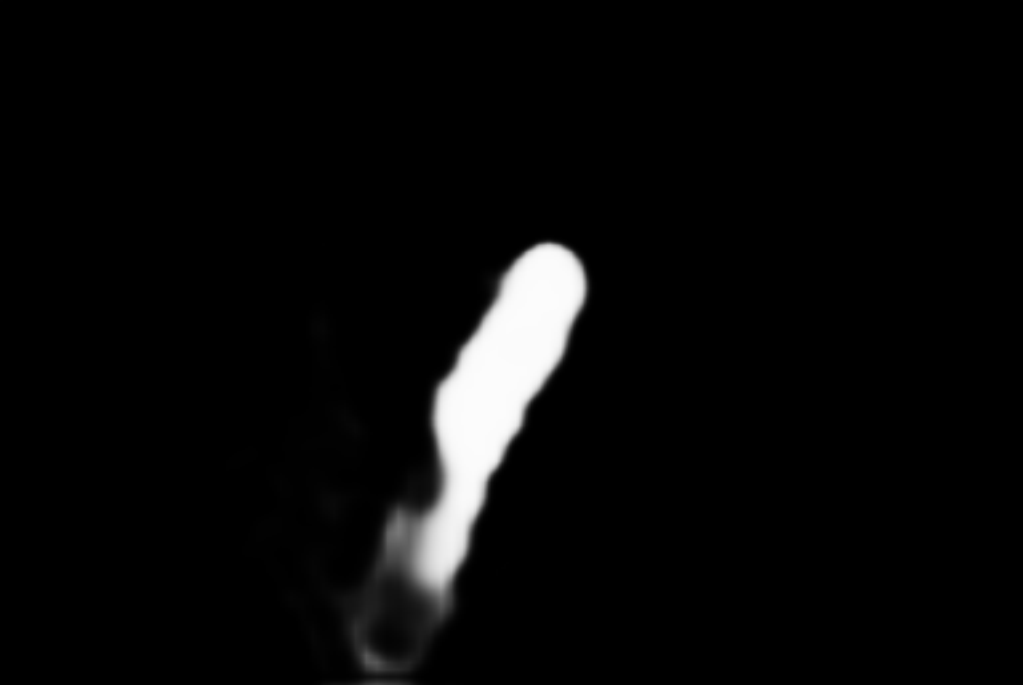}
	\end{subfigure}
    \begin{subfigure}{0.12\textwidth}
		\includegraphics[width=\textwidth]{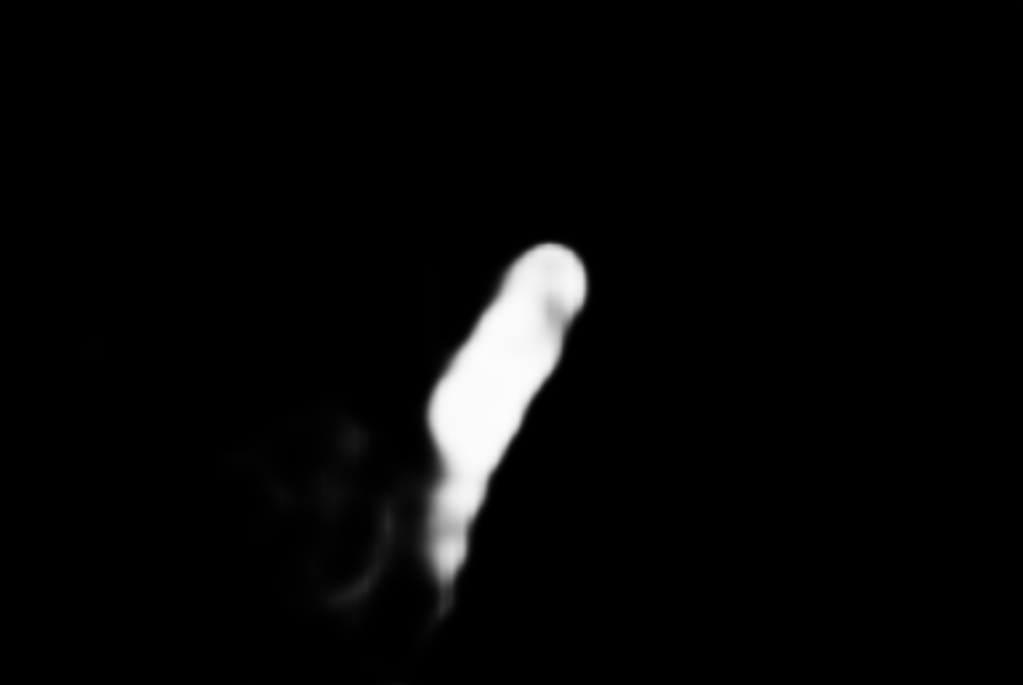}
	\end{subfigure}
    \begin{subfigure}{0.12\textwidth}
		\includegraphics[width=\textwidth]{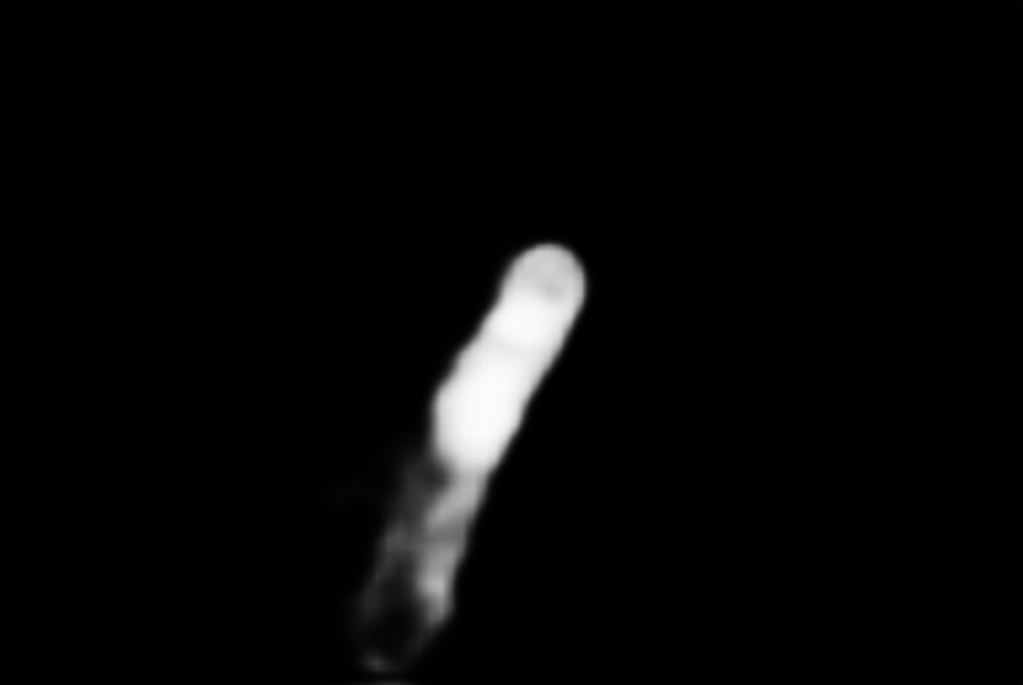}
	\end{subfigure}
	\ \\
	\vspace*{0.5mm}
	\begin{subfigure}{0.12\textwidth}
		\includegraphics[width=\textwidth]{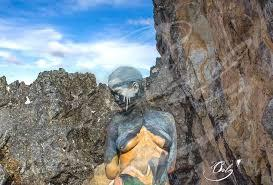}
	\end{subfigure}
	\begin{subfigure}{0.12\textwidth}
		\includegraphics[width=\textwidth]{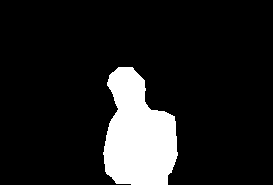}
	\end{subfigure}
	\begin{subfigure}{0.12\textwidth}
		\includegraphics[width=\textwidth]{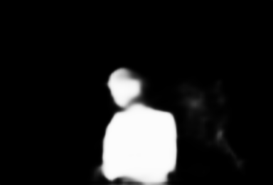}
	\end{subfigure}
    \begin{subfigure}{0.12\textwidth}
		\includegraphics[width=\textwidth]{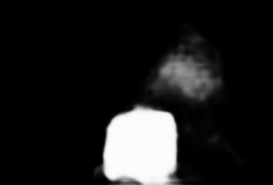}
	\end{subfigure}
    \begin{subfigure}{0.12\textwidth}
		\includegraphics[width=\textwidth]{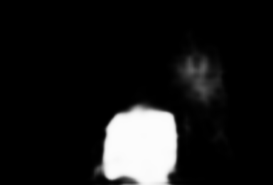}
	\end{subfigure}
	\begin{subfigure}{0.12\textwidth}
		\includegraphics[width=\textwidth]{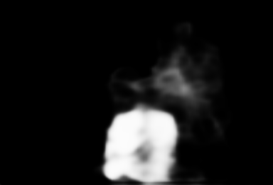}
	\end{subfigure}
    \begin{subfigure}{0.12\textwidth}
		\includegraphics[width=\textwidth]{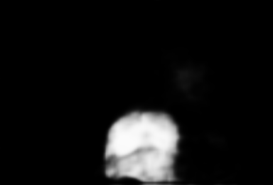}
	\end{subfigure}
	\ \\
	\vspace*{0.5mm}
	\begin{subfigure}{0.12\textwidth} \includegraphics[width=\textwidth]{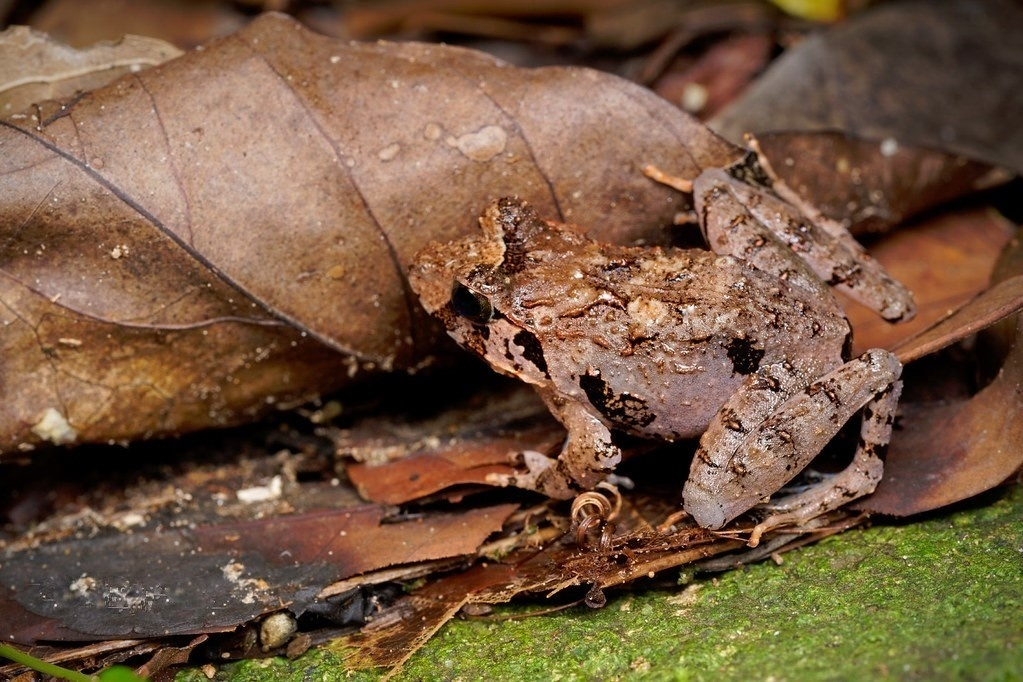}
	\end{subfigure}
	\begin{subfigure}{0.12\textwidth} \includegraphics[width=\textwidth]{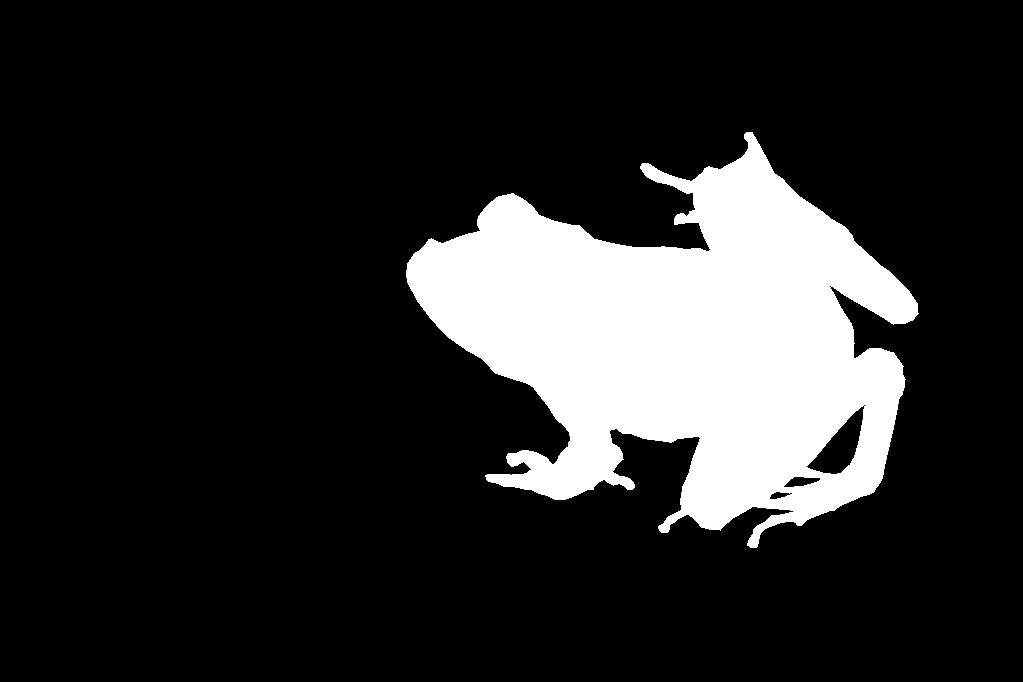}
	\end{subfigure}
	\begin{subfigure}{0.12\textwidth} \includegraphics[width=\textwidth]{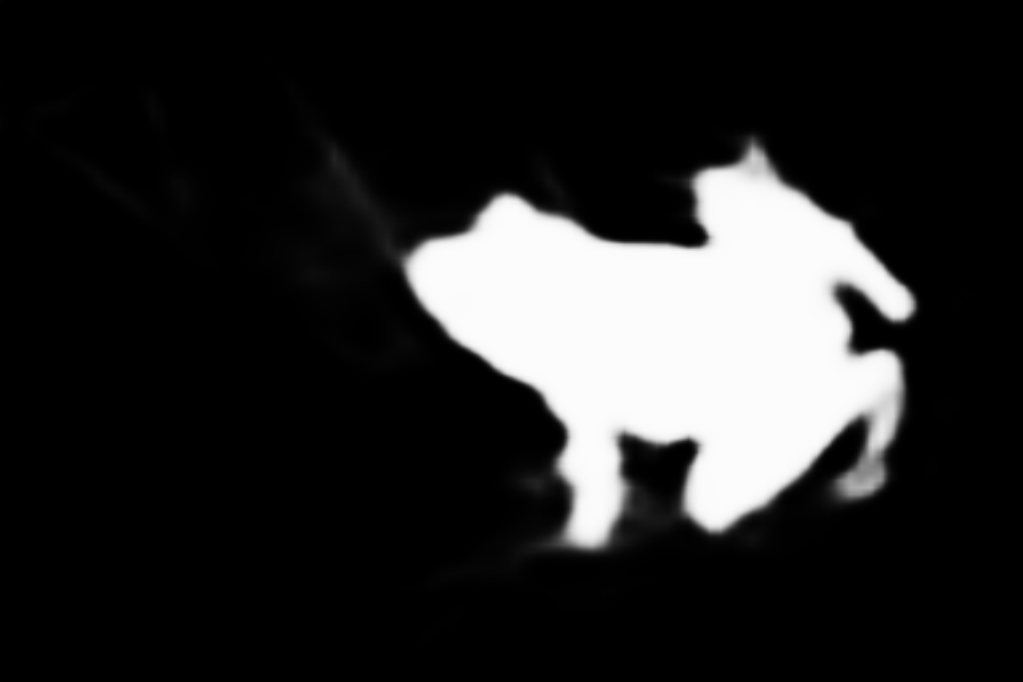}
	\end{subfigure}
    \begin{subfigure}{0.12\textwidth}	\includegraphics[width=\textwidth]{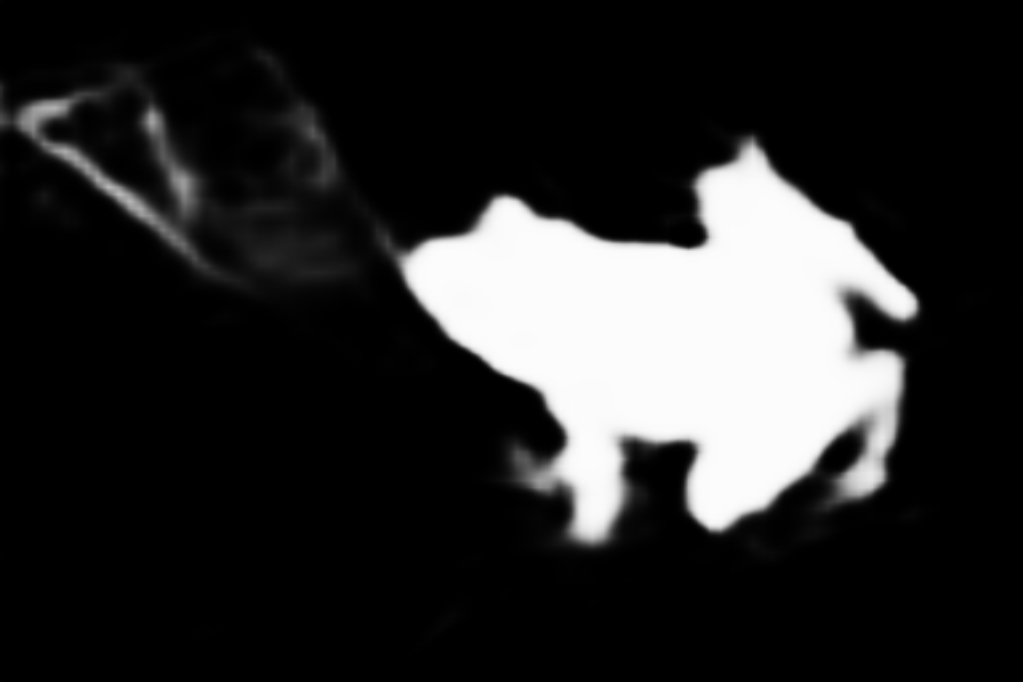}
	\end{subfigure}
    \begin{subfigure}{0.12\textwidth} \includegraphics[width=\textwidth]{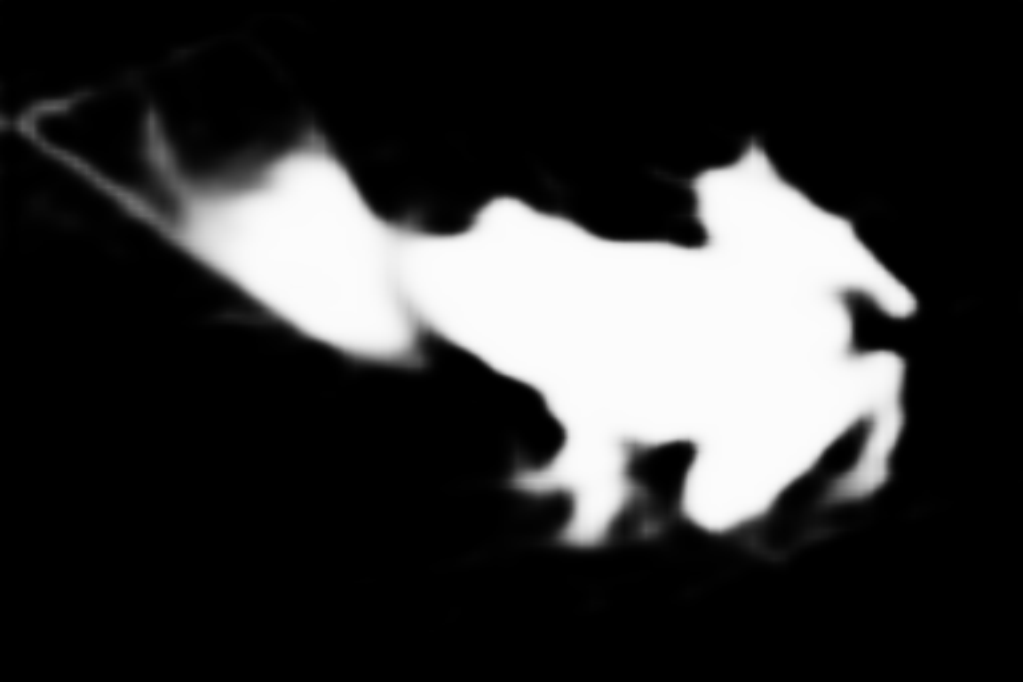}
	\end{subfigure}
	\begin{subfigure}{0.12\textwidth} \includegraphics[width=\textwidth]{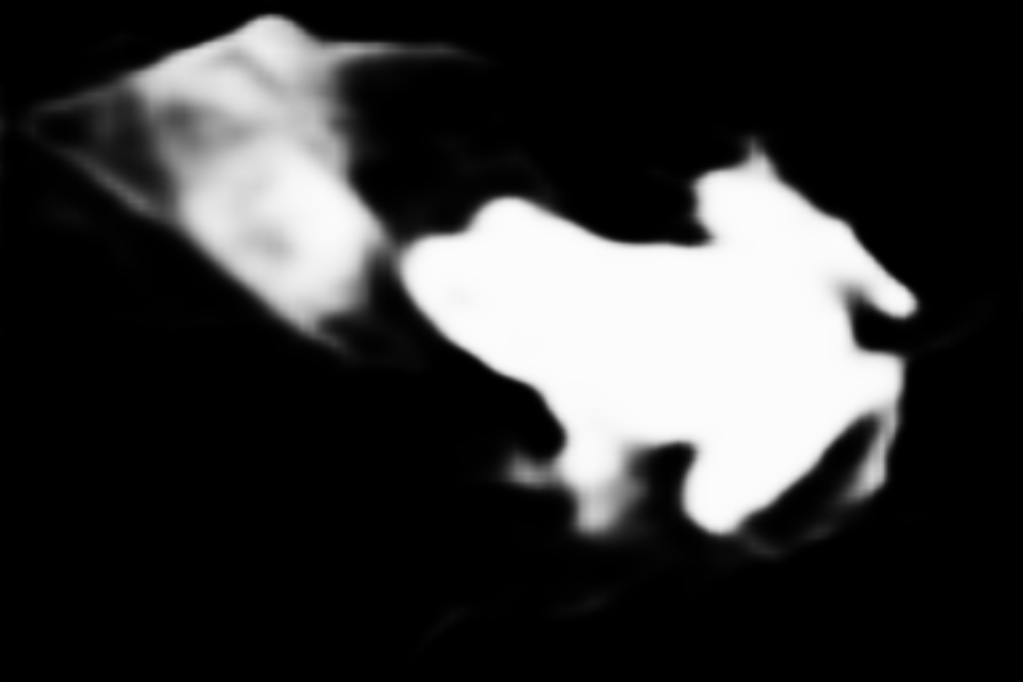}
	\end{subfigure}
	\begin{subfigure}{0.12\textwidth} \includegraphics[width=\textwidth]{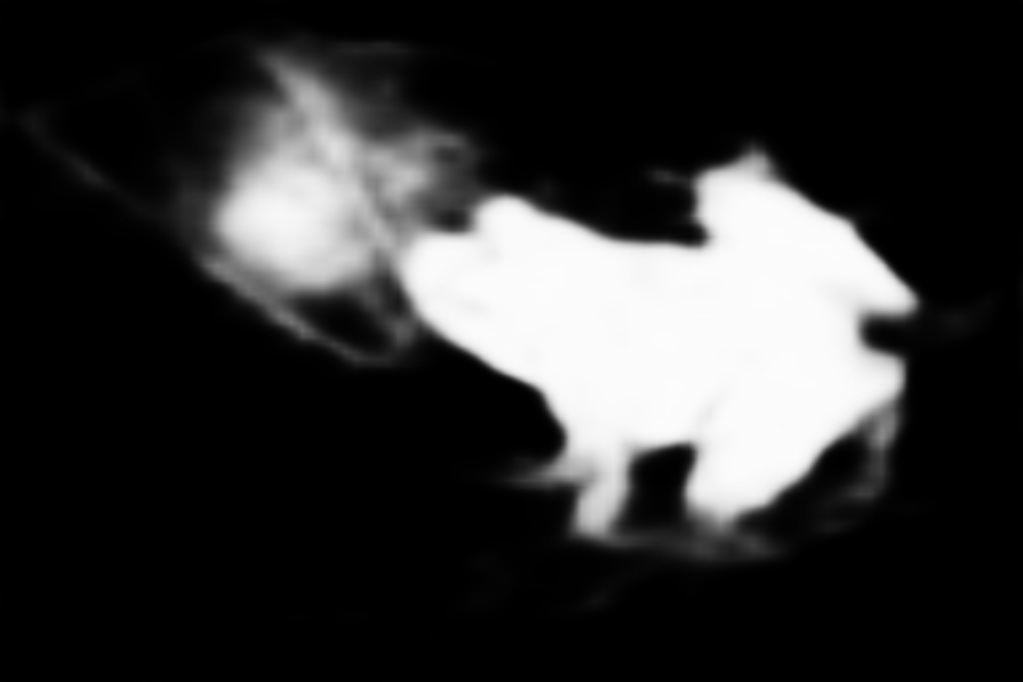}
	\end{subfigure}
    \ \\
    \vspace*{0.5mm}
	\begin{subfigure}{0.12\textwidth} \includegraphics[width=\textwidth]{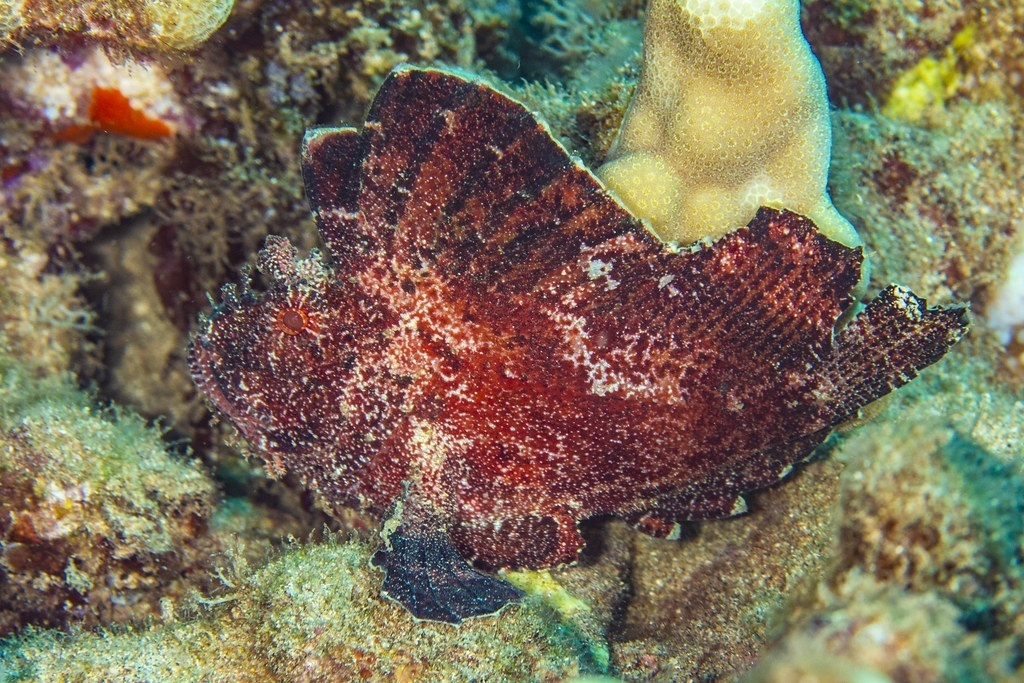}
	\end{subfigure}
	\begin{subfigure}{0.12\textwidth} \includegraphics[width=\textwidth]{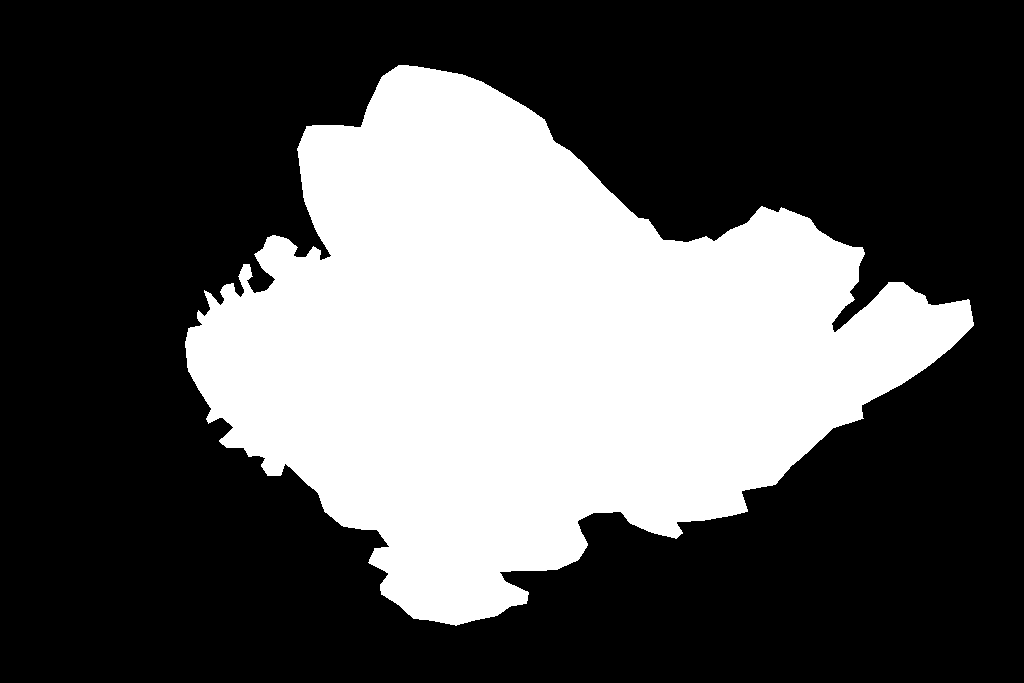}
	\end{subfigure}
	\begin{subfigure}{0.12\textwidth} \includegraphics[width=\textwidth]{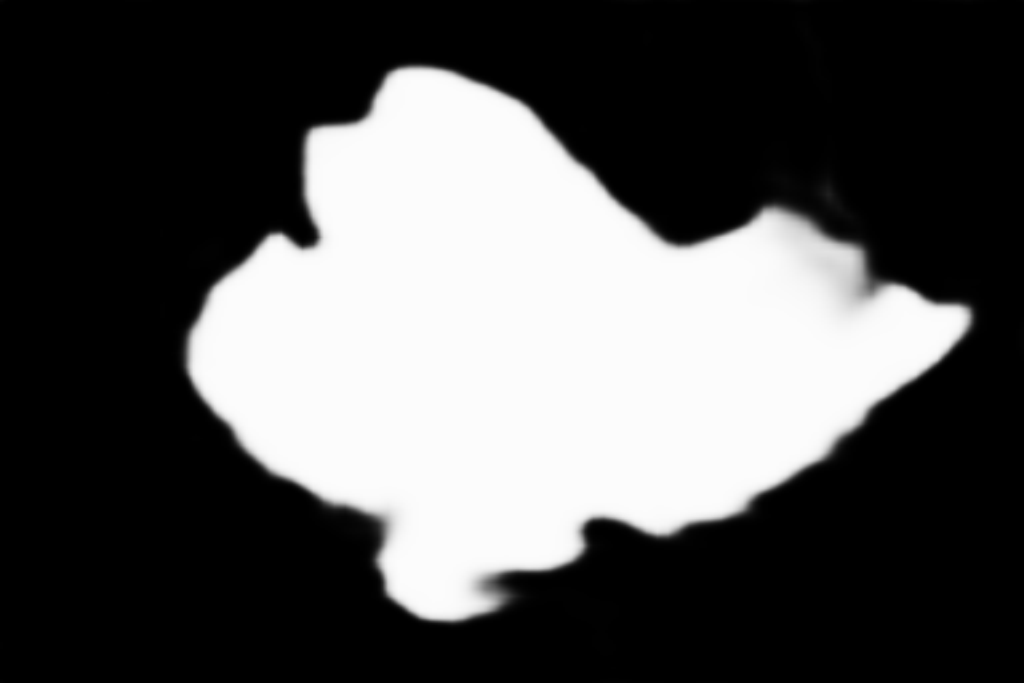}
	\end{subfigure}
    \begin{subfigure}{0.12\textwidth}	\includegraphics[width=\textwidth]{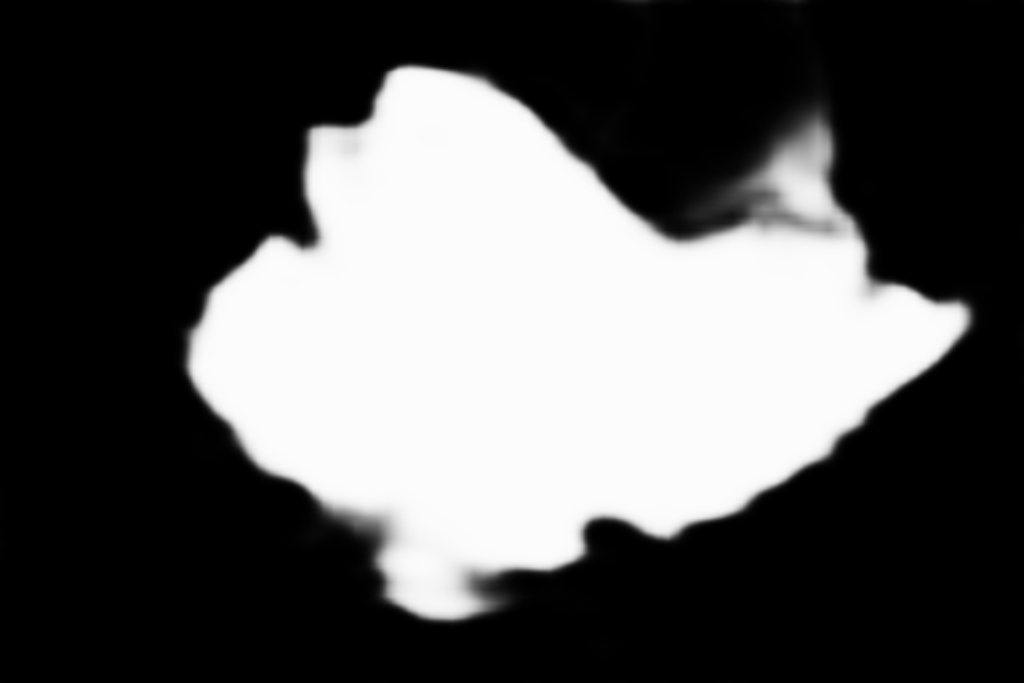}
	\end{subfigure}
    \begin{subfigure}{0.12\textwidth} \includegraphics[width=\textwidth]{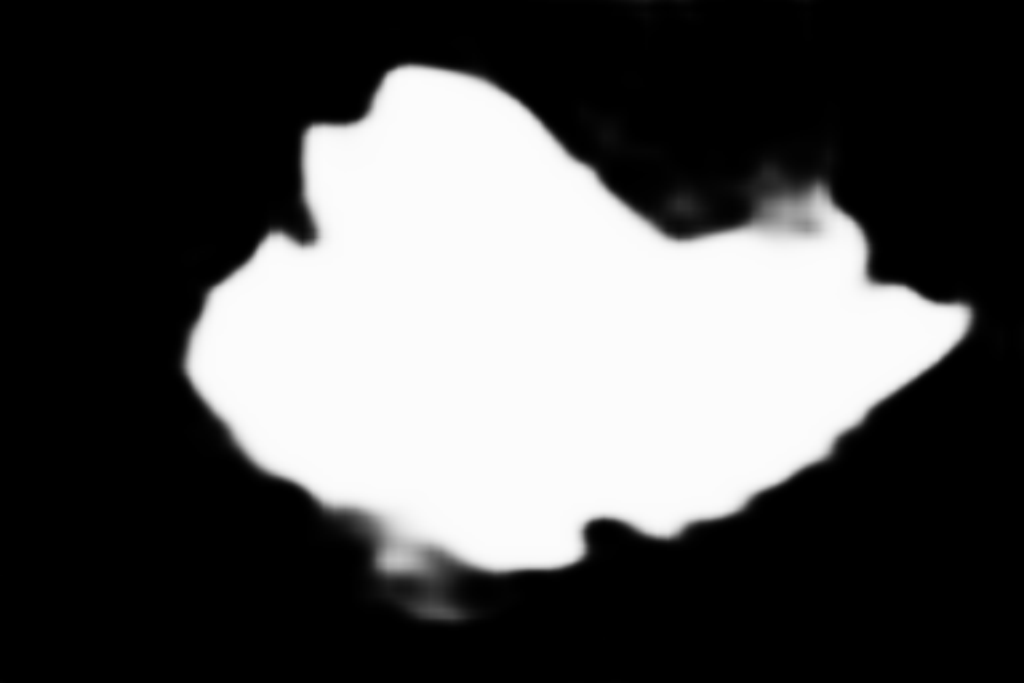}
	\end{subfigure}
	\begin{subfigure}{0.12\textwidth} \includegraphics[width=\textwidth]{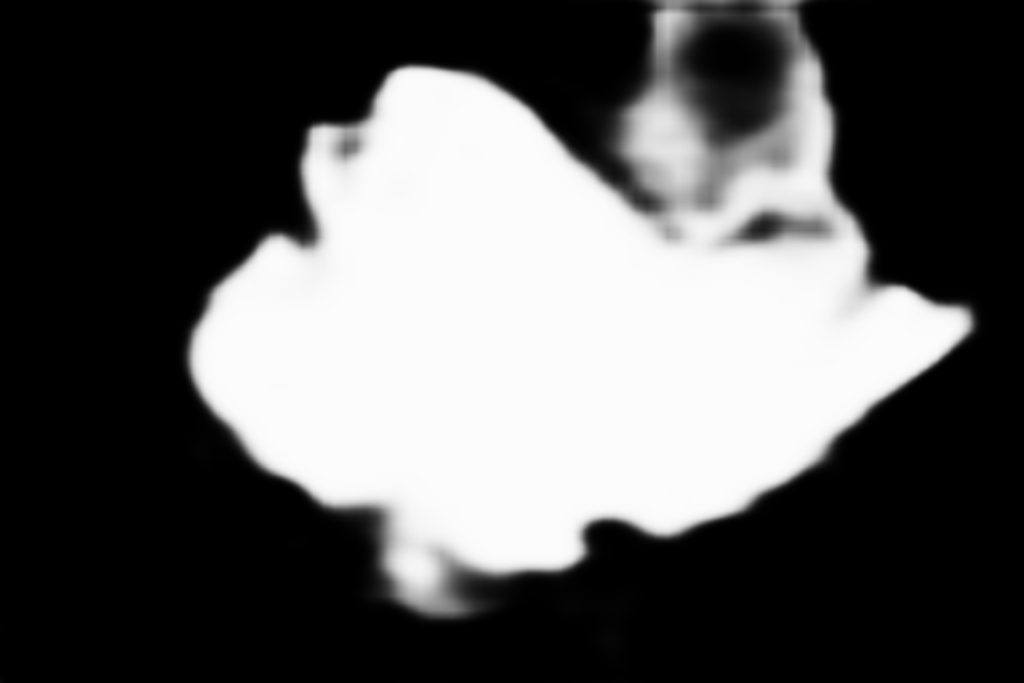}
	\end{subfigure}
	\begin{subfigure}{0.12\textwidth} \includegraphics[width=\textwidth]{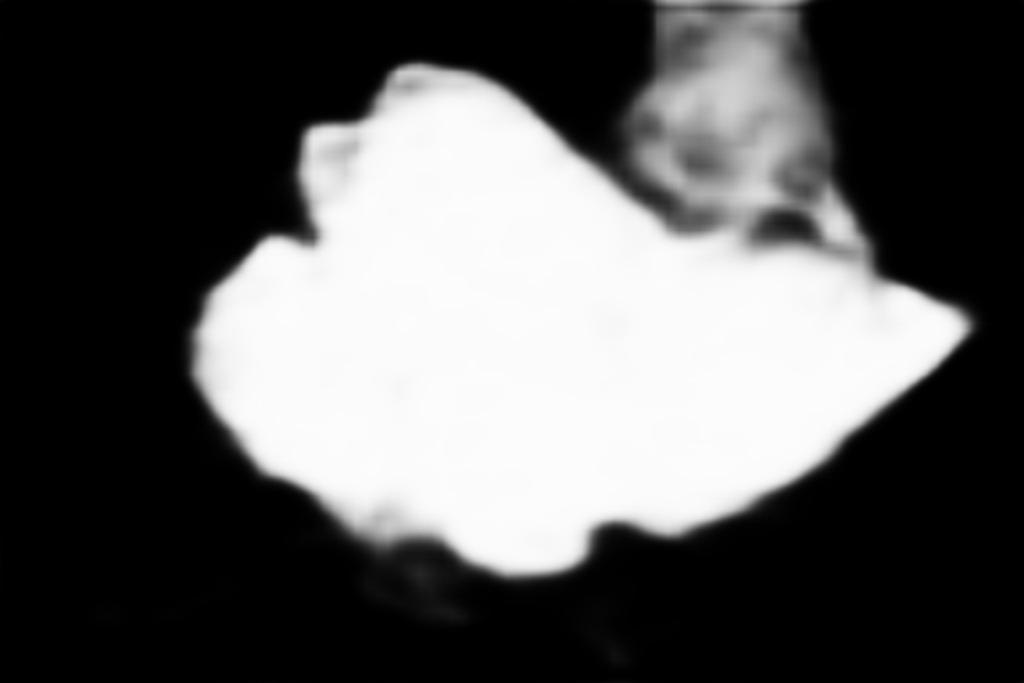}
	\end{subfigure}
    \ \\
     \vspace*{0.5mm}
	\begin{subfigure}{0.12\textwidth} \includegraphics[width=\textwidth]{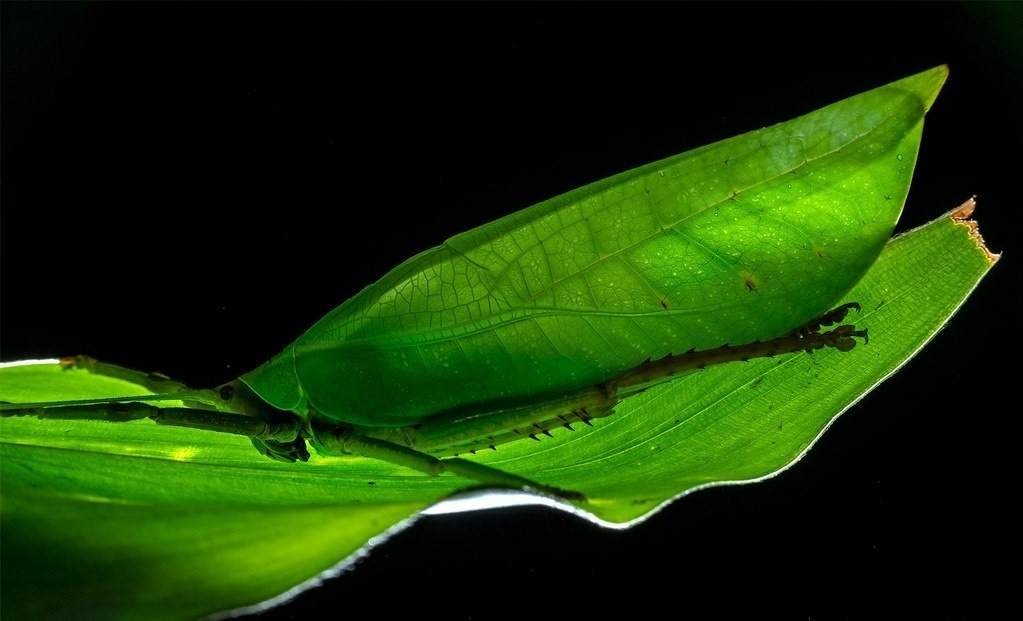}
	\end{subfigure}
	\begin{subfigure}{0.12\textwidth} \includegraphics[width=\textwidth]{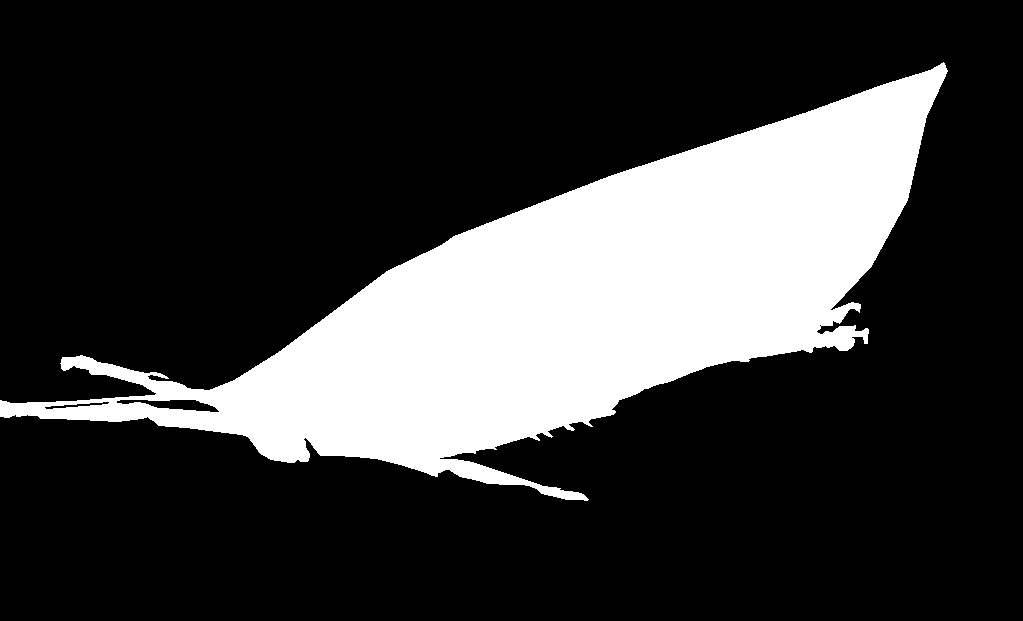}
	\end{subfigure}
	\begin{subfigure}{0.12\textwidth} \includegraphics[width=\textwidth]{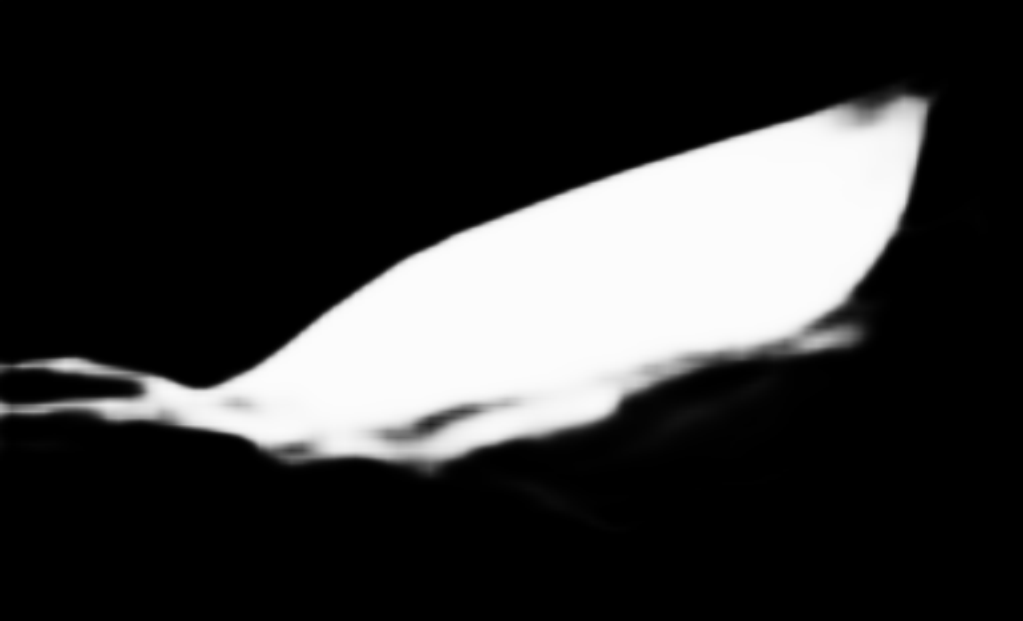}
	\end{subfigure}
    \begin{subfigure}{0.12\textwidth}	\includegraphics[width=\textwidth]{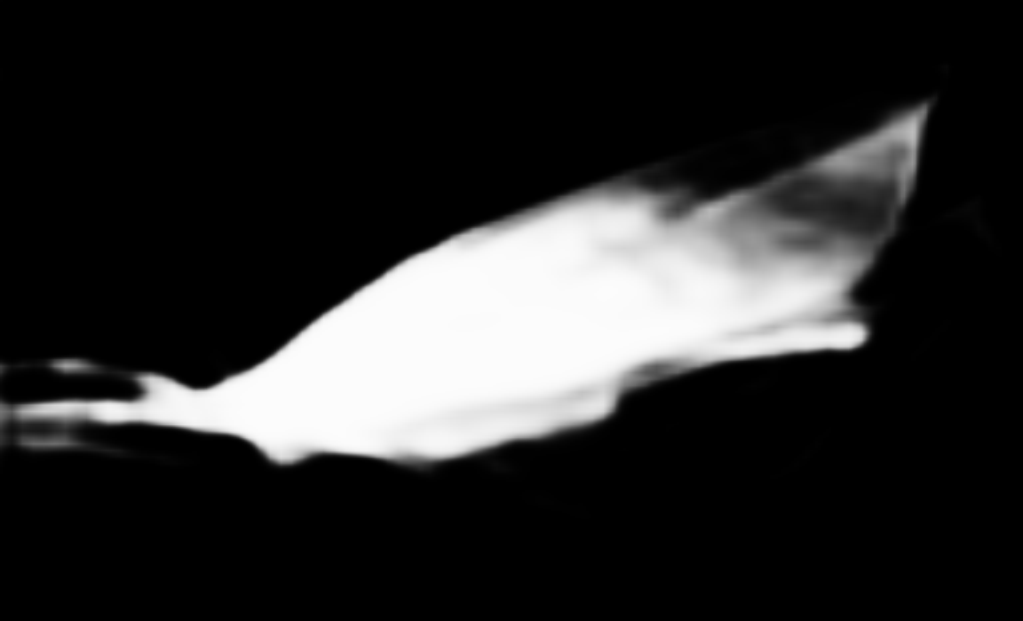}
	\end{subfigure}
    \begin{subfigure}{0.12\textwidth} \includegraphics[width=\textwidth]{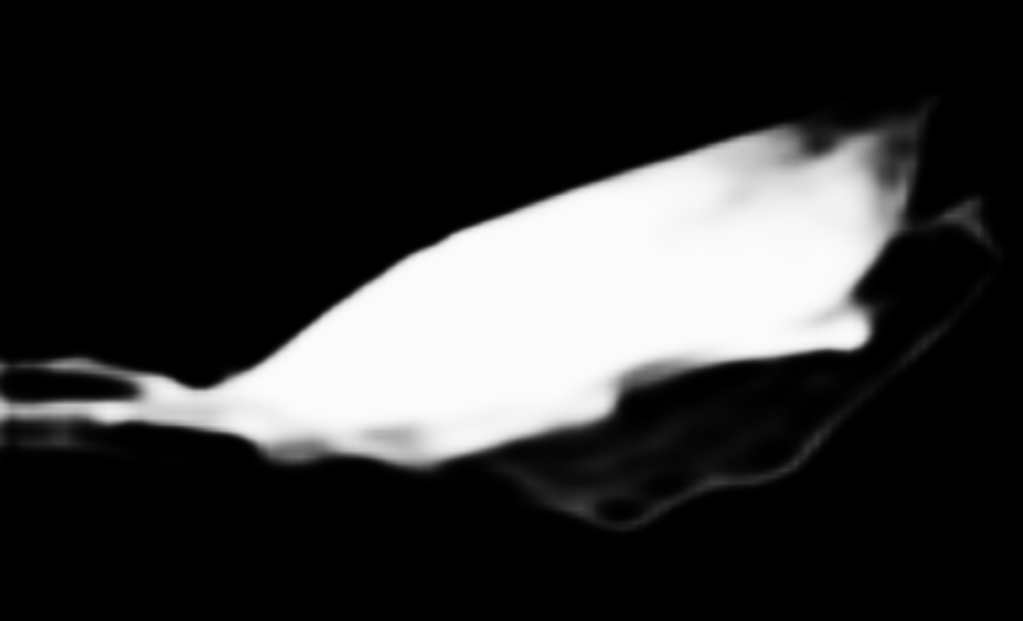}
	\end{subfigure}
	\begin{subfigure}{0.12\textwidth} \includegraphics[width=\textwidth]{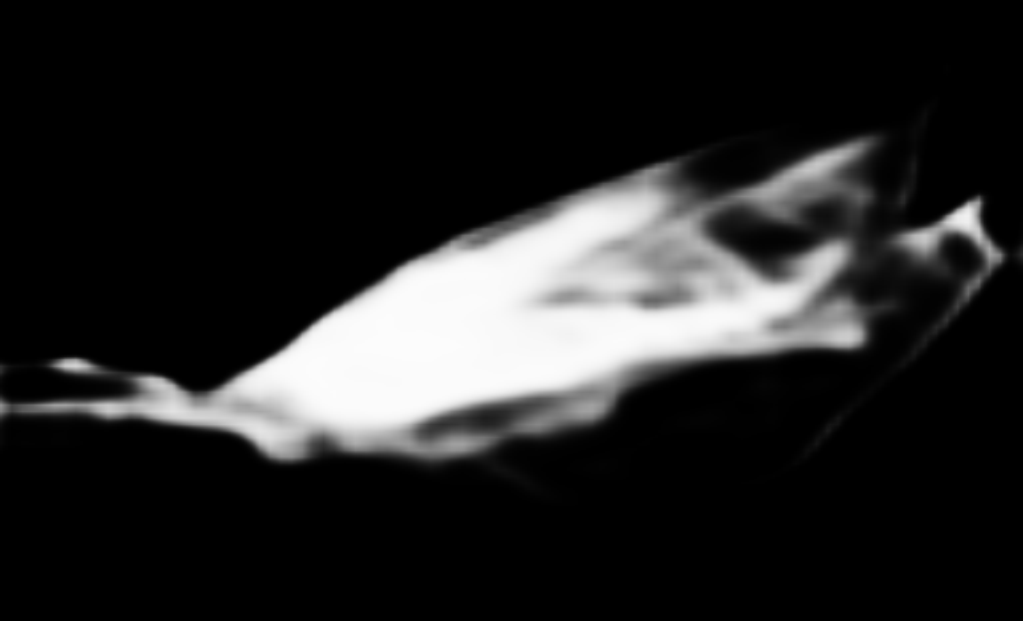}
	\end{subfigure}
	\begin{subfigure}{0.12\textwidth} \includegraphics[width=\textwidth]{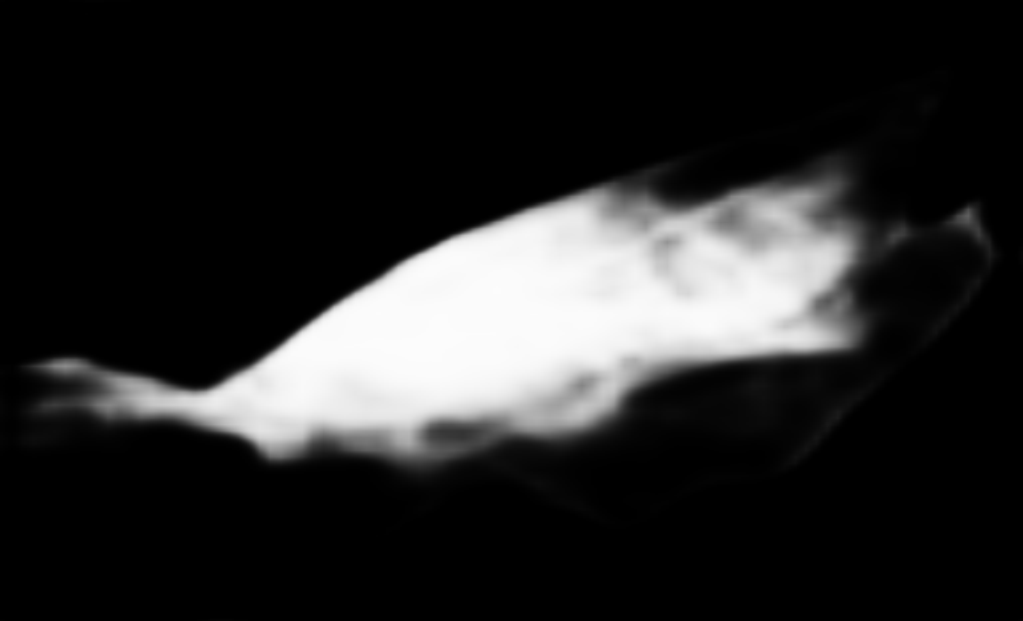}
	\end{subfigure}
    \ \\
	\vspace*{0.5mm}
	\begin{subfigure}{0.12\textwidth}
		\includegraphics[width=\textwidth]{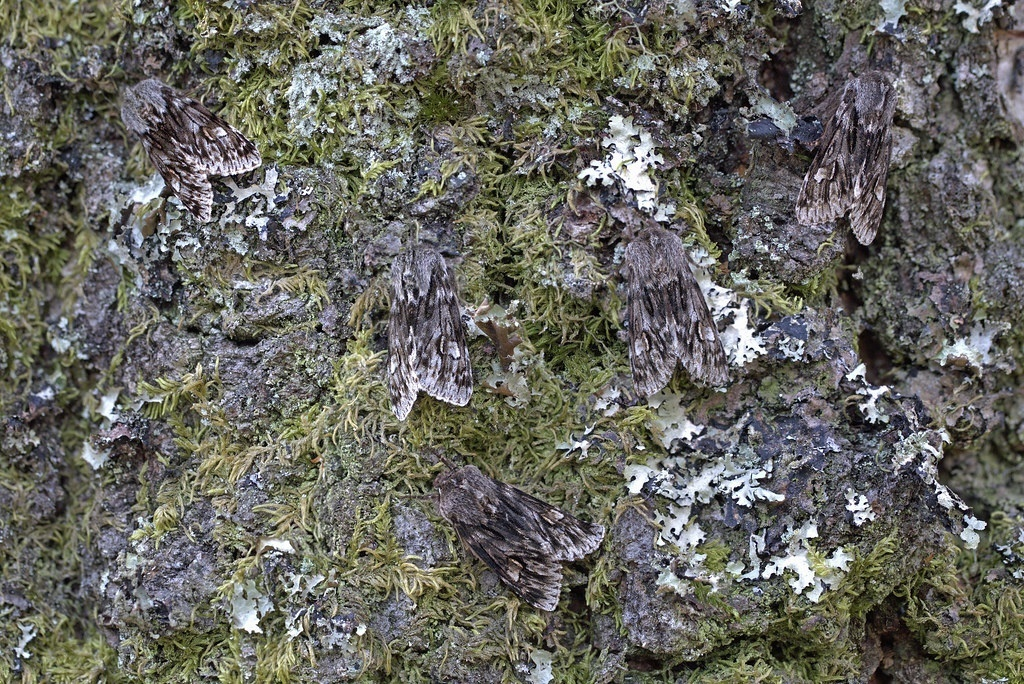}
		\vspace{-5.5mm} \caption{inputs}
	\end{subfigure}
	\begin{subfigure}{0.12\textwidth}
		\includegraphics[width=\textwidth]{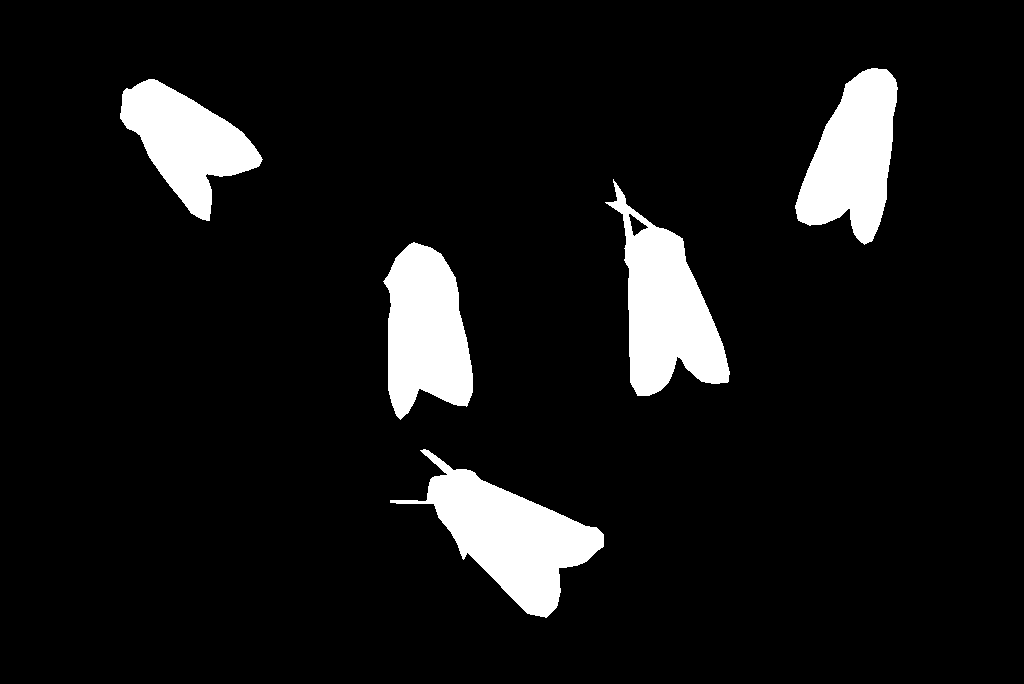}
		\vspace{-5.5mm} \caption{{\footnotesize ground truth}}
	\end{subfigure}
	\begin{subfigure}{0.12\textwidth}
		\includegraphics[width=\textwidth]{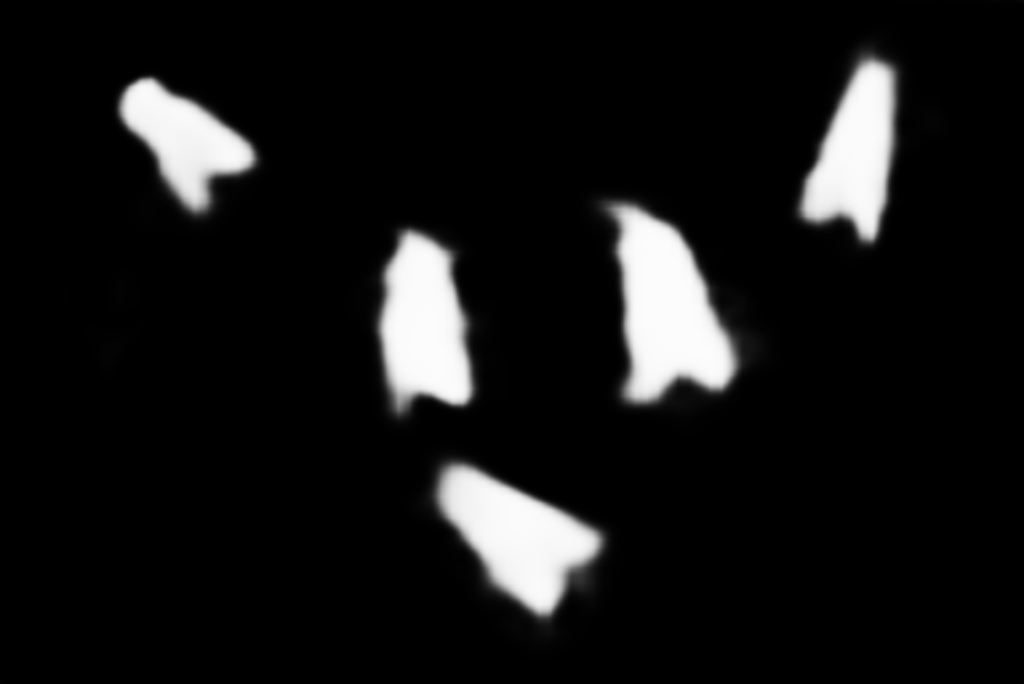}
		\vspace{-5.5mm} \caption{{\footnotesize full pipeline}}
	\end{subfigure}
    \begin{subfigure}{0.12\textwidth}
		\includegraphics[width=\textwidth]{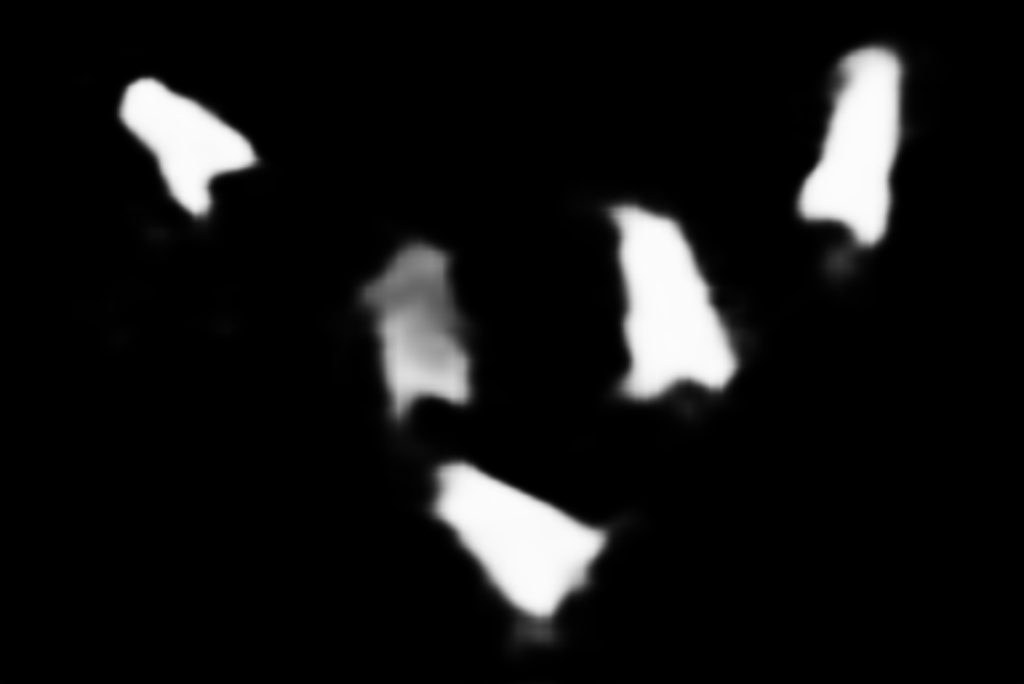}
		\vspace{-5.5mm} \caption{$M_4$}
	\end{subfigure}
    \begin{subfigure}{0.12\textwidth}
		\includegraphics[width=\textwidth]{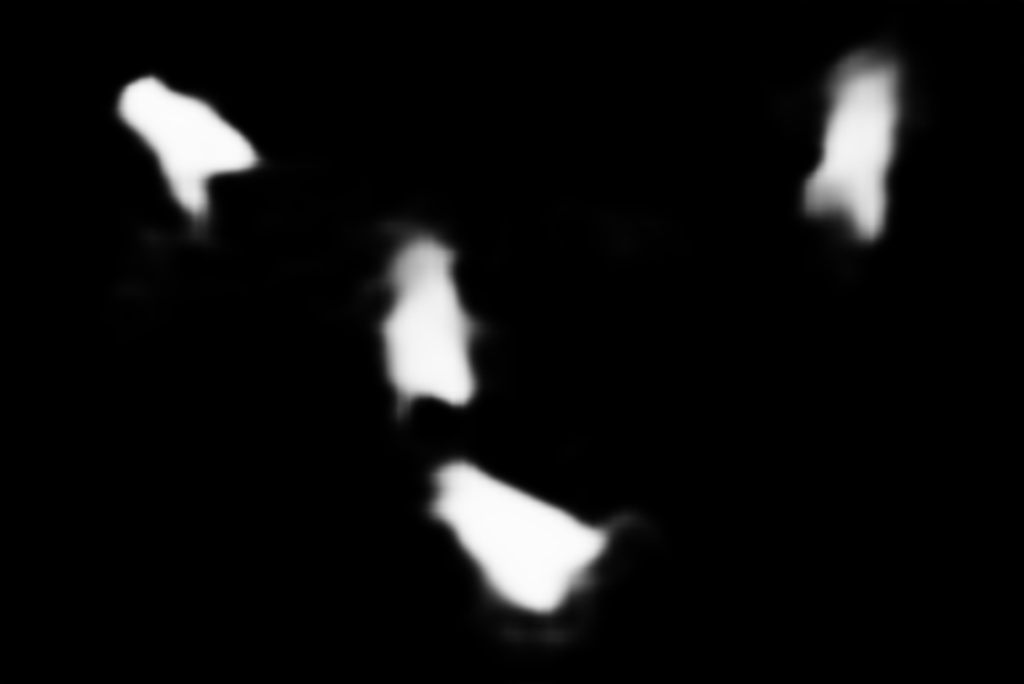}
		\vspace{-5.5mm} \caption{$M_3$}
	\end{subfigure}
		 \begin{subfigure}{0.12\textwidth}
		\includegraphics[width=\textwidth]{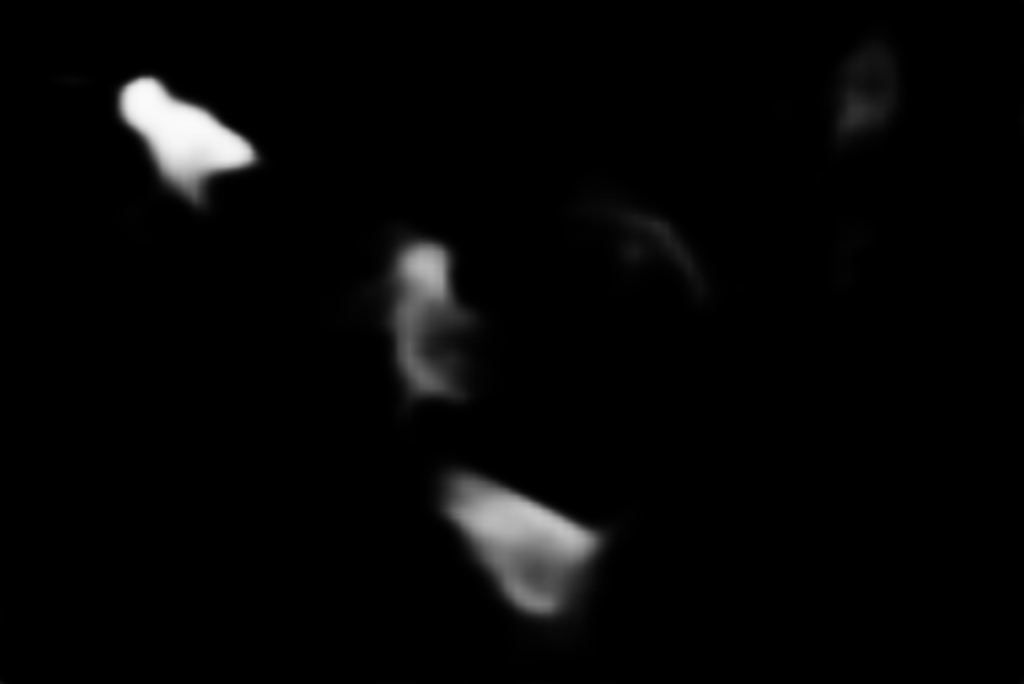}
		\vspace{-5.5mm} \caption{$M_2$}
	\end{subfigure}
	\begin{subfigure}{0.12\textwidth}
		\includegraphics[width=\textwidth]{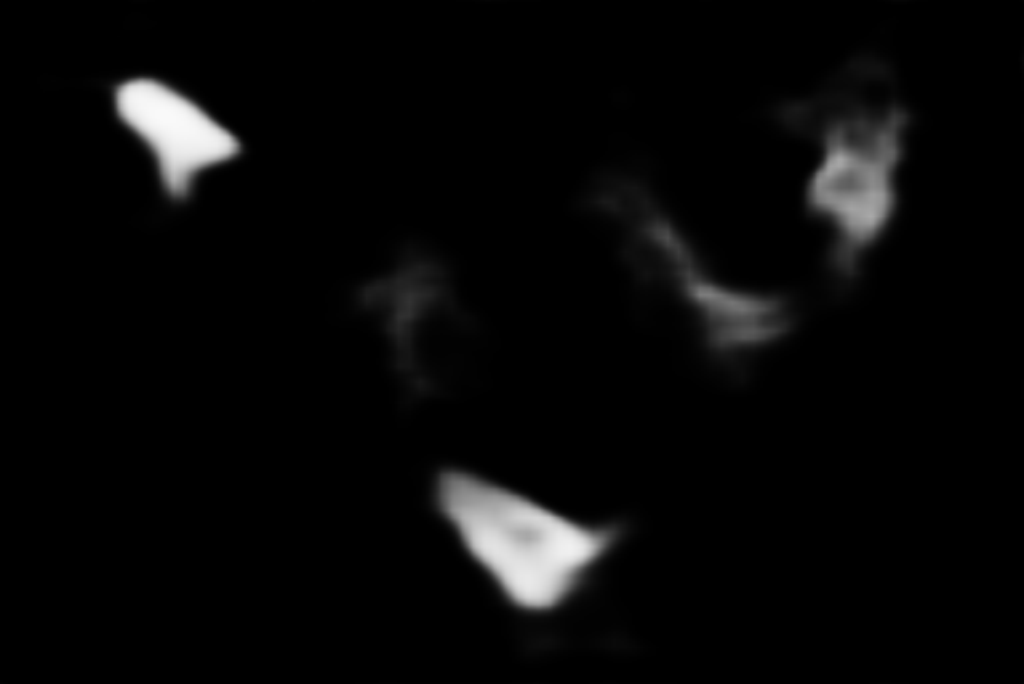}
		\vspace{-5.5mm} \caption{$M_1$}
	\end{subfigure}
	\ \\
    \vspace{-1.5mm}
	\caption{Visual comparison of camouflaged object detection maps produced by our method and baseline networks. (a) input images; (b) ground truths; camouflaged object detection maps produced by (c) our method with a full pipeline, (d)-(g) $M_4$, $M_3$, $M_2$, and $M_1$ (see Table~\ref{table:AB} for $M_1$ to $M_4$). Apparently, our full pipeline can better identify camouflaged objects than all the compared detectors.}
	\label{fig:comparison_real_photos_ablation}
\end{figure*}

\subsubsection{Comparison with SOTA Methods}

We compare our method against 13 cutting-edge methods, including (1) FPN~\cite{lin2017feature}, 
(2) MaskRCNN~\cite{he2017mask}, 
(3) PSPNet~\cite{zhao2017pyramid}, 
(4) UNet++~\cite{zhou2018unet++}, 
(5) PiCANet~\cite{liu2018picanet}, 
(6) MSRCNN~\cite{huang2019mask}, 
(7) BASNet~\cite{qin2019basnet}, 
(8) PFANet~\cite{zhao2019pyramid}, 
(9) CPD~\cite{wu2019cascaded}, 
(10) HTC~\cite{chen2019hybrid}, 
(11) EGNet~\cite{zhao2019egnet}, 
(12)ANet-SRM~\cite{le2019anabranch},
and (13) SINet~\cite{fan2020camouflaged}.
Note that SINet has reported the quantitative results and released camouflaged object maps predicted by all compared COD methods.
Hence, we use these pubic results for conducting fair comparisons.

\paragraph{Quantitative comparisons.} 
Table~\ref{table:state-of-the-art_SD} summarizes the quantitative results of different COD methods on three benchmark datasets.
Apparently, SINet, as a dedicated COD method, has a superior performance on four metrics over other semantic segmentation methods and saliency detectors on three COD benchmark datasets.
Compared to SINet, our method has larger $S_\alpha$, $E_\phi$, and $F_\beta^w$ scores, but a smaller $M$ score, which demonstrates that our method can more accurately identify camouflaged objects.
Specifically, our method has a 3.98\% improvement on the average $S_\alpha$, a 5.21\% improvement on the average $E_\phi$, a 11.41\% improvement on the average $F_\beta^w$, and a 18.26\% improvement on the average $M$ on the three benchmark datasets.
We also re-train our method by taking ResNet-50 as the backbone network, and report the results “TANet v1” in Table~\ref{table:state-of-the-art_SD}, where our method still achieves the best performance.

\paragraph{Visual comparisons.}
Figure~\ref{fig:comparison_real_photos_part1} visually compares the COD maps produced by our network and compared methods.
Apparently, all compared methods tend to include many non-camouflaged regions or neglect parts of camouflaged objects in their COD maps.
On the contrary, our method can more accurately detect camouflaged objects, and our results (see Figure~\ref{fig:comparison_real_photos_part1} (c)) are most consistent with the ground truths shown in Figure~\ref{fig:comparison_real_photos_part1} (b).
\emph{More visual comparisons can be found in the supplementary material.}


\subsection{Ablation Analysis}

We conduct ablation study experiments to verify  the effectiveness of TARMs, and the boundary-consistency loss; see Figure~\ref{fig:TRM}. 
The first baseline (``basic'') equals to remove all RRBs and all TARMs from our network.
The second baseline (``basic+RRB'') is to add  RRBs into ``basic'' to merge features at the adjacent CNN layers. It equals to remove all TARMs from our network.
The third baseline (``basic+RRB+TARM w/o BCL'') is to add TARMs without boundary-consistency loss into the second baseline.
Compared to the four baselines, our method is equal to add boundary-consistency loss of TARMs into the third baseline.

Figure~\ref{fig:comparison_real_photos_ablation} shows the visual comparison results of camouflaged object detection maps produced by our full pipeline and other baseline methods, demonstrating that our full pipeline better identifies camouflaged objects than others. 
Table~\ref{table:AB} compares the results of our network and four baseline networks on three benchmark datasets, \ie, CHAMELEON, CAMO-Test, and COD10K-Test.

\paragraph{Effectiveness of RRBs.} From the results of Table~\ref{table:AB}, ``basic+RRB'' has better metric results than ``basic'', indicating RRBs blocks help our network detect camouflaged objects.

\paragraph{Effectiveness of the affinity loss of TARMs w/o BCL.} As shown in Table~\ref{table:AB}, ``basic+RRB+TARM w/o BCL'' has larger $S_{\alpha}$, $E_{\phi}$, $F_{\beta}^{w}$ scores and a smaller $M$ score than ``basic+RRB'' on three benchmark datasets, demonstrating that it enables our network to better capture the texture difference between camouflaged objects and the background and thus benefits camouflaged object detection.

\paragraph{Effectiveness of the boundary-consistency loss of TARMs.} The superior metric results of Our method over ``basic+RRB+TARM w/o BCL'' on the three benchmark datasets indicate that computing the boundary-consistency loss in TARMs can further improve the camouflaged object detection accuracy of our network by enhancing the consistency across boundaries.


\section{Conclusion}
\label{sec:conclusion}
This paper designs a novel deep network architecture for camouflaged object detection by learning deep texture-aware features.
Our key idea is to amplify the texture difference between camouflaged objects and their surroundings, thus benefiting to identify camouflaged objects from the background. 
In our network, we design a texture-aware refinement module (TARM) that computes the covariance matrices of feature responses among feature channels to represent the texture structures and adopt the affinity loss to learn a set of parameter maps that perform a linear transformation of the convolutional features to separate the texture between camouflaged objects and the background. A boundary-consistency loss is further proposed to learn the object details.
In this way, we can obtain deep texture-aware features and formulate the TANet by embedding multiple TARMs in a deep neural network for the task.
In the end, we evaluate our method on the benchmark dataset, compare it with various state-of-the-art methods, and show the superiority of our method both qualitatively and quantitatively.
In the future, we will explore the potential of our network for generic image segmentation in more complex environments, especially for objects having similar color to the background.

{\small
\bibliographystyle{ieee_fullname}
\bibliography{egbib}
}

\end{document}